\documentclass[10pt]{article} 

\usepackage[preprint]{rlj}

\usepackage{amssymb}            
\usepackage{mathtools}          
\usepackage{mathrsfs}           
\usepackage{graphicx}           
\usepackage{subcaption}         
\usepackage[space]{grffile}     
\usepackage{url}                
\usepackage{lipsum}             

\usepackage{amsmath}
\usepackage[utf8]{inputenc} 
\usepackage[T1]{fontenc}    
\usepackage{url}            
\usepackage{booktabs}       
\usepackage{tabularx}       
\usepackage{amsthm}         
\usepackage{amsmath}         
\usepackage{amsfonts}       
\usepackage{nicefrac}       
\usepackage{graphicx}       
\usepackage{microtype}      
\usepackage{xcolor}         
\usepackage{float} 
\usepackage{placeins}
\usepackage{wrapfig}
\usepackage{tikz-cd}

\usepackage{caption} 
\usepackage{wasysym}

\usepackage{amssymb}            
\usepackage{mathtools}          
\usepackage{mathrsfs}           
\usepackage{graphicx}           
\usepackage{subcaption}         
\usepackage[space]{grffile}     
\usepackage{url}                
\usepackage{lipsum}             

\usepackage{float}
\usepackage{tabularx}       
\usepackage{booktabs}       
\usepackage{placeins}

\newtheorem{theorem}{Theorem}

\usepackage{titlesec}

\usepackage{lipsum}
\usepackage{wasysym}
\usepackage{comment}
\usepackage{ bm}

\usepackage[ruled,vlined]{algorithm2e}
\usepackage[noend]{algpseudocode}

\usepackage[nameinlink, capitalize]{cleveref}
\usepackage{subcaption}
\makeatletter

\newcounter{mSubsec}
\newcommand{\msection}[1]{%
  \section*{M Methods}%
  \addcontentsline{toc}{section}{M Methods}%
  \setcounter{mSubsec}{0}%
  #1%
}
\newcommand{\msubsection}[1]{%
  \refstepcounter{mSubsec}%
  \subsection*{M.\themSubsec\quad #1}%
  \addcontentsline{toc}{subsection}{M.\themSubsec\quad #1}%
  \phantomsection%
  \def\@currentlabelname{#1}%
  \label{m:\themSubsec}%
}

\newcounter{rSubsec}
\renewcommand{\therSubsec}{\arabic{rSubsec}}
\newcommand{\rsection}[1]{%
  \section*{R Related Work}%
  \addcontentsline{toc}{section}{R Related Work}%
  \setcounter{rSubsec}{0}%
  #1%
}
\newcommand{\rsubsection}[1]{%
  \refstepcounter{rSubsec}%
  \subsection*{R.\therSubsec\quad #1}%
  \addcontentsline{toc}{subsection}{R.\therSubsec\quad #1}%
  \phantomsection%
  \def\@currentlabelname{#1}%
  \label{r:\therSubsec}%
}

\newcounter{eSubsec}
\renewcommand{\theeSubsec}{\arabic{eSubsec}}
\newcommand{\esection}[1]{%
  \section*{E Additional Experiments}%
  \addcontentsline{toc}{section}{E Additional Experiments}%
  \setcounter{eSubsec}{0}%
  #1%
}
\newcommand{\esubsection}[1]{%
  \refstepcounter{eSubsec}%
  \subsection*{E.\theeSubsec\quad #1}%
  \addcontentsline{toc}{subsection}{E.\theeSubsec\quad #1}%
  \phantomsection%
  \def\@currentlabelname{#1}%
  \label{e:\theeSubsec}%
}

\newcounter{pSubsec}
\renewcommand{\thepSubsec}{\arabic{pSubsec}}
\newcommand{\psection}[1]{%
  \section*{P Proofs}%
  \addcontentsline{toc}{section}{P Proofs}%
  \setcounter{pSubsec}{0}%
  #1%
}
\newcommand{\psubsection}[1]{%
  \refstepcounter{pSubsec}%
  \subsection*{P.\thepSubsec\quad #1}%
  \addcontentsline{toc}{subsection}{P.\thepSubsec\quad #1}%
  \phantomsection%
  \def\@currentlabelname{#1}%
  \label{p:\thepSubsec}%
}

\newcounter{aSubsec}
\renewcommand{\theaSubsec}{\arabic{aSubsec}}
\newcommand{\asection}[1]{%
  \section*{A Algorithms}%
  \addcontentsline{toc}{section}{A Algorithms}%
  \setcounter{aSubsec}{0}%
  #1%
}
\newcommand{\asubsection}[1]{%
  \refstepcounter{aSubsec}%
  \subsection*{A.\theaSubsec\quad #1}%
  \addcontentsline{toc}{subsection}{A.\theaSubsec\quad #1}%
  \phantomsection%
  \def\@currentlabelname{#1}%
  \label{a:\theaSubsec}%
}

\makeatother
\title{The challenge of hidden gifts in multi-agent reinforcement learning}

\setrunningtitle{The challenge of hidden gifts}

\author{Dane Malenfant\textsuperscript{*\enskip$\dag$\enskip$\mercury$}, Blake A. Richards\textsuperscript{\enskip$\dag$ \enskip$\mercury$\enskip$\P$\enskip$\ddag$}}

\emails{dane.malenfant@mail.mcgill.ca, \ \ blake.richards@mila.quebec}

\affiliations{
$^{\dag}$\hspace*{0.5em}\textbf{School of Computer Science, McGill University}\\
$^{\mercury}$\hspace*{0.5em}\textbf{Mila - The Québec AI Institute}\\
$^{\P}$\hspace*{0.5em}\textbf{ CIFAR Learning in Machines and Brains}\\
$^{\ddag}$\hspace*{0.5em}\textbf{Dept of Neurology \& Neurosurgery, and Montreal Neurological Institute, McGill University }
\par *corresponding author
}

\contribution{We introduce the Manitokan task, which induces a structural credit assignment problem involving hidden gifts.}
{Prior MARL environments study cooperation through shared rewards or social dilemmas, and gifting mechanisms have been explored as reward transfer \citep{hughes2018inequity, peysakhovich2018prosocial}. However, existing tasks do not isolate the credit assignment challenges created when cooperative actions are hidden and reciprocity becomes uncertain.}

\contribution{We provide empirical evidence that several state-of-the-art MARL credit assignment algorithms fail to solve the Manitokan task, even with recurrent policies, despite the small state and action space of the environment.}
{Many MARL algorithms have been designed to address credit assignment in cooperative tasks through centralized critics, value factorization, or specialized architectures. Our results show that these approaches can still struggle with hidden gifts.}

\contribution{We present a theoretical analysis of the Manitokan credit assignment problem and derive a learning-awareness correction term inspired by gradient steering in learning-aware MARL \citep{foerster2017learning}.}
{Learning-aware MARL methods model how one agent's policy update influences another agent's learning \citep{foerster2017learning, willi2022cola}. By analyzing this problem, we discoverer another learning aware gradient correction term. }

\contribution{We propose a self learning-awareness correction term that is fully \textit{decentralized}, assuming the return can be decomposed into collective and individual rewards , that does not require access to other agents' policies and empirically reduces variance while improving performance on the task \citep{bernstein2002complexity}.}
{Prior learning-aware MARL methods typically require access to other agents' policies or gradients \citep{foerster2017learning, willi2022cola}, which limits their applicability in decentralized settings.}

\keywords{Cooperation, Multi-Agent, Trust, Reinforcement Learning, Self-Learning Awareness} 

\summary{
Agents can sometimes benefit from the actions of others even when those actions are unobserved. For example, a neighbor leaving a parking spot open benefits you even if you never see them choose not to take it. We refer to such actions as ``hidden gifts,'' which create a challenging credit assignment problem in multi-agent reinforcement learning (MARL).

We study this problem using a simple grid-world task. Agents must unlock their own doors for individual rewards, while a larger collective reward is obtained only if all agents unlock their doors. However, there is a single shared key, so the collective reward requires agents to drop the key for others after using it. Because agents cannot observe or infer when others drop the key, these acts of cooperation are hidden gifts.

We show that several state-of-the-art MARL algorithms fail to learn the collective solution in this task. Decentralized actor-critic policy gradient agents can solve it when given their own action history, but MARL-specific architectures still fail. We further derive a learning-aware correction term for policy gradient agents that reduces variance and improves convergence to collective success. These results highlight the difficulty of credit assignment under hidden cooperation and show that decentralized self learning-awareness can help address it.
}

\begin{document}

\makeCover  
\maketitle  

\begin{abstract}
Sometimes we benefit from actions that others have taken even when we are uncertain that they took those actions. For example, if your neighbor chooses not to take a parking spot in front of your house when you are not there, you can benefit, even without being aware that they took this action. These "hidden gifts" represent an interesting challenge for multi-agent reinforcement learning (MARL), since assigning credit when the beneficial actions of others are hidden is non-trivial. Here, we study the impact of hidden gifts with a simple MARL task. In this task, agents in a grid-world environment have individual doors to unlock in order to obtain individual rewards. As well, if all the agents unlock their door the group receives a larger collective reward. However, there is only one key for all of the doors, such that the collective reward can only be obtained when the agents drop the key for others after they use it. Notably, there is nothing to indicate to an agent that the other agents have dropped the key, thus the act of dropping the key for others is a "hidden gift". We show that several different state-of-the-art MARL algorithms, including MARL specific architectures, fail to learn how to obtain the collective reward in this simple task. Interestingly, we find that decentralized actor-critic policy gradient agents can solve the task when we provide them with information about their own action history, but MARL agents still cannot solve the task with action history. Finally, we derive a correction term for these policy gradient agents, inspired by learning aware approaches, which reduces the variance in learning and helps them to converge to a collective success more reliably. These results show that credit assignment in multi-agent settings can be particularly challenging in the presence of "hidden gifts", and demonstrate that self learning-awareness in decentralized agents can benefit these settings.
\end{abstract}

\addtocontents{toc}{\protect\setcounter{tocdepth}{-10}}
\section{Introduction}

In the world we often rely on other people to help us accomplish our goals. Sometimes, people help us even when we are not aware of it or haven't communicated a need for it. One simple example would be if someone decides not to take the last cookie in the pantry, leaving it for others. Another interesting example is the historical ``Manitokan'' practice of the plains Indigenous nations of North America. In an expansive environment with limited opportunities for communication, people would cache goods for others to use at effigies \citep{manitoh}. Notably, in these cases there was no explicit agreement of a trade or articulation of a ``tit-for-tat''\citep{axelrod1980effective}. Rather, people simply engaged in altruistic acts that others could then benefit from, even without knowing who had taken the altruistic act. We refer to these undeclared altruistic acts as ``hidden gifts''.

Hidden gifts represent an interesting challenge for credit assignment in multi-agent reinforcement learning (MARL). If one leaves a hidden gift, assigning credit to the actions of another is essentially impossible, since the action was never made clear to the beneficiary. As such, standard Bellman-back-ups \citep{bellman1954theory} would likely be unable to identify the critical steps that led to success in the task. Moreover, unlike a scenario where cooperation and altruistic acts can emerge through explicit agreement or a strategic equilibrium \citep{nash1950equilibrium}, as in general sum games \citep{axelrod1980effective}, with hidden gifts the benefits of taking an altruistic action are harder to identify or reciprocate.

To explore the challenge of hidden gifts for MARL we built a  grid-world task where hidden gifts are required for optimal behavior \citep{chevalier2023minigrid}. We call it the Manitokan task, in reference to the "take what you need, leave what you don't need" inspiration from Manitokan of plains Indigenous communities.  In the Manitokan task, two-or-more agents are placed in an environment where each agent has a ``door'' that they must open in order to obtain an individual, immediate, small reward. As well, if all of the agents successfully open their door then a larger, collective reward is given to all of them. To open the doors, the agents must use a key, which the agents can both pick up and drop. However, there is only a single key in the environment. As such, if agents are to obtain the larger collective reward then they must drop the key for others to use after they have used it themselves. The agents receive an egocentric, top-down partial image of the environment as their observation in the task, and they can select actions of moving in the environment, picking up a key, dropping a key, or opening a door.  Since the agents do not have access to other agent's decision making process, key drops represent a form of hidden gift -- which make the credit assignment problem challenging. In particular: \textbf{1.} The task is fully cooperative so there is no disincentive for leaving the key, and \textbf{2.} dropping the key only leads to the collective reward if the other agents exploits the gift.  

We tested several state-of-the-art MARL algorithms on the Manitokan task. Specifically we tested Value Decomposition Networks (VDN, QMIX and QTRAN) \citep{sunehag2017value,son2019qtran,rashid2020monotonic}, Multi-Agent and Independent Proximal Policy Optimization (MAPPO and IPPO) \citep{schulman2017proximal, yu2022surprising}, counterfactual multi-agent policy gradients (COMA) \citep{foerster2018counterfactual,ata}, Multi-Agent Variational Exploration Networks (MAVEN) \citep{mahajan2019maven}, an information bottleneck based Stateful Active Facilitator (SAF) \citep{liu2023stateful} and standard actor-critic policy gradients (PG) with Actor-Critic \citep{williams1992simple, sutton1999policy,sutton1998reinforcement, ata}. Notably, we found that none were capable of learning to drop the key and obtain the collective reward reliably. In fact, many of the MARL algorithms exhibited a total removal of key-dropping behavior, leading to less than random performance on the collective reward. These failures held even when we provided the agents with objective relevant information, providing inputs indicating which doors were open and whether the agents were holding the key. 

Interestingly, when we also provided the agents with a history of their own actions as one-hot vectors, we observed that policy gradient agents without proximal policy optimization could now solve the collective task, whereas others still failed. However, these successful agents' showed high variability in cooperation. Based on this, we analyzed the value estimation problem for this task formally, and observed that the value function necessitates an approximation of a non-constant reward. That is, the collective reward is conditioned on the other agent's policy which is non-stationary between policy updates. Inspired by learning awareness \citep{willi2022cola,foerster2017learning}, we derived a new term in the policy gradient theorem which corresponds to the Hessian of the collective reward objective partitioned by the other agent's policy with respect to the collective reward. Using this correction term, we show that we can reduce the variance in the performance of the PG agents and achieve consistent learning to drop the key for others.


\section{The Manitokan task for studying hidden gifts} \label{sec:manito}

\begin{figure}[htbp!]
  \centering
  \includegraphics[height=6.5cm]{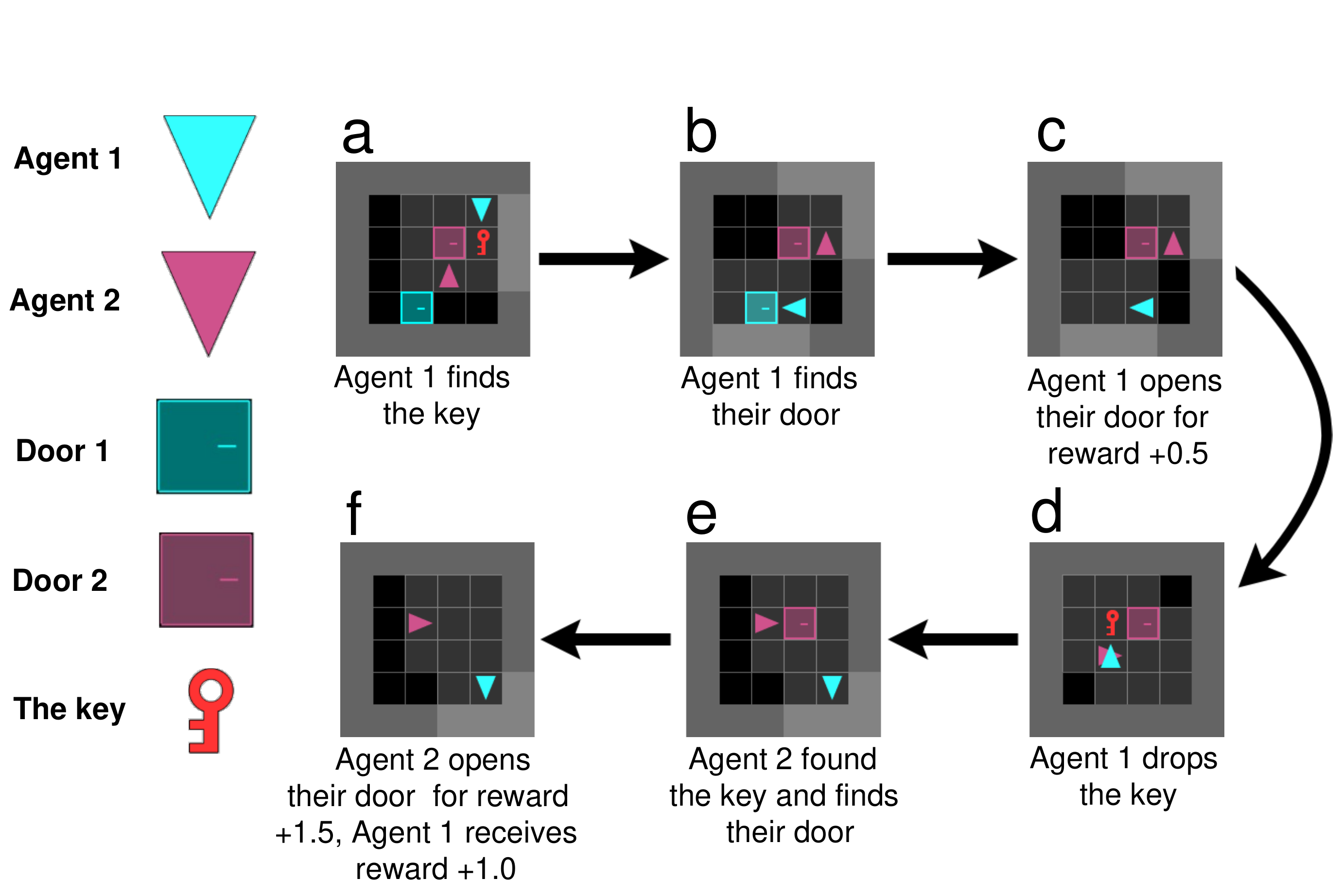}
  \caption{The deceivingly simple steps to success in the Manitokan task. a) Agent 1 finds the key; b) Agent 1 then finds their door; c) Agent 1 opens their door; d) Agent 1 drops the key as a ``hidden gift''; e) Agent 3 finds their door; f) Agent 2 opens their door.}
  \label{fig:manitohkan}                    %
\end{figure}

The Manitokan task is a cooperative MARL task in a grid world\footnote{Although the agents act in a discrete gird world, each agent's policy is an unobservable part of this world and may be continuous} (see Fig.\ref{fig:manitohkan}). The task has been designed to be more complex than matrix games, such as Iterative Prisoner's Dilemma \citep{axelrod1980effective, chammah1965prisoner}, but capable for mathematical analysis of strategic behaviour and different from past cooperative environments (See \ref{related}). At the beginning of an episode each agent is assigned a locked door (Fig.\ref{fig:manitohkan}A) that they can only open if they hold a key. Agents can pick up the key if they move to the grid location where it is located (Fig.\ref{fig:manitohkan}B). Once an agent has opened their door it disappears and that agent receives a small individual reward immediately (Fig.\ref{fig:manitohkan}C). However, there is only one key for all agents to share and the agents can drop the key at any time if they hold it (Fig.\ref{fig:manitohkan}D). Once the key has been dropped the other agents can pick it up (Fig.\ref{fig:manitohkan}E) and use it to open their door as well (Fig.\ref{fig:manitohkan}F). If all doors are opened a larger collective reward is given to all agents, and at that point, the task terminates. The conditions for the rewards \cref{eq:reward_function} are not mutually exclusive.

We now define the notation that we will use for describing the Manitokan task and analyzing formally. The environment is a decentralized partially observable Markov decision process (Dec-POMDP) with the caveat that the collective reward requires individual rewards \citep{goldman2004decentralized, bernstein2002complexity}. Dec-POMDPs are also a type of partially observable stochastic games \citep{10.5555/1597148.1597262}.

Let $M=(\mathcal{N}, T, \mathcal{T},\mathcal{O}, \mathcal{A}, \Pi, \mathcal{R}, \gamma )$, where: $\mathcal{N}:= \{1,2,\dots,N\}$ is the set of $N$ agents, $T\in \mathbb{N}$ is the maximum timesteps in an episode, $\mathcal{O}:=\times_{i\in \mathcal{N}}O^i$ is the joint observation space for the $N$ agents and $o^i_t\in O^i\rightarrow \mathbb{N}^{3\times 3}$ is a partial observation for an agent $i$ at timestep $t$. This is the only input agents take so the state $\mathcal{S}=\mathcal{O}$, $\mathcal{A}:=\times_{i\in \mathcal{N}} A^i$ is the joint action space and $a^i_t\in A^i$ is the action of agent $i$ at time $t$, $\Pi:=\times_{i\in\mathcal{N}}\pi^i$ is the joint space of individual agent policies, $\mathcal{R}\rightarrow \mathbb{R}$ is the reward function composed of both individual rewards, $r^i_t$, which agents receive for opening their own door (i.e. an individual objective), and the collective reward, $r^c$, which is given to all agents when all doors are opened (i.e. a collective objective) (See equation \ref{eq:reward_function} below.), $\mathcal{T}:\mathcal{O}\times\mathcal{A}\rightarrow\Delta(\mathcal{O})$ is the transition function specifying the probability $\mathcal{T}(o^{i'},\mathcal{R}^i(o^i,a^i)|o^i,a^i)$ that agent $i$ transitions to $o^{i'}$ from $o^i$ by taking action $a^i$ for a reward $\mathcal{R}^i$, and $\gamma\in[0,1)$ is the discount factor.

The observations, $o^i_t$, that each agent receives are egocentric images of the 9 grid locations surrounding the current position of the agent (see the lighter portions in Fig. 1). The key, the doors, and the other agents are all visible if they are in the field of view, but not otherwise (hence the task is partially observable). The actions the agents can select, $a^i_t$, consist of `\textit{move forward}', `\textit{turn left}', `\textit{turn right}', `\textit{pick up the key}', `\textit{drop the key}', and `\textit{open the door}'. Episodes last for $T=150$ timesteps at maximum, and are terminated early if all doors are opened.

The monotonic reward function $\mathcal{R}^i$ is defined as:

\begin{equation}
\label{eq:reward_function}
    \mathcal{R}^i(o_t^i,a_t^i):= 
    \begin{cases} r_t^i=r^i \text{ door opened}\\  
      
    r^c = \sum_j^N r^j  \text{ all doors opened } \\
\end{cases}
\end{equation}
But in correspondence with multi-objective problems, $\mathcal{R}^i$ is scalarized as $\hat{\mathcal{R}}^i=r_i+\omega(t)r^c$ where the preference weighting $\omega(t)$ is the other agent's policy so $\hat{\mathcal{R}}^i=r_i+{}^e\pi^j(a^j_t|o^j_t)r^c$ for agent $i$ and episode $e$ \citep{mossalam2016multi}.
The Manitokan task is unique from other credit assignment work in MARL due to the number of keys being strictly less than the number of agents (see. Section \ref{coop}). This scarcity requires the coordination of gifting the key between agents as a necessary critical step for success and maximizing the cumulative return. But, notably, unlike most other MARL settings the act of dropping the key is not actually observable by other agents when learning a policy. When an agent picks up the key they do not know if they were the first agent to do so or if other agents had held the key and dropped it for them. Thus, key drop acts are ``hidden gifts'' between agents and the task represents a deceptively simple, but actually complex structural credit assignment problem across learning dynamics \citep{tumer2002learning, agogino2004unifying, gupta2021structural}. 

Importantly, the collective reward is delayed relative to any key drop actions. Moreover, key drop actions only lead to reward if the other agents have learned to accomplish their individual tasks. It then follows that the delay between a key drop action and the collective reward being received will be proportional in expectation to the number of agents, rendering a more difficult credit assignment problem for higher values of $N$. In the presented data, we focus on the canonical two-player setting from game theory, where ($N=2$), for analytical tractability and interpretability of a Dec-POMDP.

\section{Results} \label{sec:results}

We begin by testing the ability of various state-of-the-art model-free RL algorithms to solve this task, both multi-agent, and decentralized. For the multi-agent algorithms, we selected ones that are prominently used as baselines for credit assignment in fully cooperative MARL tasks. These included the counterfactual model COMA, the centralized critic multi-agent PPO (MAPPO), and global value mixer algorithms VDN, QMIX and QTRAN \citep{foerster2018counterfactual, yu2022surprising, sunehag2017value,rashid2020monotonic,son2019qtran}. We used actor-critic policy gradient methods, and  gradient decoupled IPPO without a value function. \citep{williams1992simple,sutton1999policy, schulman2017proximal}. In order to alleviate problems with exploration and changing policies we also tested MAVEN (which provides more robust exploration) and SAF (which is a meta-learning approach with a communication protocol network for learning with multiple policies) \citep{mahajan2019maven, liu2023stateful}. All algorithms were built with recurrent components in their policy (specifically, Gated Recurrent Units, GRUs \citep{cho-etal-2014-learning}) in order to provide agents with some information about task history. (See methods in the supplementary \label{supp}for more details on design and training.) In our initial tests we provided only the egocentric (i.e agent's "self" is included) observations as input for the agents. Hyperparameters were optimized by tuning from the sets provided in the original papers with a search to avoid overfitting on the immediate reward.  As well, we trained 10 simulations with different seeds that initialized 32 parallel environments also with different random seeds. These parallel environments make the reward signals in each batch less sparse. For each simulation we ran 10,000 episodes for each 32 parallel environments, except in Figure \ref{fig:6} where we did 26,000 episodes. Training was done with 2 CPUs for each run  and SAF required an additional A100 GPU per run. An emulator was also used to improve environment step speed \citep{suarez2024pufferlib}.

\subsection{All algorithms fail in the basic Manitokan  task} \label{sec:fail}
\begin{figure}[htbp]
    \centering
    \begin{subfigure}[b]{0.40\textwidth}
        \centering
        \includegraphics[width=5.cm]{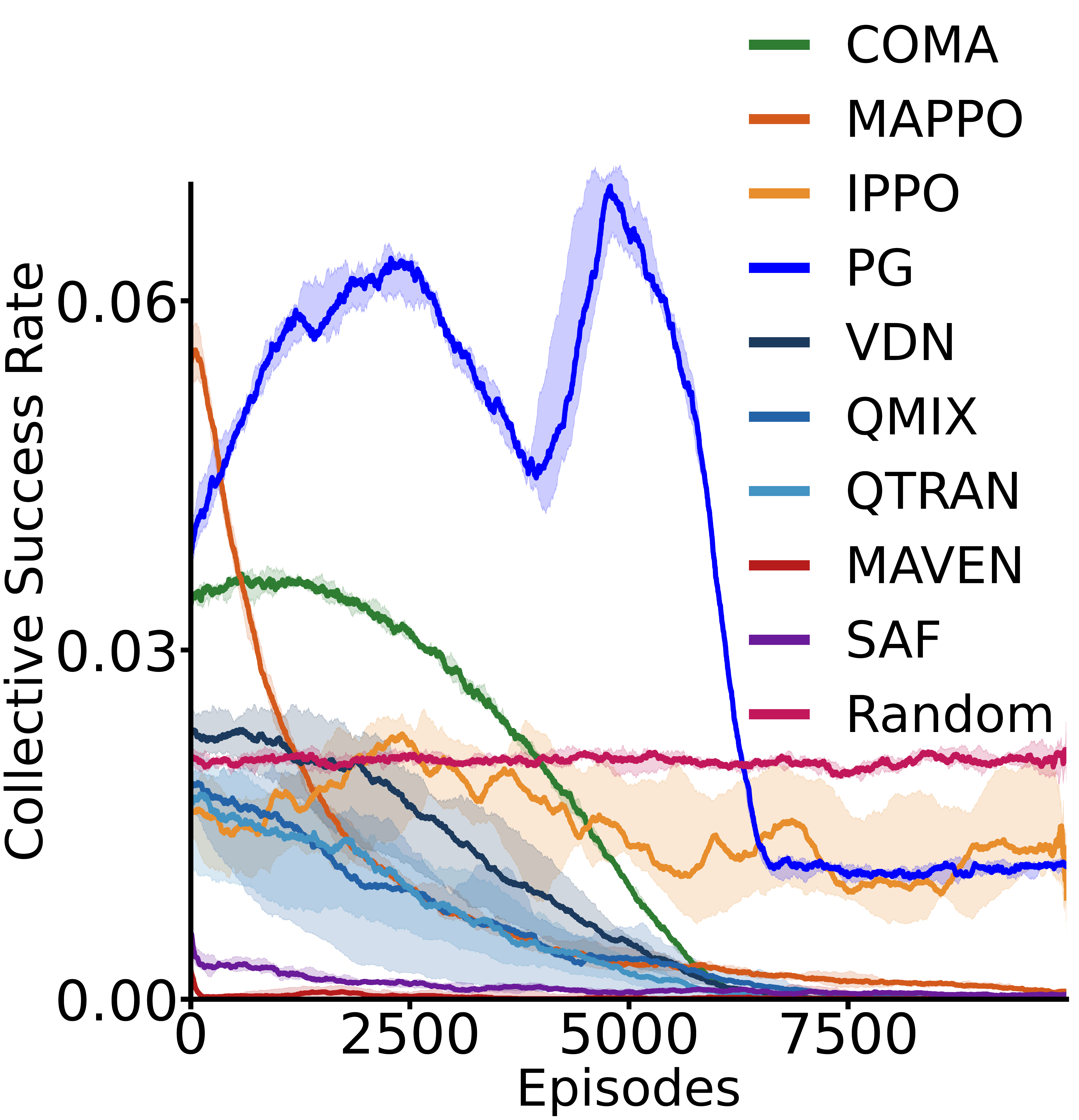}
        \caption{Collective Success Rate}
        \label{fig:2a}
    \end{subfigure}
    \hspace{0.001\textwidth}
    \begin{subfigure}[b]{0.40\textwidth}
        \centering
       \includegraphics[width=6.cm]{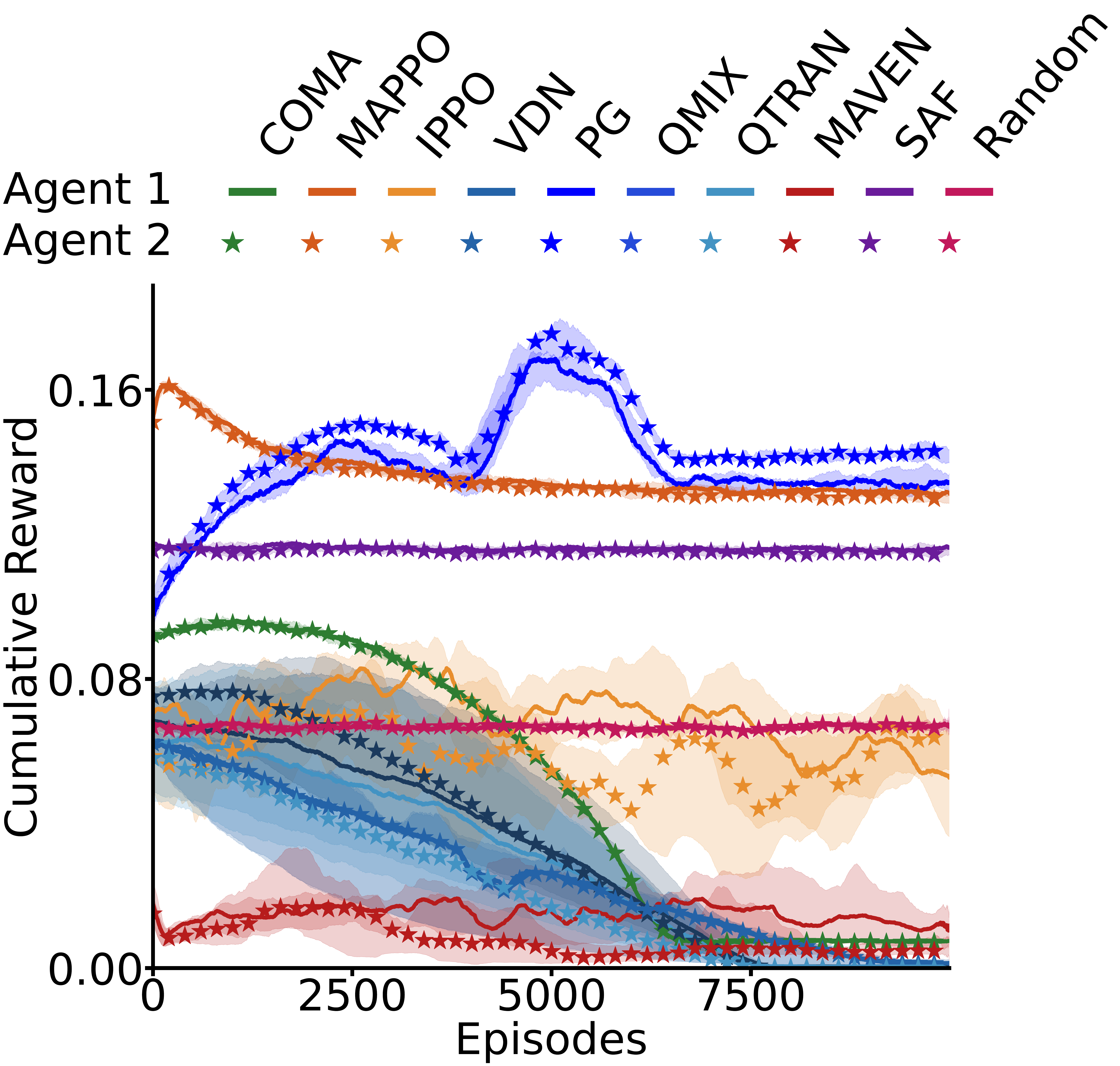}
        \caption{Cumulative Reward}
        \label{fig:2b}
    \end{subfigure}
    \caption{a) Success rate for the collective reward, i.e. percentage of trials where both agents opened their doors. b) Cumulative reward of both agents across 10000 episodes with 32 parallel environments limited to 150 timesteps each.
 }
    \label{fig:2}
\end{figure}

To our surprise, everything we tested converged to a level of success in obtaining the collective reward that was \textit{below} the level achieved by a fully random policy (Fig. \ref{fig:2a}) even though reward was being maximized and the single agent key-to-door task is solvable (see E.\ref{e:single_agents}). In fact, with the sole exception of MAPPO, all of the MARL algorithms we tested (COMA, VDN, QMIX, QTRAN) exhibited full collapse in hidden gift behavior: these algorithms all converged to policies that involved \textit{less} than random key dropping frequency. Randomizing the policy can slightly improve success rate but reduced cumulative reward (E. \ref{rand}). Notably, the agents that didn't show full collapse in collective success (MAPPO, IPPO and SAF) were still successfully opening their individual doors, since their cumulative reward was higher than that of a random policy (Fig. \ref{fig:2b}).  But, the MARL agents that showed total collapse of collective behavior also showed collapse in the individual rewards. We believe that this was due to the impact of asymmetric state information and shared value updates. With shared value updates the reward signal could be swamped by noise from the unrewarded agents in the absence of key drops, and became confused by a lack of reward obtained when agents' dropped the key before opening their doors (See more below in section \ref{sec:formal}). The key drop rate is optimal at $1$, eg. one drop after using the key, all agents had a near zero drop rate or did not seem to learn (E.\ref{e:keys}).

\subsection{Observability of door and key status does not rescue performance in the Manitokan task} \label{sec:obs}
To receive the collective reward, agents needed to learn to pick up the key, use it, then drop it. If they did these actions out of sequence (e.g. dropping the key before using it), then they can not succeed. As such, one potential cause for collapse in performance could have been the fact that agents did not have an explicit signal for their door being opened or that they are holding the key (i.e. the task was partially observable with respect to these variables). To make the task easier, we provided the agents with more decentralized information, one which indicated whether their door was open, the other which indicated whether they held the key. The agents now always have a cue when their individual task is completed.
\FloatBarrier
\begin{figure}[!htbp]
    \centering
    \begin{subfigure}[b]{0.40\textwidth}
        \centering
        \includegraphics[width=4.8cm]{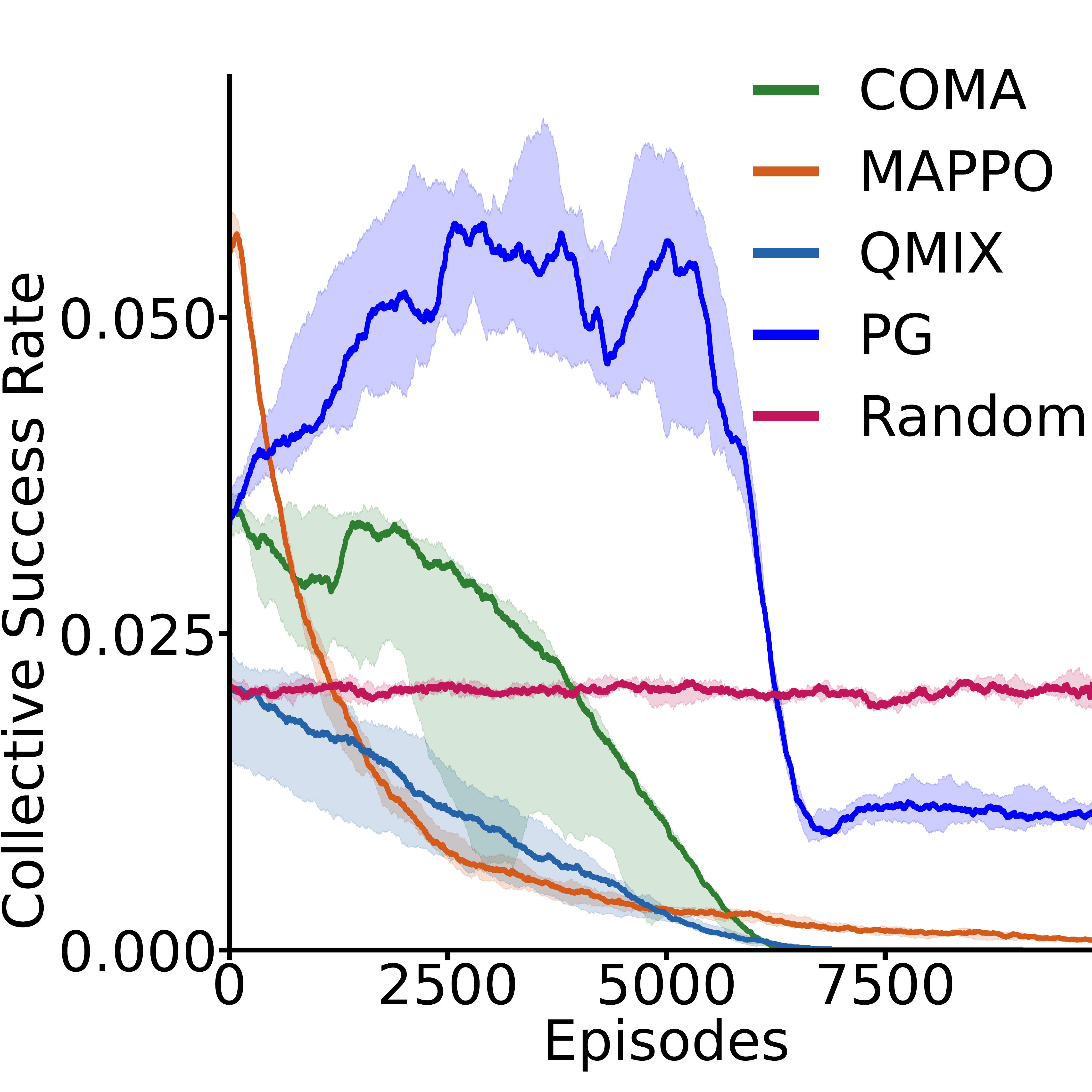}
         \caption{Collective Success Rate}
        \label{fig:3a}
    \end{subfigure}
    \hspace{0.01\textwidth}
    \begin{subfigure}[b]{0.40\textwidth}
        \centering
       \includegraphics[width=4.8cm]{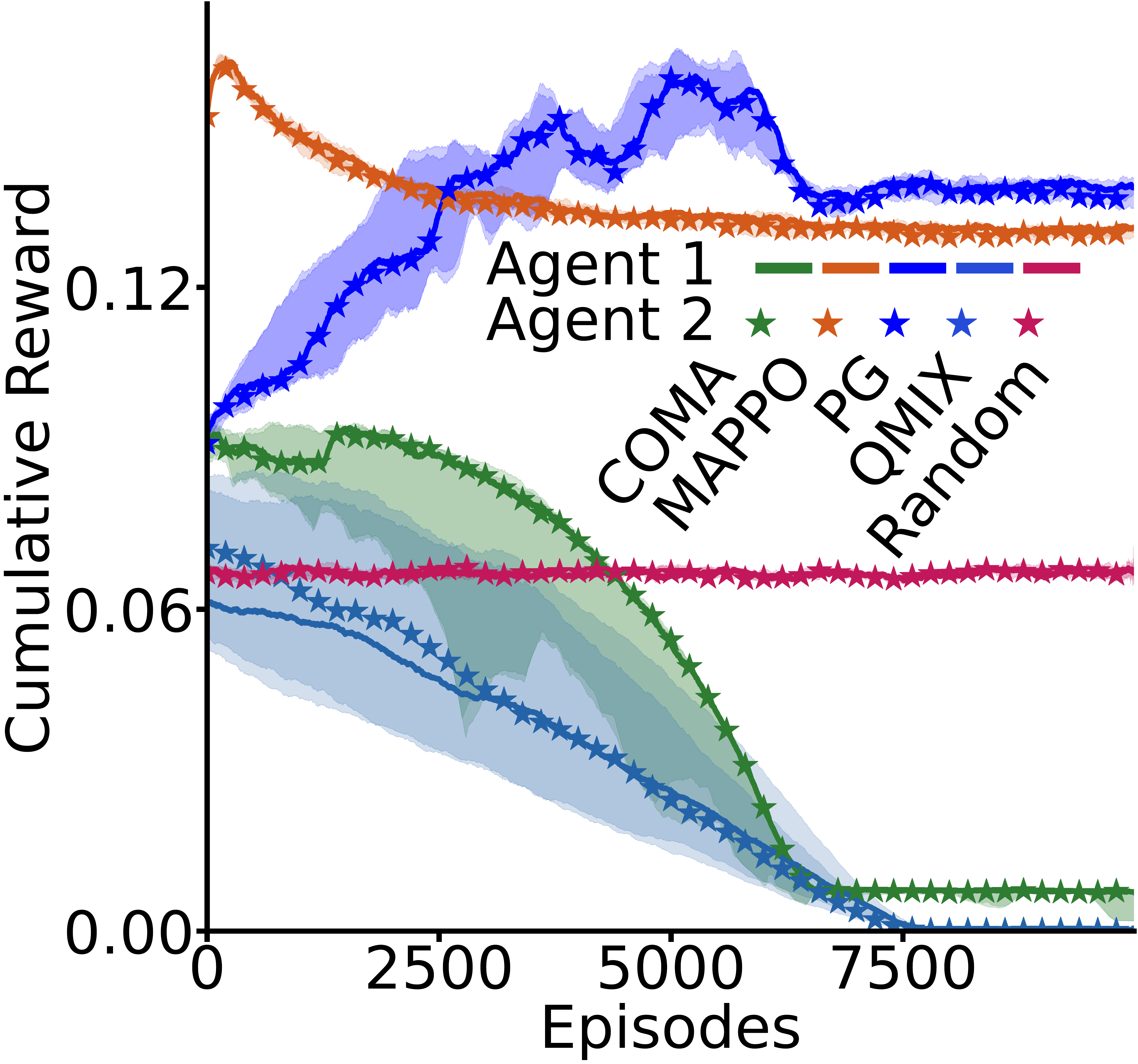}
        \caption{Cumulative Reward}
        \label{fig:3b}
    \end{subfigure}
    \caption{a) Success rate when each agent receives information about whether they have opened their door or not and if they have the key or not. b) Cumulative reward of both agents with information about whether they have opened their door or not and if they have the key or not.
 }
    \label{fig:3}
\end{figure}
Surprisingly, the agents we tested all failed to achieve collective success rates above random. In fact, the same behavior occurred, with the MARL agents (MAPPO, QMIX, COMA) showing total collapse, and the decentralized PG agents showing some collective success, but still below random (Fig. \ref{fig:3a}). As before, We found that only MAPPO and decentralized PG showed any learning in the task, with QMIX and COMA showing collapse in the individual success rate as well (Fig. \ref{fig:3b}). Thus, the lack of information about the status of the door and key was not the cause of failure of the Manitokan task.




\subsection{Adding action history helps decentralized agents but not MARL agents} \label{sec:hist}
Next, we reasoned that a cause of failure was that agents could not see themselves drop the key. To alleviate the credit assignment, we provided the agents with the last action they took as a one-hot vector. Coupled with the recurrence, this would permit the agents to know that they had dropped the key in the past when the collective reward was obtained.

\begin{figure}[htbp]
    \centering
    \begin{subfigure}[b]{0.40\textwidth}
        \centering
        \includegraphics[width=5.cm]{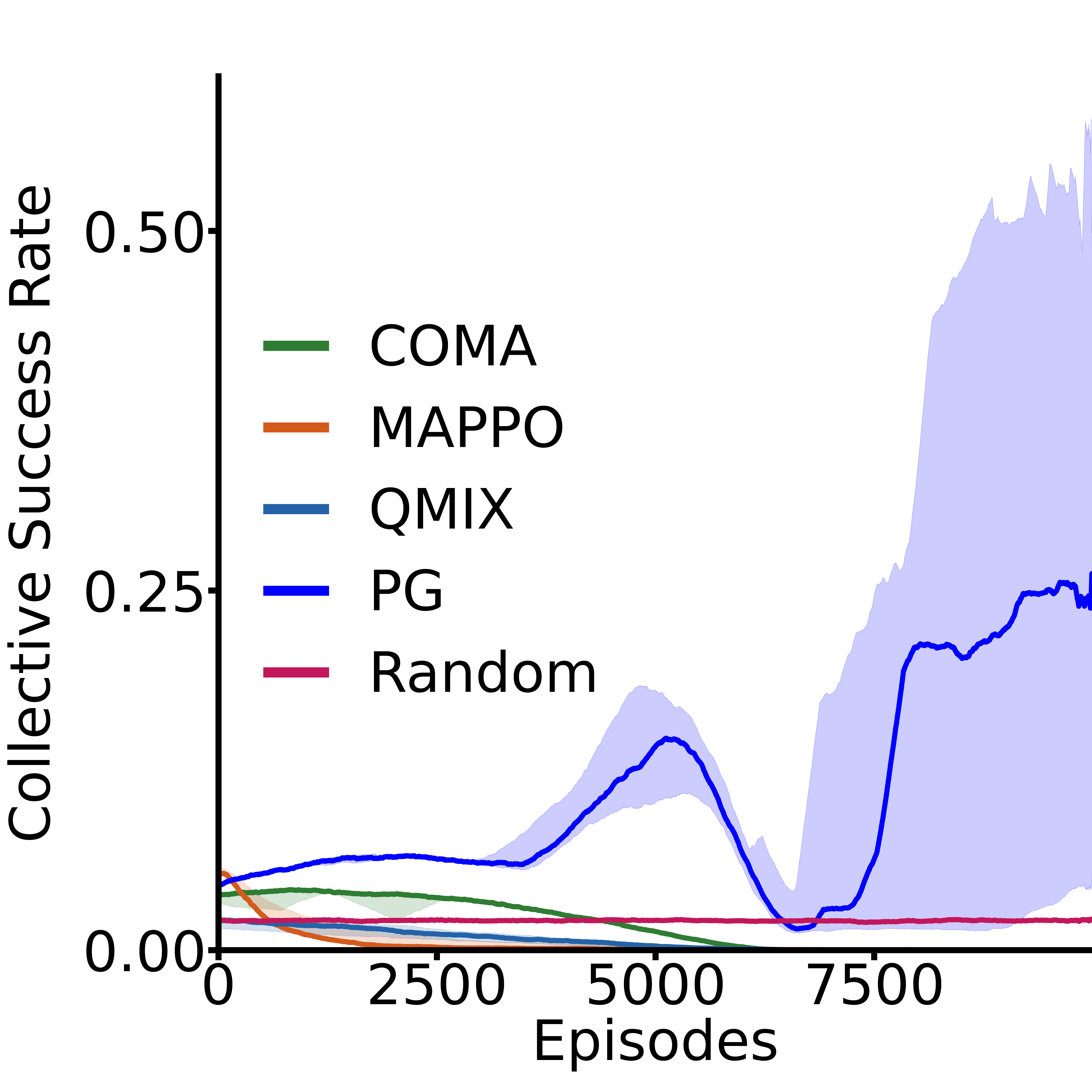}
        \caption{Collective Success Rate}
        \label{fig:4a}
    \end{subfigure}
    \hspace{0.01\textwidth}
    \begin{subfigure}[b]{0.40\textwidth}
        \centering
       \includegraphics[width=5.cm]{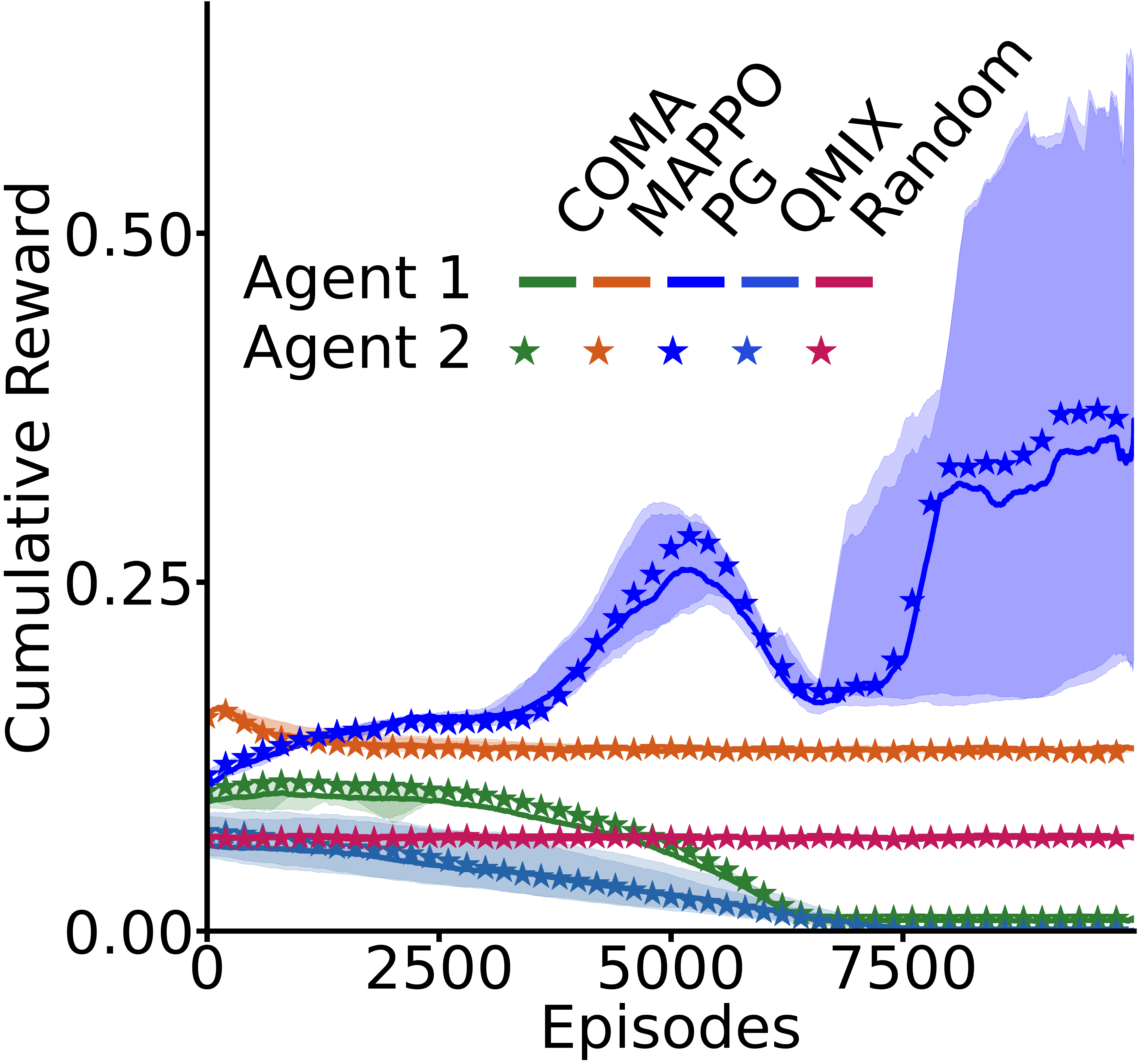}
        \caption{Cumulative Reward}
        \label{fig:4b}
    \end{subfigure}
    \caption{a) Success rate when each agent receives their last action in the observation. b) Cumulative reward of both agents with last action  information.}
    \label{fig:4}
\end{figure}

When we added the past action to the observation, we found that the PG agents now showed signs of obtaining the collective reward, much better than random (Fig. \ref{fig:4}). This also led to better cumulative reward (Fig. \ref{fig:4}). However, interestingly, the other agents showed no ability to learn this task, exhibiting the same collapse in collective success rate and same low levels of cumulative reward as before (Fig. \ref{fig:4a} \& \ref{fig:4b}). These results indicated that there is something about the credit assignment problem in the Manitokan task that can be addressed by the standard policy gradient objective, but not trust region mechanisms. Further modifying the reward function can  help or inhibit these agents but removing one of the rewards harms success (E. \ref{e:rmod}). Changing or randomizing agent turn order also reduces success rate (E. \ref{e:turnorder}). Overall, the PG agents still exhibited very high variance in their collective success rate (Fig. \ref{fig:4}), suggesting more to the credit assignment problem. We then formally analyzed the value function of the task to better understand the credit assignment problem therein.

\section{Formal analysis and a correction term}
\label{sec:formal}

For ease of Dec-POMDP analysis we again focus on the situation where $N=2$, i.e. there are only two agents, and borrow the language of \textit{sub-policies} from options and hierarchical learning \citep{sutton1999between, bakker2004hierarchical,andreas2017modular} to disassociate what needs to be learned to acquire both scalar rewards in an episode which notably are in conflict: Finding and holding the key is opposite from avoiding it. 

We begin by considering the objective function $J$ for agent $i$ with parameters $\Theta^i$, for an entire successful episode of the Manitokan task where expectations are over trajectories $\tau$ sampled from the the current policies $\pi^i,\pi^j$ of the agents given a randomly initialized observation:

\begin{equation}\label{eq:2}
    J(\Theta^i) = \mathbb{E}_{\tau\sim \pi^i,\pi^j}[\sum_{t=0}^T \gamma^t\mathcal{\hat{R}}^i(o_t^i,a_t^i)]
\end{equation}

Taking this objective from the perspective of agent $i$, we can factorize it into the individual and collective terms:

\begin{equation}\label{eq:2}
\begin{split}
    J(\Theta^i) 
    & = \mathbb{E}_{\tau\sim \pi^i, \pi^j}[\sum_{t=0}^T \gamma^tr_t^i+ \gamma^tr^c] \\ 
    & = \mathbb{E}_{\tau\sim \pi^i, \pi^j}[\sum_{t=0}^{T-1} \gamma^tr_t^i] + \mathbb{E}_{\tau\sim \pi^i, \pi^j}[\sum_{t=T}^T \gamma^tr^c] \\ 
    & = \mathbb{E}_{\tau\sim \pi^i, \pi^j}[\sum_{t=0}^{T-1} \gamma^tr_t^i] + \mathbb{E}_{\tau\sim \pi^i, \pi^j}[ \gamma^Tr^c] \\
    &= J_d(\Theta^i) + J_c(\Theta^i)
\end{split}
\end{equation}

If we consider the sub-objective related solely to the collective reward by assuming agent $i$ acquired the individual reward $r^i_t$ first $ J_c(\Theta^i) = J(\Theta^i) - J_d(\Theta^i) = \mathbb{E}_{\tau\sim \pi^i, \pi^j}\!\left[\sum_{t=0}^{T-1} \gamma^t r^{c,i}\right] = \mathbb{E}_{\tau\sim \pi^i, \pi^j}\!\left[\gamma^T r^c\right]$\footnote{The collective reward $r^c$ is a terminal reward and can only appear at the terminal time step $T$.}, we can then also consider the sub-policies of the agent related to the collective reward. $\pi_c^i$ is the sub policy for the collective reward which can only be returned at the terminal state so
$\pi^i_c:=\pi^i_T(a^i_T|o^i_T)|\hat{\mathcal{R}}_T(a^i_T,o^i_T)=r^c$. $\pi_d$ is the sub policy for the individual reward where $\pi^i_d:=\pi^i_t(a^i_t|o^i_t)|\hat{\mathcal{R}}_t(a^i_t,o^i_t)=r_t^i$. Notably, both these policies output a scalar probability for an action $a^i_t$ where $o^i_t$ is the observation where the sub policy terminates. 

Since the collective reward $r_T^c$ only depends on the other agent $j$'s policy after agent $i$ acquired $r^i_t$, we need to consider agent $j$'s policy in the calculation of the gradient of the objective. But, agent $j$'s policy is not a stationary distribution. Therefore, the collective objective from agent $i$'s perspective must include an estimate of how agent $j$'s policy may change. 

Of course, agent $i$ cannot exactly know how agent $j$'s policy will change, but they can assume that agent $j$ is another agent learning through policy gradients. Any agent learning via a policy gradient will update their policy proportional to the log of their policy. Therefore, to obtain a better estimate of the collective reward that accounts for the non-stationarity in agent $j$'s policy, we will consider an \emph{estimate} of the collective objective with an appropriate weighting based on agent $j$'s log policy $\hat{J_c}(\Theta^i)$ and differentiate with respect to agent $j$'s parameters:

\begin{equation}\label{eq:3}
\nabla_{\Theta^i} \hat{J_c}(\Theta^i)=\mathbb{E}_{\tau\sim\pi^j}\big[\nabla_{\Theta^i}\log \pi_c^j(a^j|o^j)\,
      r^c_T\big] = \mathbb{E}_{\tau\sim\pi^j}\big[\log \pi_c^j(a^j|o^j)\,
      \nabla_{\Theta^i}r^c_T\big] 
\end{equation}

Passing the gradient with respect to agent $j$'s policy through agent $i$'s objective is usually $0$ but since agent $j$'s value function is predicting the collective reward, it's values change with agent $j$'s policy. The collective policy of agent $i$ is statistically independent from effect of agent $j$'s collective policy:

\begin{equation}\label{eq:3}
\begin{split}
\nabla_{\Theta^i} \hat{J_c}(\Theta^i)
&= \mathbb{E}_{\tau\sim\pi^j}\big[\log \pi_c^j(a^j|o^j)\big]\,
   \mathbb{E}_{\tau\sim\pi^j}\big[\nabla_{\Theta^i}r^c_T\big]
\end{split}
\end{equation}

Since the collective objective is a scalar quantity, we can get a surrogate for an agent's gradient of the collective reward $\mathbb{E}_{\tau\sim  \pi^j}\big[\nabla_{\Theta^j}r^c_T\big]$ by dividing the terms by $\mathbb{E}_{\tau\sim\pi^j}\big[\log \pi_c^i(a^i|o^i)\big]$ which is a scalar quantity and equivalent to the entropy of the policy at the terminal state $\mathbb{E}_{\tau\sim\pi^j}\big[\log \pi_c^j(a^j|o^j)\big] = \sum_{a^j}^{|\mathcal{A}|}\pi_c^j(a^j|o^j)\log \pi_c^j(a^j|o^j)=\mathbb{H}\big[\pi_c^j(.|o^j)\big]$. Notably, the policy at the terminal state is unaffected by the reward. Per learning aware methods \citep{foerster2017learning}, the policy entropy between agents is swapped so agents are aware of each other's predictions. The gradient of the collective reward prediction is inversely related to the entropy of the other agent's policy.

\begin{theorem}
\label{thm:correction}
Let $r^c$ be the collective reward returned for agent $i$ at the terminal state. By \cref{eq:3} with the inverse of the first term, the correction term is

\begin{equation}\label{eq:4}
\nabla_{\Theta^i}r^c= \mathbb{E}_{\tau\sim\pi^j}[\frac{\nabla_{\Theta^i}\hat{J_c}(\Theta^i)}{\log \pi_c^j(a^j|o^j)}]  = \mathbb{E}_{\tau\sim\pi^j}[\nabla_{\Theta^i}\hat{J_c}(\Theta^i)\Psi(\pi_c^j,a^j,o^j)]= \nabla_{\Theta^i}J_c(\Theta^i)
\end{equation}

where the scalar reciprocal is the other agent's expected score function $\Psi(\pi_c^j,a^j,o^j) = \frac{1}{\mathbb{E}_{\tau\sim\pi^j}[\log \pi_c^j(a^j|o^j)]}$ and $i\not =j$. Since the collective reward is a scalar and the correction term corrects for the change in the $r^c$ which is also a scalar.
\end{theorem}

See  P.\ref{p:correction} for the full proof. As a sketch, we rely on two key assumptions. The first key assumption is the episode was successful so by the reward function, the collective reward was returned in the episode. As a result, there was a sequence of time steps where  agent $j$ was not holding the key up until the collective reward was returned. The second key assumption is that the other agent's collective reward policy is differentiable. With those assumptions we can then use the objective of agent $j$ as a surrogate for the collective reward in the look-ahead step of the policy gradient derivation \citep{sutton1998reinforcement}, similar to mutual learning aware update rules \citep{willi2022cola,foerster2017learning}. The correction term does not conflict with individual objectives (see P.\ref{p:conf}) and can be computed with a finite difference method. The corrected gradient objective for agent $j$ from P.\ref{p:correction} is:

\begin{equation}\label{eq:5}
\nabla_{\Theta^j}J(\Theta^{j}) = \mathbb{E}_{\tau^i\sim\pi^i,\tau^j\sim\pi^j}[\nabla_{\Theta^j}\log{\pi^j(a^j|o^j)Q(o^j,a^j)}+ \nabla_{\Theta^j}\nabla_{\Theta^i}J_c(\Theta^i)\Psi(\pi_c^j,a^j,o^j)]
\end{equation}

\subsection{Use of a correction term in the policy gradient}

Correcting the policy gradient for how the collective reward changes with the other agent's policy \cref{eq:5}  should reduce the variance in success by stabilizing their value estimate with respect to each other's policies updating. Furthermore, since the collective reward is shared and the optimal policy is the same between both agents $\pi^{i*} = \pi^{j*}$, it is reasonable to further decentralize the correction term by only using an agent's own policy and parameters: $ \mathbb{E}_{\tau\sim\pi^i}[\nabla_{\Theta^i}\hat{J_c}(\Theta^i)\Psi(\pi_c^i,a^i,o^i)]$  which we call ``Self Correction''. Hence, we evaluated these policy gradient correction terms Fig.\ref{fig:6}.

\begin{figure}[!htbp]
    \centering

    \caption*{\textbf{Improved success rate and variance with the derived correction term}}

    \begin{subfigure}[b]{0.29\textwidth}
        \centering
        \includegraphics[width=4.cm]{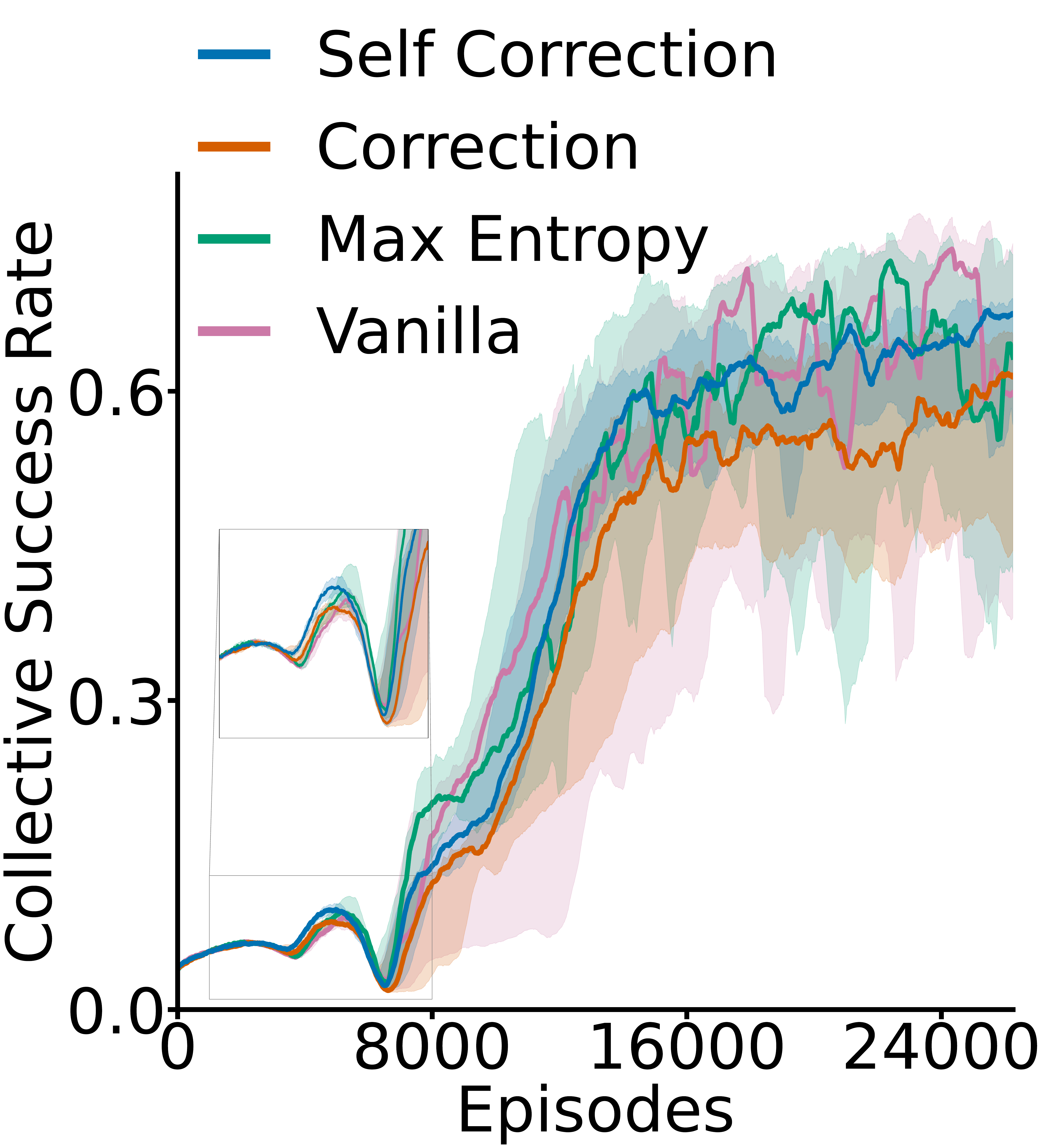}
        \caption{Collective Success Rate}
        \label{fig:5a}
    \end{subfigure}
        \begin{subfigure}[b]{0.29\textwidth}
        \centering
        \includegraphics[width=4.cm]{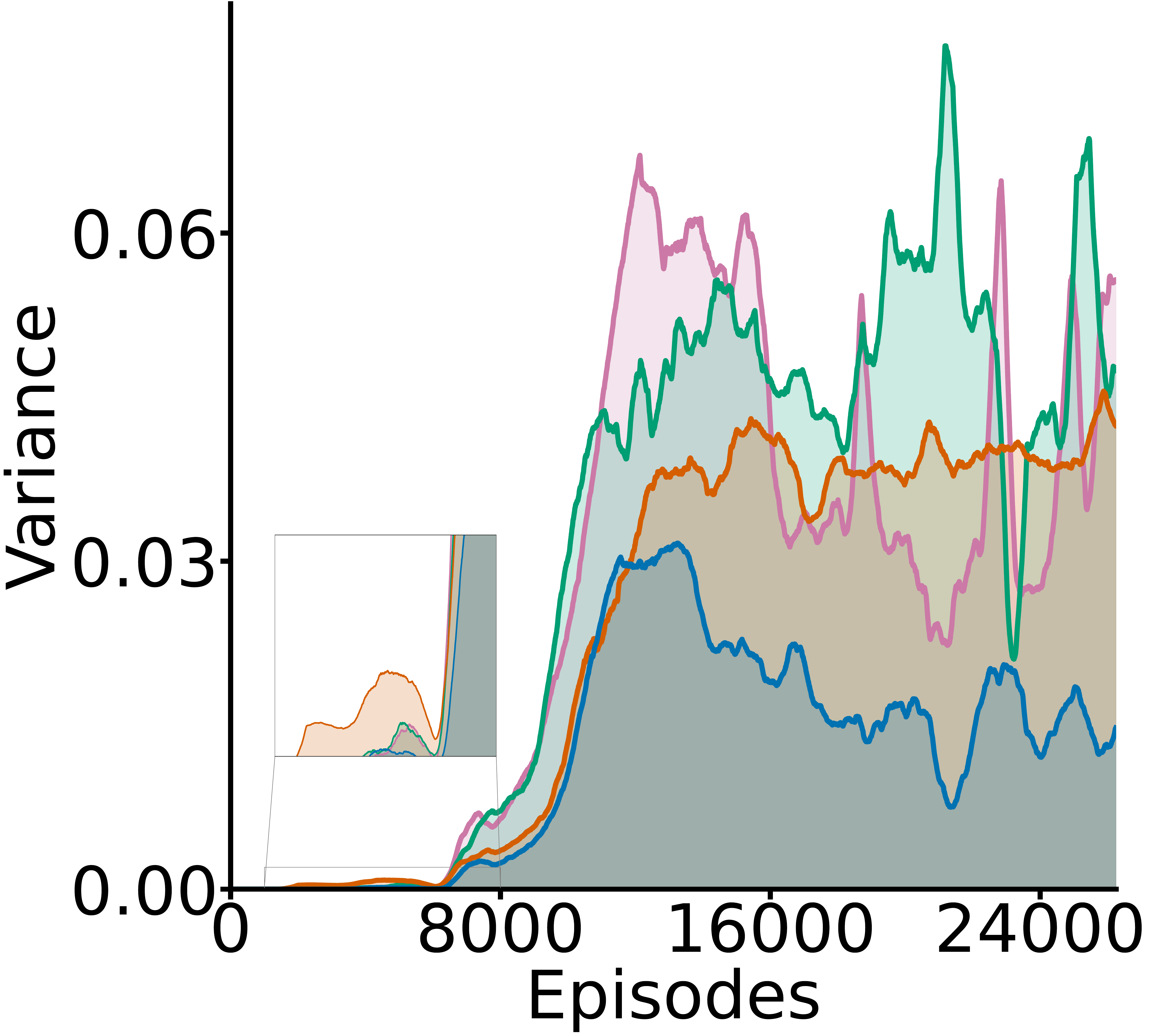}
        \caption{Variance}
        \label{fig:5d}
    \end{subfigure}
    \begin{subfigure}[b]{0.32\textwidth}
        \centering
        \includegraphics[width=4.cm]{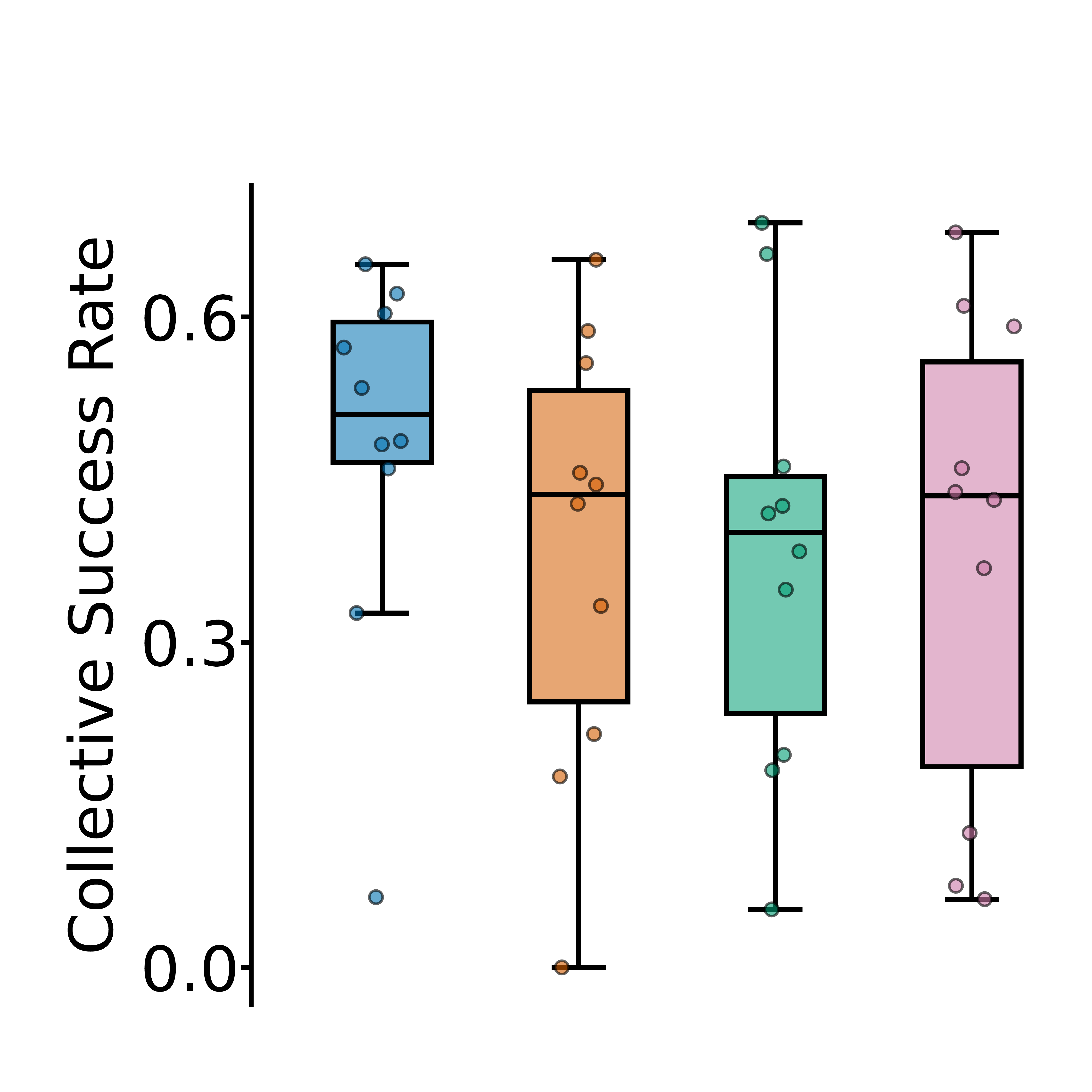}
        \caption{Global Collective Success }
        \label{fig:5c}
    \end{subfigure}

    \caption{Collective success rate of PG agents comparing the vanilla model against one with a maximum entropy term, with the correction term, and the self-correction term. (b) Variance in collective success rate across episodes.  (c)   Global collective success rate of comparing PG agents with, maximum entropy, the correction PG agents}
    \label{fig:6}
\end{figure}

With action history inputs, we trained PG agents with and without the correction and self-correction terms over seven days to ensure convergence. Additionally, we examined PG agents with a maximum entropy term, which should also reduce the variance in the learned policies \citep{ahmed2019understanding,haarnoja2018soft, eysenbach2022maximum}. We found that all of the agents converged to a fairly high success rate over time (Fig. \ref{fig:6a}), high cumulative reward (Fig. \ref{fig:6b}) and reduced the distance between themselves (Fig.\ref{fig:behav2}b). Notably, the collective success variance was markedly different. The variance of the standard PG agents was quite high with steep spikes, and the variance of the max-entropy agents were not any different throughout the majority of the episodes, with the exception of the very early episodes (Fig. \ref{fig:6c}). In contrast, the variance of the agents with the correction term was a bit lower but more stable. Interestingly, the agents with the self-correction term showed the lowest variance. We believe that this may be due to added noise from considering multiple policies in the update. Altogether, these results show that the correction term reduces variance in performance in the hidden gift problem, but is more prominent when decentralized with self-correction. This is interesting because it shows that it may be possible to resolve the complexities of hidden gift credit assignment using self learning-awareness rather than collective learning-awareness.  




\subsection{Self correction outperforms LOLA under reward factorization assumptions}

Self correction was inspired by learning awareness, namely how an agent learns can affect reward acquisition and ultimately reciprocal collective behavior. a key advantage of self correction is that it does not require access to the other agent's policy to compute the correction term and is a scalar surrogate for how a policy conditions shared rewards nor requires an additional learning rate to . However, it does require \emph{a priori} knowledge that the reward can be factorized into a shared collective component and individual components. In contrast, the original learning-aware correction, LOLA, explicitly uses the other agent's policy in the correction term over the entire returned trajectory but does not rely on reward factorization \citep{foerster2017learning, willi2022cola}. Under this comparison, one might expect LOLA to outperform self correction. To test this, we implemented and tuned two variants of LOLA.

\begin{figure}[!htbp]
    \centering


    \begin{subfigure}[b]{0.29\textwidth}
        \centering
        \includegraphics[width=4.cm]{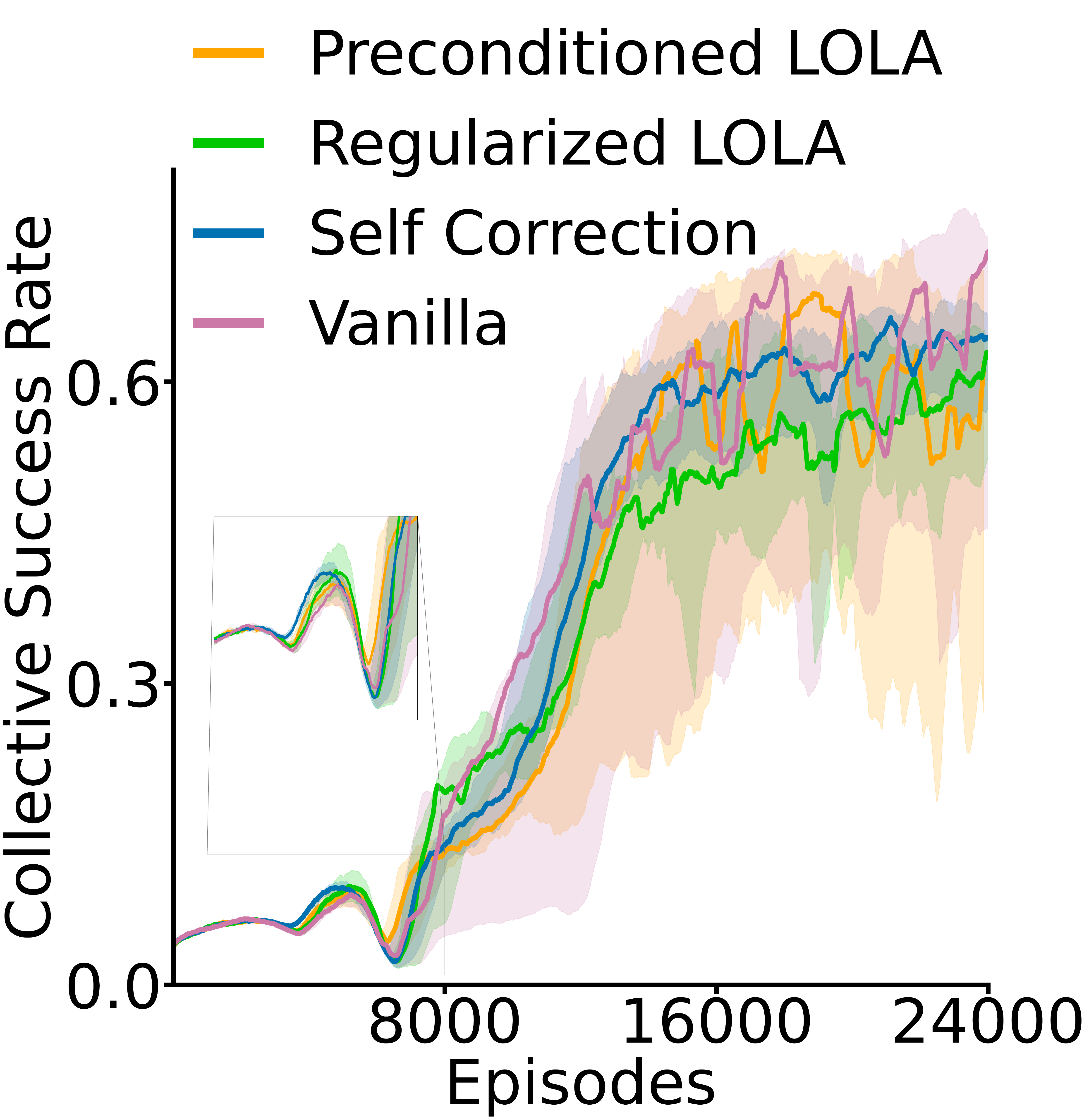}
        \caption{Collective Success Rate}
        \label{fig:6a}
    \end{subfigure}
        \begin{subfigure}[b]{0.29\textwidth}
        \centering
        \includegraphics[width=4.cm]{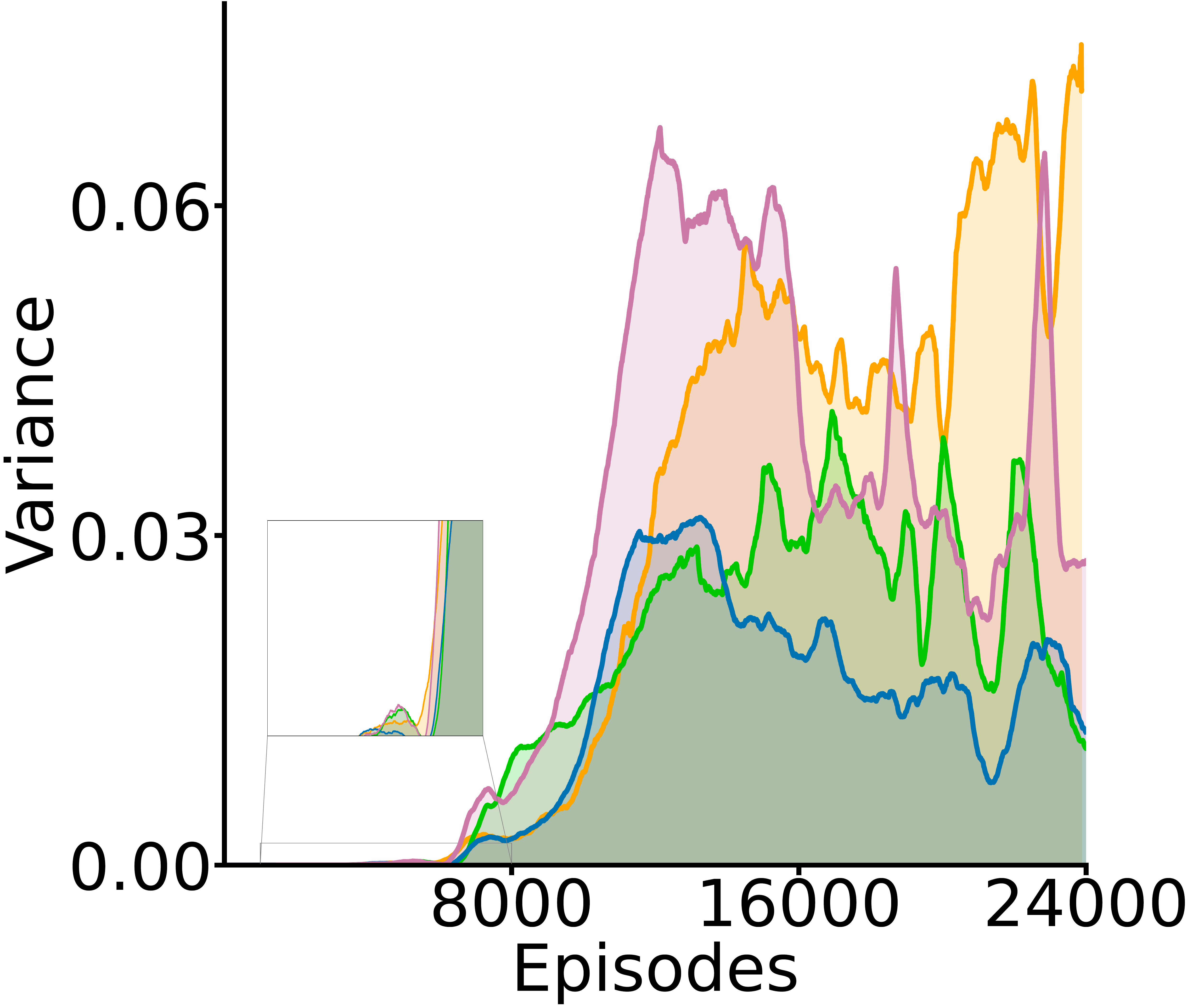}
        \caption{Variance}
        \label{fig:6b}
    \end{subfigure}
    \begin{subfigure}[b]{0.33\textwidth}
        \centering
        \includegraphics[width=4.cm]{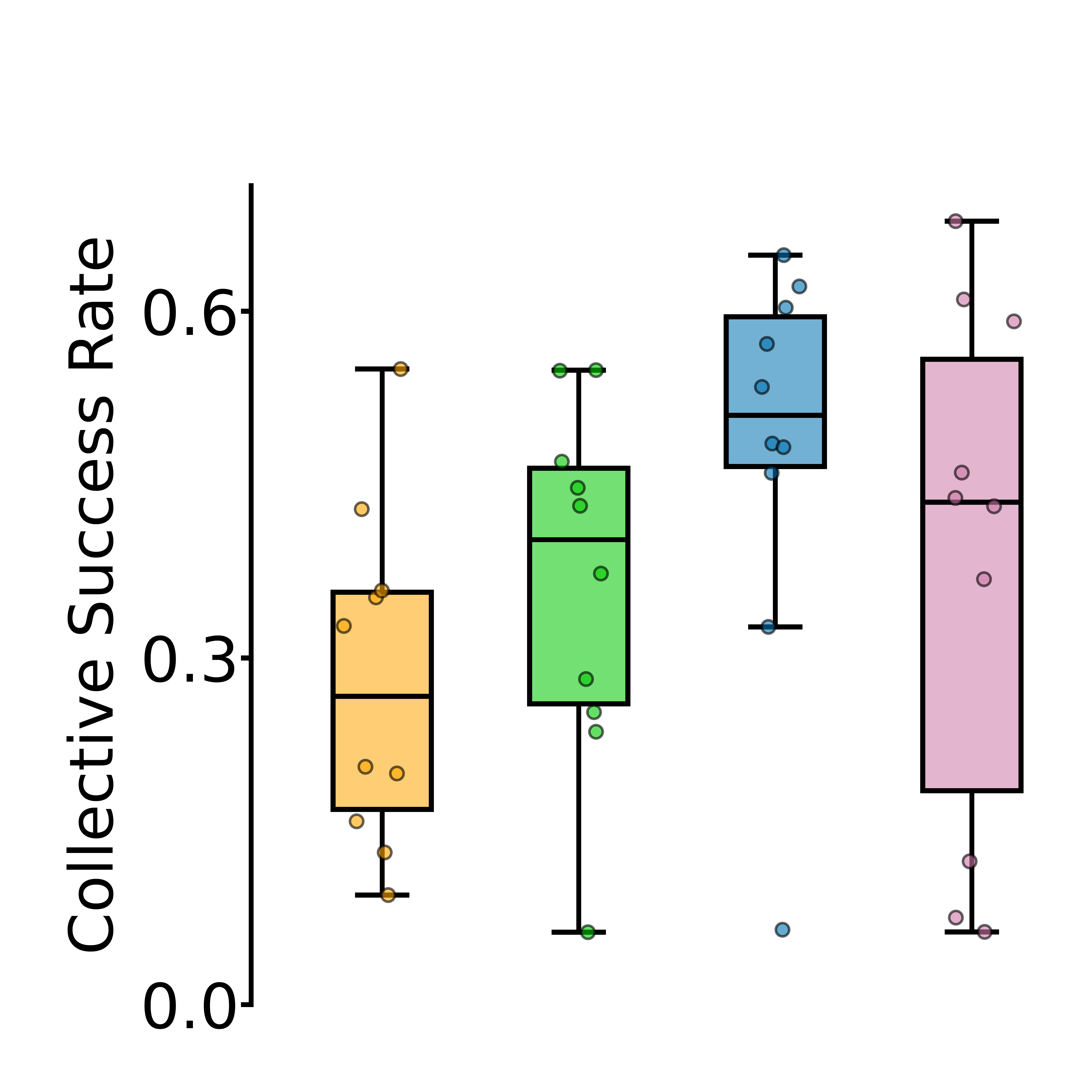}
        \caption{Global Collective Success }
        \label{fig:6c}
    \end{subfigure}

    \caption{Collective success rate of PG agents comparing the vanilla model against one with a self correction term, with a preconditioned LOLA term, and a regularizing LOLA term. (b) Variance in collective success rate across episodes.  (c) Global collective success rate of comparing vanilla, LOLA, and self correction PG agents }
    \label{fig:6}
\end{figure}

The first variant follows the original LOLA update and preconditions parameters to approximate a lookahead step through the other agent's learning. The best learning rate for this variant was $2\times10^{-5}$. The second variant uses the LOLA term as a regularizer added to the gradient update, with a best learning rate of $5\times10^{-4}$. In our implementation, preconditioned LOLA was also more computationally expensive than regularized LOLA.

In 7 days, both LOLA variants reached median collective success rates comparable to the self-correction agent, but their learning dynamics were far less stable. Both LOLA agents exhibited higher variance than the self-correction agent (Fig. \ref{fig:6b}). Regularized LOLA reduced variance substantially relative to vanilla PG, but instability persisted. This instability negatively impacted global collective success, where the median performance of LOLA agents fell below that of vanilla PG (Fig. \ref{fig:6c}).

Overall, these results indicate that LOLA can solve the Manitokan task without assuming reward factorization, but behaved similarly to vanilla policy gradients in terms of stability and aggregate performance. In contrast, self correction achieves more reliable learning and stronger global collective success under the reward factorization assumption.

\section{Discussion} \label{sec:disc}
In this work we developed a MARL task to explore the complexities of learning in the presence of ``hidden gifts'', i.e. cooperative acts that are not revealed to the recipient. The Manitokan task we developed, inspired by the concept in Indigenous plains communities across North America, requires agents to open doors using a single shared key in the environment. Agents must drop the key for other agents after they have used it if they are to obtain a larger collective reward. But, these key drop acts are not apparent to the other agents, making it difficult to assign credit between policy updates.

We observed that in the basic version of the Manitokan task none of the algorithms tested were able to solve it. This included both  policy gradient agents (PG, PPO), meta-learning agents (SAF), enhanced exploration agents (MAVEN), counterfactual agents (COMA), and agents with collective value functions (VDN, QMIX, QTRAN, and MAPPO). When we added additional information to the observations the more sophisticated algorithms tested were still not able to solve this task. However, with previous action information,  the actor-critic PG agents could solve the task, though with high variance. Formal analysis of the value function for the Manitokan task showed that it contains a second-order term related to the collective reward that can reduce instability in learning. We used this to derive a gradient correction for the PG agents that successfully reduced the variance in their performance and improved the collective success rate.  Altogether, our results demonstrate that hidden gifts introduce challenging credit assignment problems that many state-of-the-art MARL architectures were not designed to overcome.

\subsection{Limitations} \label{sec:limit}
We used a grid-world task to induce the hidden gift credit assignment problem while enabling a tractable formal analysis. However, real-world settings are rich in sensory input, and biological agents typically operate over much larger action spaces. Denser observations and less sparse reward signals may make credit assignment easier by providing more information to leverage or infer, although they may also introduce additional confounds that are absent in simplified environments.

Additionally, biological agents can engage in explicit, structured inter-agent communication, which may aid coordination by enabling agents to negotiate objectives, roles, or plans \citep{wu2024group}. This differs from SAF's latent communication protocol \citep{liu2023stateful}, in which coordination must be expressed implicitly through internal state rather than explicit commitments. In settings where agents can communicate, it may be easier to establish gifting commitments beforehand, and such commitments may later become implicit norms that require less overt communication \citep{velez2022representational}.

Further, the limited memory capacity of the GRU architecture may inhibit credit assignment over longer horizons. Incorporating more explicit memory alongside action history (e.g., a long-context transformer) could enable agents to better attribute outcomes to prior gifting behavior \citep{ni2023transformers, chen2021decision, cross2025hypothetical}. A retrieval-augmented temporal memory mechanism \citep{hung2019optimizing} might also help model-free agents avoid learning policies that drift away from, or discount, collective objectives when the relevant dependency structure is non-Markovian \citep{NEURIPS2023_08342dc6}. This temporal retrieval mechanism is distinct from SAF's spatial mechanism \citep{liu2023stateful}.

Finally, PPO-based agents (MAPPO, IPPO, SAF) were ineffective at acquiring the collective reward, and this persisted even after augmentation with action history. While the cause is not fully resolved here, the result raises the possibility that PPO’s clipped surrogate objective may require additional bias-correction mechanisms, similar to correction term used in PG agents (see Section \ref{sec:formal}). Given related observations that PPO can struggle with non-stationary rewards in continual learning tasks, it may also be fruitful to treat policy adaptation under non-stationarity as a continual learning problem and leverage mechanisms from continual RL \citep{dohare2024loss, Elelimy2023ContinualRL, khetarpal2020towards}.


\subsection{Rethinking reciprocity}\label{sec:rep}

A broader implication of our work is that reciprocity in multi-agent settings becomes more complex when reciprocal actions are partially or fully unobservable. In such cases, reciprocity is temporally indirect and agents cannot easily determine whether cooperative behavior will persist \citep{nowak2005evolution, santos2021complexity}. One way to address this challenge may be to improve agents' ability to either predict the future actions of other agents or influence them through implicit information \citep{jaques2019social, xie2021learning}. Both mechanisms could reduce uncertainty about whether altruistic gifts will be exploited and may therefore support reciprocity in MARL environments where gifting behavior is hidden or asymmetric.

More generally, reciprocity under hidden information may benefit from agents that can accurately model the behavior of others when observations are incomplete. The correction term derived in our formal analysis was motivated by the gradient-steering effects studied in several learning-aware approaches \citep{willi2022cola, foerster2017learning, meulemans2025multiagentcooperationlearningawarepolicy, aghajohari2024loqa}. Although learning-aware methods are not always optimal (see E.~\ref{e:lola_fail}), these results suggest that some of their underlying principles may have broader applicability beyond the domains for which they were originally designed. For example, similar mechanisms might be used not only to encourage cooperation but also to inhibit undesirable coordination, such as harmful cooperation or collusion when the reward structure incentivizes it (see E. \ref{e:safe}).

\section{Acknowledgments}

DM would like to thank Vedant Shah for the experimental ideas for further validating the results, Michael Noukhovitch for the insightful perspectives on state-information experiments and profound MARL knowledge and Arna Ghosh for verifying and questioning the correction term derivation as well has Hessian approximations.

BAR received support from NSERC (Discovery Grant: RGPIN-2020-05105; Discovery Accelerator Supplement: RGPAS-2020-00031) and
CIFAR (Canada AI Chair; Learning in Machine and Brains Fellowship). DM received support from an
NSERC CGSM and a Rathlyn Fellowship from the Indigenous Studies Department of McGill. This research was enabled in part by support provided by (Calcul Québec) (https://www.calculquebec.ca/en/)
9and the Digital Research Alliance of Canada (https://alliancecan.ca/en). The authors acknowledge
the material support of NVIDIA in the form of computational resources.

\bibliographystyle{rlj}

\newpage
\beginSupplementaryMaterials
\label{supp}


\rsection{}\label{related}

\rsubsection{Coordination and Gifting in MARL}\label{coop}
Fully cooperative coordination games feature a single team objective requiring agents to act jointly, often reducible to a single-agent problem with a large action space. Previous tasks include navigation \citep{mordatch2017emergence, lowe2017multi}, cooking coordination \citep{carroll2019utility, gessler2025overcookedv2}, battles \citep{samvelyan2019starcraft2, ellis2023smacv2}, and social-dilemmas \citep{leibo2017multi, lerer2017maintaining, christianos2020shared}. These are often studied under the centralized training with decentralized execution, with methods such as COMA \citep{foerster2018counterfactual} and QMIX \citep{rashid2020monotonic} leveraging global states during training to stabilize coordination. Additionally sharing collective rewards across agents is common and promotes cooperation but can also create “lazy-agent” credit assignment behavior \citep{liu2023lazy}. Individualized rewards can mitigate this but risk pulling policies away from team objectives \citep{wang2022individual}.

Within this cooperative context, “gifting” has been proposed as a mechanism for reward transfer, where one agent deliberately allocates part of its payoff to another to foster cooperation or reciprocity \citep{hughes2018inequity, peysakhovich2018prosocial, lupu2020gifting}. This can be seen as a bounded, targeted form of social influence. In single-agent RL this gifting can be interpreted as an intrinsic “self-gift,” i.e., intrinsically generated rewards that support exploration or long-horizon credit assignment \citep{schmidhuber1991possibility, arjona2019rudder, sun2023contrastive}. In multi-agent settings, intrinsic rewards have also been used to shape others’ behavior through causal social influence \citep{jaques2019social}. However, this gifting is treated only as scalar reward signals, not as the transfer of tangible, task-critical resources.
\rsubsection{Multi-Objective RL}
Many decision-making problems involve objectives whose relative importance shifts over time, creating a non-stationary optimization landscape where fixed-weight multi-objective RL (MORL) methods falter \citep{van2014multi,roijers2013survey}. Dynamic-weights MORL addresses this by conditioning policies or value functions on the current weight vector $w(t)$, enabling a single policy to adapt across changing trade-offs without retraining. Approaches include weight-conditioned DQNs \citep{mossalam2016multi}, policy gradients with weight inputs \citep{abels2019dynamic}, and replay strategies for stability under shifting scalarizations \citep{yang2019generalized}.

In multi-agent settings, MORL has been used to balance individual and collective goals \citep{hayes2022practical}, but prior work assumes known or designed $w(t)$, rather than treating another agent’s policy itself as a dynamic weight. Seldom in the world do we have ever complete control of our incentives.

\newpage
\msection{Methods}
 \label{meth}
This section contains the hyperparameters for the results, hardware details for training and minor details on the task setup. 

\msubsection{Hyperparameters} \label{m:hyp}
\begin{table}[!htbp]
\centering
\caption{Model architecture and hyperparameters used for MAPPO.}
\begin{tabularx}{\textwidth}{l X}
\toprule
\textbf{Component} & \textbf{Specification} \\
\midrule
Policy Network Architecture (Joint) & 1-layer CNN (outchannels = 32, kernel = 3, ReLU), 1-layer MLP (input = 32, output=64, ReLU), 1 layer MLP (input = 32, output=64, ReLU), 1 layer MLP (input = 64, output=64, ReLU), 1 layer GRU (input = 64, output = 64, with LayerNorm), 1 layer Categorical (input=64, output=6)  \\
Value Network Architecture (Joint) &  1-layer CNN (outchannels = 32, kernel = 3, ReLU), 1-layer MLP (input = 32, output=64, ReLU), 1 layer MLP (input = 32, output=64, ReLU), 1 layer MLP (input = 64, output=64, ReLU), 1 layer GRU (input = 64, output = 64, with LayerNorm), 1 layer MLP(input = 64, output = 1, ReLU) \\
Optimizer & Adam, learning rate: $1 \times 10^{-5}$ \\
Discount Factor $\gamma$ & 0.99 \\
GAE Parameter $\lambda$ & 0.95 \\
PPO Clip Ratio $\epsilon$ & 0.2 \\
Entropy Coefficient & 0.0001 \\
Data chunk length & 10\\
Parallel Environments & 32   \\
Batch Size & Parallel Environments × Data chunk length  × number of agents \\
Mini-batch Size & 1 \\
Epochs per Update & 15 \\
Gradient Clipping & 10 \\
Value Function Coef. & 1 \\
Gain & 0.01 \\
Loss & Huber Loss with delta 10.00 \\
\bottomrule
\end{tabularx}
\label{tab:mappo_hyperparams}
\end{table}
\FloatBarrier
\begin{table}[!htbp]
\centering
\caption{Model architecture and hyperparameters used for IPPO.}
\begin{tabularx}{\textwidth}{l X}
\toprule
\textbf{Component} & \textbf{Specification} \\
\midrule
Policy Network Architecture (Disjoint) & 1-layer CNN (outchannels = 32, kernel = 3, ReLU), 1-layer MLP (input = 32, output=64, ReLU), 1 layer MLP (input = 32, output=64, ReLU), 1 layer MLP (input = 64, output=64, ReLU), 1 layer GRU (input = 64, output = 64, with LayerNorm), 1 layer Categorical (input=64, output=6)  \\
Value Network Architecture (Disjoint) &  1-layer CNN (outchannels = 32, kernel = 3, ReLU), 1-layer MLP (input = 32, output=64, ReLU), 1 layer MLP (input = 32, output=64, ReLU), 1 layer MLP (input = 64, output=64, ReLU), 1 layer GRU (input = 64, output = 64, with LayerNorm), 1 layer MLP(input = 64, output = 1, ReLU) \\
Optimizer & Adam \\
Learning rate & $1 \times 10^{-5}$ \\
Discount Factor $\gamma$ & 0.99 \\
GAE Parameter $\lambda$ & Not used \\
PPO Clip Ratio $\epsilon$ & 0.2 \\
Entropy Coefficient & 0.0001 \\
Data chunk length & 10\\
Parallel Environments & 32   \\
Batch Size & Parallel Environments × Data chunk length  × number of agents \\
Mini-batch Size & 1 \\
Epochs per Update & 15 \\
Gradient Clipping & 10 \\
Value Function Coef. & 1 \\
Gain & 0.01 \\
Loss & Huber Loss with delta 10.00 \\
\bottomrule
\end{tabularx}
\label{tab:ippo_hyperparams}
\end{table}
\FloatBarrier
\begin{table}[!htbp]
\centering
\caption{Model architecture and hyperparameters used for PG.}
\begin{tabularx}{\textwidth}{l X}
\toprule
\textbf{Component} & \textbf{Specification} \\
\midrule
Policy Network Architecture (Joint) &  1-layer MLP (input = 27, output=64, ReLU), 1 layer GRU (input = 64, output = 64), 1 layer MLP (input=64, output=6)  \\
Critic Network Architecture (Disjoint) &  1-layer MLP (input = 27, output=64, ReLU), 1-layer MLP (input = 64, output=64, ReLU), 1-layer MLP (input=64, output=1) \\

Target Critic Network Architecture (Disjoint) &  1-layer MLP (input = 27, output=64, ReLU), 1-layer MLP (input = 64, output=64, ReLU), 1-layer MLP (input=64, output=1) \\

Actor optimizer & RMSprop, alpha $0.99$, epsilon $1\times10^{-5}$ \\
Critic optimizer & RMSprop, alpha $0.99$, epsilon $1\times10^{-5}$ \\
Discount factor $\gamma$ & 0.99 \\
Target network update interval & 1 episode\\
Learning rate & $5 \times 10^{-5}$ \\
TD Lambda & 1.0 \\
Replay buffer size & 32 \\
Parallel environment & 32 \\
Parallel episodes per buffer episode & 1 \\
Training batch size & 32 \\

\bottomrule
\end{tabularx}
\label{tab:pg_hyperparams}
\end{table}
\FloatBarrier
\begin{table}[!htbp]
\centering
\caption{Model architecture and hyperparameters used for COMA.}
\begin{tabularx}{\textwidth}{l X}
\toprule
\textbf{Component} & \textbf{Specification} \\
\midrule
Policy Network Architecture (Joint) &  1-layer MLP (input = 27, output=64, ReLU), 1 layer GRU (input = 64, output = 64), 1 layer MLP (input=64, output=6)  \\
Critic Network Architecture (Joint) &  1-layer MLP (input = 27, output=64, ReLU), 1-layer MLP (input = 64, output=64, ReLU), 1-layer MLP (input=64, output=6) \\

Target Critic Network Architecture (Joint) &  1-layer MLP (input = 27, output=64, ReLU), 1-layer MLP (input = 64, output=64, ReLU), 1-layer MLP (input=64, output=6) \\

Actor optimizer & RMSprop, alpha $0.99$, epsilon $1\times10^{-5}$ \\
Critic optimizer & RMSprop, alpha $0.99$, epsilon $1\times10^{-5}$ \\
Discount factor $\gamma$ & 0.99 \\
Target network update interval & 1 episode\\
Learning rate & $5 \times 10^{-5}$ \\
TD Lambda & 1.0 \\
Replay buffer size & 320 \\
Parallel environment & 32 \\
Parallel episodes per buffer episode & 1 \\
Training batch size & 32 \\
\bottomrule
\end{tabularx}
\label{tab:coma_hyperparams}
\end{table}
\FloatBarrier
\begin{table}[!htbp]
\centering
\caption{Model architecture and hyperparameters used for SAF.}
\begin{tabularx}{\textwidth}{l X}
\toprule
\textbf{Component} & \textbf{Specification} \\
\midrule
Policy Network Architecture (Disjoint) & 2-layer MLP (input = 64, output=128, Tanh),  \\
Value Network Architecture (Joint) &  2-layer MLP (input = 80, output=128, Tanh), \\
Shared Convolutional Encoder (Joint) & 1-Layer CNN (outchannels = 64, kernel = 2)  \\
\multicolumn{2}{l}{Knowledge Source Architecture (Joint)} \\
\quad Query Projector &  1-layer MLP (input = 128, output=64, Tanh) \\
\quad State Projector &  1-layer MLP (input = 128, output=64, Tanh)\\
\quad Perceiver Encoder& (latents = 4, latent input = 64, cross attention channels = 64, cross attention heads = 1, self attention heads = 1, self attention blocks = 2 with 2 layers each)    \\
\quad  Cross Attention & (heads = 1, query input = 64, key-value input = 64, query-key input = 64, value channels = 64, dropout = 0.0) \\
\quad Combined State Projector & 1-layer MLP (input = 128, output=64, Tanh) \\
\quad Latent Encoder & 1-layer MLP (input = 128, output=64, Tanh), 1-layer MLP (input = 64, output=64, Tanh ),1-layer MLP (input = 64, output=16, Tanh ) \\
\quad Latent Encoder Prior & 1-layer MLP (input = 64, output=64, Tanh), 1-layer MLP (input = 64, output=64, Tanh ),1-layer MLP (input = 64, output=16, Tanh ) \\
\quad Policy Projector & 1-layer MLP (input = 128, output=164, Tanh) \\

Optimizer & Adam, epsilon $1\times 10^{-5}$ \\
learning rate & $3 \times 10^{-4}$ \\
Discount Factor $\gamma$ & 0.99 \\
GAE Parameter $\lambda$ & GAE not used \\
PPO Clip Ratio $\epsilon$ & 0.2 \\
Entropy Coefficient & 0.01 \\
Data chunk length & 10\\
Parallel Environments & 32   \\
Batch Size & Parallel Environments × Data chunk length  × number of agents \\
Mini-batch Size & 5 \\
Epochs per Update & 15 \\
Gradient Clipping & 9 \\
Value Function Coef. & 1 \\
Gain & 0.01 \\
Loss & Huber Loss with delta 10.00 \\
Number of policies & 4 \\
Number of slot keys & 4 \\
\bottomrule
\end{tabularx}
\label{tab:saf_hyperparams}
\end{table}
\FloatBarrier
\begin{table}[!htbp]
\centering
\caption{Model architecture and hyperparameters used for VDN.}
\begin{tabularx}{\textwidth}{l X}
\toprule
\textbf{Component} & \textbf{Specification} \\
\midrule
Policy Network Architecture (Joint) &  1-layer MLP (input = 27, output=64, ReLU), 1 layer GRU (input = 128, output = 64), 1 layer MLP (input=128, output=6)  \\
Target Policy Network Architecture (Joint) &  1-layer MLP (input = 27, output=64, ReLU), 1 layer GRU (input = 128, output = 64), 1 layer MLP (input=128, output=6)  \\
Mixer Network Architecture &  Tensor sum of states\\
Policy optimizer & Adam, alpha $0.99$, epsilon $1\times10^{-5}$ \\
Target policy optimizer & Adam, alpha $0.99$, epsilon $1\times10^{-5}$ \\
Start epsilon greedy & 1.0 \\
Minimum epsilon greedy &  0.05 \\
Discount factor $\gamma$ & 0.99 \\
Target network update interval & 1 episode\\
Start learning rate & $1 \times 10^{-2}$ \\
Minimum learning rate & $1 \times 10^{-6}$ \\
TD Lambda & 1.0 \\
Replay buffer size & 1000 \\
Parallel environment & 32 \\
Parallel episodes per buffer episode & 32 \\
Training batch size & 32 \\
Warm up buffer episodes & 32 \\
\bottomrule
\end{tabularx}
\label{tab:vdn_hyperparams}
\end{table}
\FloatBarrier
\begin{table}[!htbp]
\centering
\caption{Model architecture and hyperparameters used for QMIX.}
\begin{tabularx}{\textwidth}{l X}
\toprule
\textbf{Component} & \textbf{Specification} \\
\midrule
Actor Network Architecture (Joint) &  1-layer MLP (input = 27, output=64, ReLU), 1 layer GRU (input = 64, output = 64), 1 layer MLP (input=64, output=6)  \\
Target Actor Network Architecture (Joint) &  1-layer MLP (input = 27, output=64, ReLU), 1 layer GRU (input = 64, output = 64), 1 layer MLP (input=64, output=6)  \\

\multicolumn{2}{l}{Mixing Network Architecture (Joint)} \\
\quad Hypernet Weights 1 &  1-layer MLP (input =54, output=64, ReLU),  1-layer MLP (input = 64, output=52) \\
\quad Hypernet Biases 1 &  1-layer MLP (input =54, output=64) \\
\quad Hypernet Weights 2 &  1-layer MLP (input =54, output=32, ReLU),  1-layer MLP (input = 64, output=32) \\
\quad Hypernet Bias 2 &  1-layer MLP (input =54, output=64, ReLU),  1-layer MLP (input = 64, output=1) \\

\multicolumn{2}{l}{Target Mixing Network Architecture (Joint)} \\
\quad Hypernet Weights 1 &  1-layer MLP (input =54, output=64, ReLU),  1-layer MLP (input = 64, output=52) \\
\quad Hypernet Biases 1 &  1-layer MLP (input =54, output=64) \\
\quad Hypernet Weights 2 &  1-layer MLP (input =54, output=32, ReLU),  1-layer MLP (input = 64, output=32) \\
\quad Hypernet Bias 2 &  1-layer MLP (input =54, output=64, ReLU),  1-layer MLP (input = 64, output=1) \\

Policy optimizer & Adam, alpha $0.99$, epsilon $1\times10^{-5}$ \\
Target policy optimizer & Adam, alpha $0.99$, epsilon $1\times10^{-5}$ \\
Start epsilon greedy & 1.0 \\
Minimum epsilon greedy &  0.05 \\
Discount factor $\gamma$ & 0.99 \\
Target network update interval & 1 episode\\
Start learning rate & $1 \times 10^{-2}$ \\
Minimum learning rate & $1 \times 10^{-6}$ \\
TD Lambda & 1.0 \\
Replay buffer size & 1000 \\
Parallel environment & 32 \\
Parallel episodes per buffer episode & 32 \\
Training batch size & 32 \\
Warm up buffer episodes & 32 \\
\bottomrule
\end{tabularx}
\label{tab:qmix_hyperparams}
\end{table}
\FloatBarrier
\begin{table}[!htbp]
\centering
\caption{Model architecture and hyperparameters used for QTRAN.}
\begin{tabularx}{\textwidth}{l X}
\toprule
\textbf{Component} & \textbf{Specification} \\
\midrule
Policy Network Architecture (Joint) &  1-layer MLP (input = 27, output=64, ReLU), 1 layer GRU (input = 64, output = 64), 1 layer MLP (input=64, output=6)  \\
Target Policy Network Architecture (Joint) &  1-layer MLP (input = 27, output=64, ReLU), 1 layer GRU (input = 64, output = 64), 1 layer MLP (input=64, output=6)  \\

\multicolumn{2}{l}{Mixing Network Architecture (Joint)} \\
\quad Query Network &  1-layer MLP (input =188, output=32, ReLU),  1-layer MLP (input = 32, output=32, ReLU),  1-layer MLP (input = 32, output=1) \\
\quad Value Network &  1-layer MLP (input =54, output=32, ReLU),  1-layer MLP (input = 32, output=32, ReLU),  1-layer MLP (input = 32, output=1) \\
\quad Action Encoding &  1-layer MLP (input =134, output=134, ReLU),  1-layer MLP (input = 134, output=134) \\

\multicolumn{2}{l}{Target Mixing Network Architecture (Joint)} \\
\quad Query Network &  1-layer MLP (input =188, output=32, ReLU),  1-layer MLP (input = 32, output=32, ReLU),  1-layer MLP (input = 32, output=1) \\
\quad Value Network &  1-layer MLP (input =54, output=32, ReLU),  1-layer MLP (input = 32, output=32, ReLU),  1-layer MLP (input = 32, output=1) \\
\quad Action Encoding &  1-layer MLP (input =134, output=134, ReLU),  1-layer MLP (input = 134, output=134) \\

Policy optimizer & Adam, alpha $0.99$, epsilon $1\times10^{-5}$ \\
Target policy optimizer & Adam, alpha $0.99$, epsilon $1\times10^{-5}$ \\
Start epsilon greedy & 1.0 \\
Minimum epsilon greedy &  0.05 \\
Discount factor $\gamma$ & 0.99 \\
Target network update interval & 1 episode\\
Start learning rate & $1 \times 10^{-2}$ \\
Minimum learning rate & $1 \times 10^{-6}$ \\
TD Lambda & 1.0 \\
Replay buffer size & 1000 \\
Parallel environment & 32 \\
Parallel episodes per buffer episode & 32 \\
Training batch size & 32 \\
Warm up buffer episodes & 32 \\
\bottomrule
\end{tabularx}
\label{tab:qtran_hyperparams}
\end{table}
\FloatBarrier
\begin{table}[!htbp]
\centering
\caption{Model architecture and hyperparameters used for MAVEN.}
\begin{tabularx}{\textwidth}{l X}
\toprule
\textbf{Component} & \textbf{Specification} \\
\midrule
Policy Network Architecture (Joint) &  1-layer MLP (input = 27, output=64, ReLU), 1 layer GRU (input = 64, output = 64), 1 layer MLP (input=64, output=6)  \\
Target Policy Network Architecture (Joint) &  1-layer MLP (input = 27, output=64, ReLU), 1 layer GRU (input = 64, output = 64), 1 layer MLP (input=64, output=6)  \\

\multicolumn{2}{l}{Noise Mixing Network Architecture (Joint)} \\
\quad Hypernet Weights 1 & 1-layer MLP (input=116, output=64) \\
\quad Hypernet Bias 1 & 1-layer MLP (input=116, output=32) \\
\quad Hypernet Weights 2 & 1-layer MLP (input=116, output=32) \\
\quad Skip Connection & 1-layer MLP (input=116, output=2) \\
\quad Value network & 1-layer MLP (input=116, output=32, ReLU), 1-layer MLP(input=32,output=1) \\

\multicolumn{2}{l}{Target Noise Mixing Network Architecture (Joint)} \\
\quad Hypernet Weights 1 & 1-layer MLP (input=116, output=64) \\
\quad Hypernet Bias 1 & 1-layer MLP (input=116, output=32) \\
\quad Hypernet Weights 2 & 1-layer MLP (input=116, output=32) \\
\quad Skip Connection & 1-layer MLP (input=116, output=2) \\
\quad Value network & 1-layer MLP (input=116, output=32, ReLU), 1-layer MLP(input=32,output=1) \\

RNN Aggregator & 1-layer GRU (input=116, output=2) \\
Discriminator  & 1-layer MLP (input=116, output=32, ReLU), 1-layer MLP (input=32, output=2),\\

Actor optimizer & RMSprop, alpha $0.99$, epsilon $1\times10^{-5}$ \\
Target actor optimizer & Adam, alpha $0.99$, epsilon $1\times10^{-5}$ \\
Use skip connection in mixer & False \\
Use RNN aggregation & False \\
Discount factor $\gamma$ & 0.99 \\
Target network update interval & 1 episode\\
Learning rate & $5 \times 10^{-5}$ \\
TD Lambda & 1.0 \\
Replay buffer size & 1000 \\
Parallel environment & 32 \\
Parallel episodes per buffer episode & 1 \\
Training batch size & 32 \\
\bottomrule
\end{tabularx}
\label{tab:maven_hyperparams}
\end{table}
\FloatBarrier
\msubsection{Compute}\label{m:comp}
For each simulation 2 CPUs were allocated and the 32 parallel environments were multithreaded. All algorithms expect for SAF were able to run without GPUs while SAF used a single A100 for each simulation. All algorithms, except for VDN, QMIX and QTRAN can finish at 10000 episodes for all 10 simulations within 4 days while the aforementioned algorithms take 7 days. It is possible to use a GPU for these value mixer mechanisms for faster data collection but this was not done to collect the data. The correction term experiments take 7 days to collect 26000 episodes and do not benefit from GPUs since their networks are too small. The Hessian term can be approximated with finite difference technique  or with Pearlmutter's trick.

\FloatBarrier
\msubsection{Manitokan task setup}
The Manitokan Task is a grid world for tractable analysis. The key, agents and doors are randomly initialized at the beginning of each episode and the actions \textit{drop} and \textit{toggle} were additionally pruned when an agent is not holding a key for reasonable environment logic but are not necessary to be removed for the task to work. The doors look the same to both agents. Everything else was described in \ref{sec:manito}.
\FloatBarrier
\newpage

\asection{}

\asubsection{Self Correction }
\DontPrintSemicolon
\SetKwComment{Comment}{\#\ }{}          
\SetKwProg{Fn}{def}{:}{}               
\SetKw{KwRet}{return}                  
\SetKwFunction{SelfCorrection}{self\_correction}
\SetKwFunction{TDLambda}{build\_td\_lambda\_targets}

\begin{algorithm}[hbtp!]
\caption{Self-correction}\label{alg:two}

\KwIn{$\hat{\mathcal{R}}_{0:T},\; Q_{0:T},\; p_{0:T} \sim \pi_{\Theta^i}(a\mid o),\; \mathcal{R},\; a_{0:T}$}
\KwOut{\texttt{Loss}}
\BlankLine
\BlankLine

\Fn{\SelfCorrection{$\hat{\mathcal{R}}_{0:T},\, Q_{0:T},\, p_{0:T},\, \mathcal{R},\, a_{0:T}$}}{
\BlankLine

\BlankLine
    $\hat{\mathcal{R}}^i :=  \hat{\mathcal{R}}_{0:T}[\mathcal{R}<1]$
    \Comment*[r]{Buffer of individual rewards}
\BlankLine
\BlankLine
    $\hat{\mathcal{R}}^c :=  \hat{\mathcal{R}}_{0:T}[\mathcal{R}\geq1]$
    \Comment*[r]{Buffer of collective rewards}
\BlankLine
\BlankLine
    $Q_{0:T}^c :=  Q_{0:T}-\hat{\mathcal{R}}_{0:T}^i$
    \Comment*[r]{Value estimates of collective rewards}
\BlankLine
\BlankLine
$G^{\lambda,c}_{0:T} = \TDLambda(\hat{\mathcal{R}}^{c},\, Q^{c}_{0:T},\, \gamma=0.95,\, \lambda=1.0)$
    \Comment*[r]{TD-$\lambda$ targets}

\BlankLine
\BlankLine

$A^{\lambda,c}_{0:T} = G^{\lambda,c}_{0:T} - Q^{c}_{0:T}$
    \Comment*[r]{TD-$\lambda$ advantage for collective objective}
\BlankLine
\BlankLine
    $\nabla_{\Theta^i} J_c
      = \nabla_{\Theta^i}\!\left(\log p_{0:T} \cdot (A^{\lambda,c}_{0:T})\right)$
    \Comment*[r]{Policy gradient for the collective objective}
\BlankLine
\BlankLine
    $H_T = \log(p_T)$
\BlankLine
\BlankLine
    $\nabla_{\Theta^i} J_{c}\Psi = \dfrac{\nabla_{\Theta^i} J_{c,T}}{H_T^i}$
    \Comment*[r]{normalize by terminal entropy/log-prob}
\BlankLine
\BlankLine

$G^{\lambda}_{0:T} = \TDLambda(\hat{\mathcal{R}}_{0:T},\, Q_{0:T},\, \gamma=0.95,\, \lambda=1.0)$
    \Comment*[r]{TD-$\lambda$ targets}

\BlankLine
\BlankLine

$A^{\lambda}_{0:T} = G^{\lambda}_{0:T} - Q_{0:T}$
    \Comment*[r]{TD-$\lambda$ advantage for total reward objective}
    
\BlankLine
\BlankLine
    $\texttt{update}  = \nabla_{\Theta^i}\dfrac{1}{T}\left(
        \sum_{t=0}^{T-1}\log(p_t^i)(A^{\lambda}_{t}) + \log(p_T^i)(A^{\lambda}_{T})+\nabla_{\Theta^i} J_{c}\Psi\right)$
\BlankLine
\BlankLine
    \KwRet{} \texttt{update} 
}
\end{algorithm}

\newpage
\esection{}
The experiments provided below offer  insights into the challenge of the Manitokan Task, and further empirical validation of the correction and self correction terms.

\esubsection{COMA's loss becomes negative} \label{coma_neg}
\FloatBarrier
\begin{figure}[htbp]
    \centering
    \begin{subfigure}[b]{0.45\textwidth}
        \centering
        \includegraphics[width=5cm]{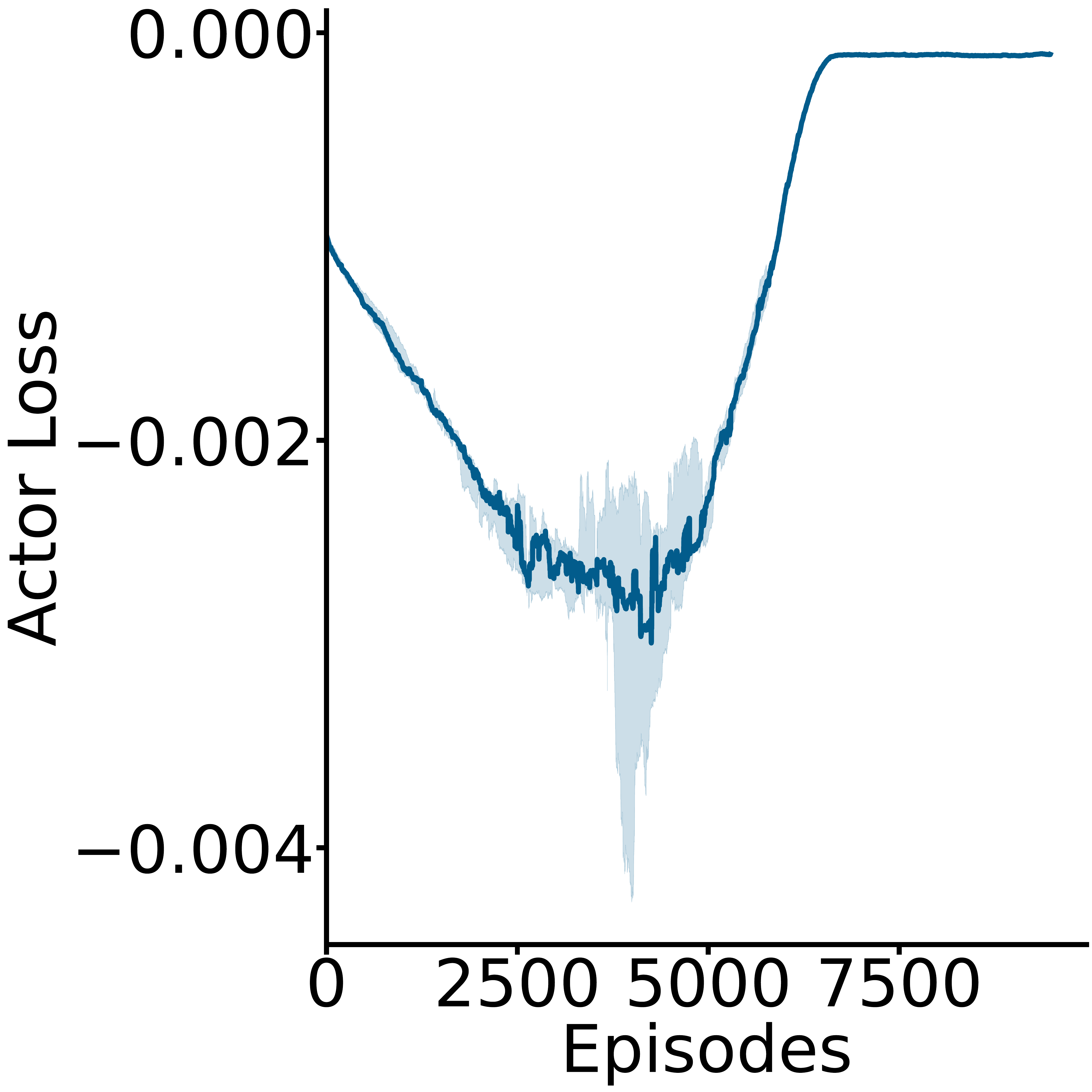}
        \caption{Actor Loss}
        \label{fig:comaa}
    \end{subfigure}
     \hspace{0.01\textwidth}
    \begin{subfigure}[b]{0.45\textwidth}
        \centering
       \includegraphics[width=5cm]{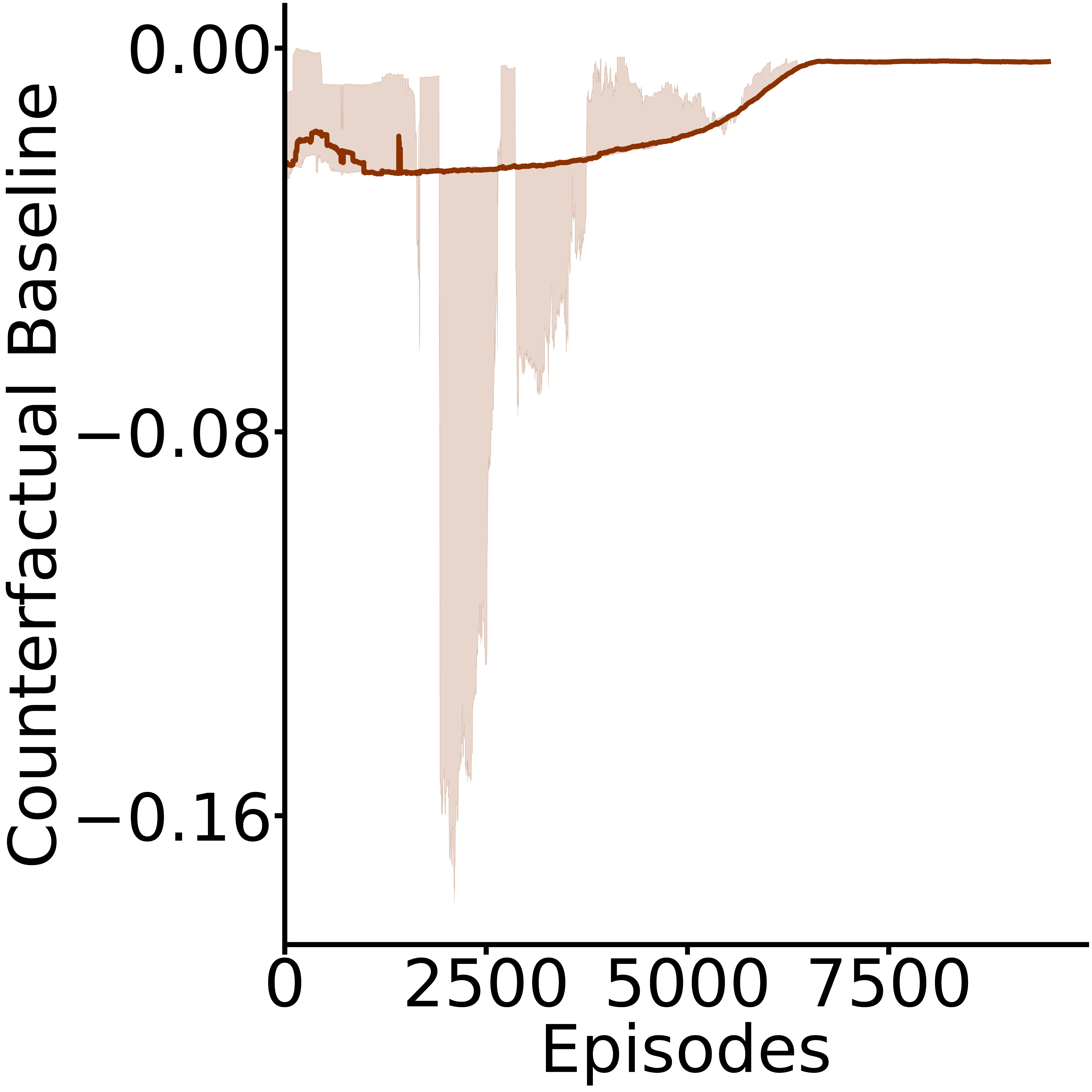}
        \caption{Counterfactual Baseline}
        \label{fig:comab}
    \end{subfigure}
    \caption{a) Policy loss of the COMA model b) Counterfactual baseline in the COMA policy update}
    \label{fig:coma}

\end{figure}

\FloatBarrier
COMA persistently collapsed even though it exhibited similar learning behaviour to PG (a closely related model). The policy loss and baseline curves show increasing instability with large variance spikes before converging to a value around 0.0. Perhaps this collapse is from the difficulty of leaving a hidden gift between individual and collective incentives. The original COMA paper \citep{foerster2018counterfactual} even mentions a struggle for an agent overcoming an individual reward, although exterior to hidden gifts, may be cause for the instability.
\newpage
\esubsection{Optimal key drop rate is unattained by all agents}
\label{e:keys}

\begin{figure}[htbp]
    \centering
    \begin{subfigure}[b]{0.45\textwidth}
        \centering
        \includegraphics[width=6.5cm]{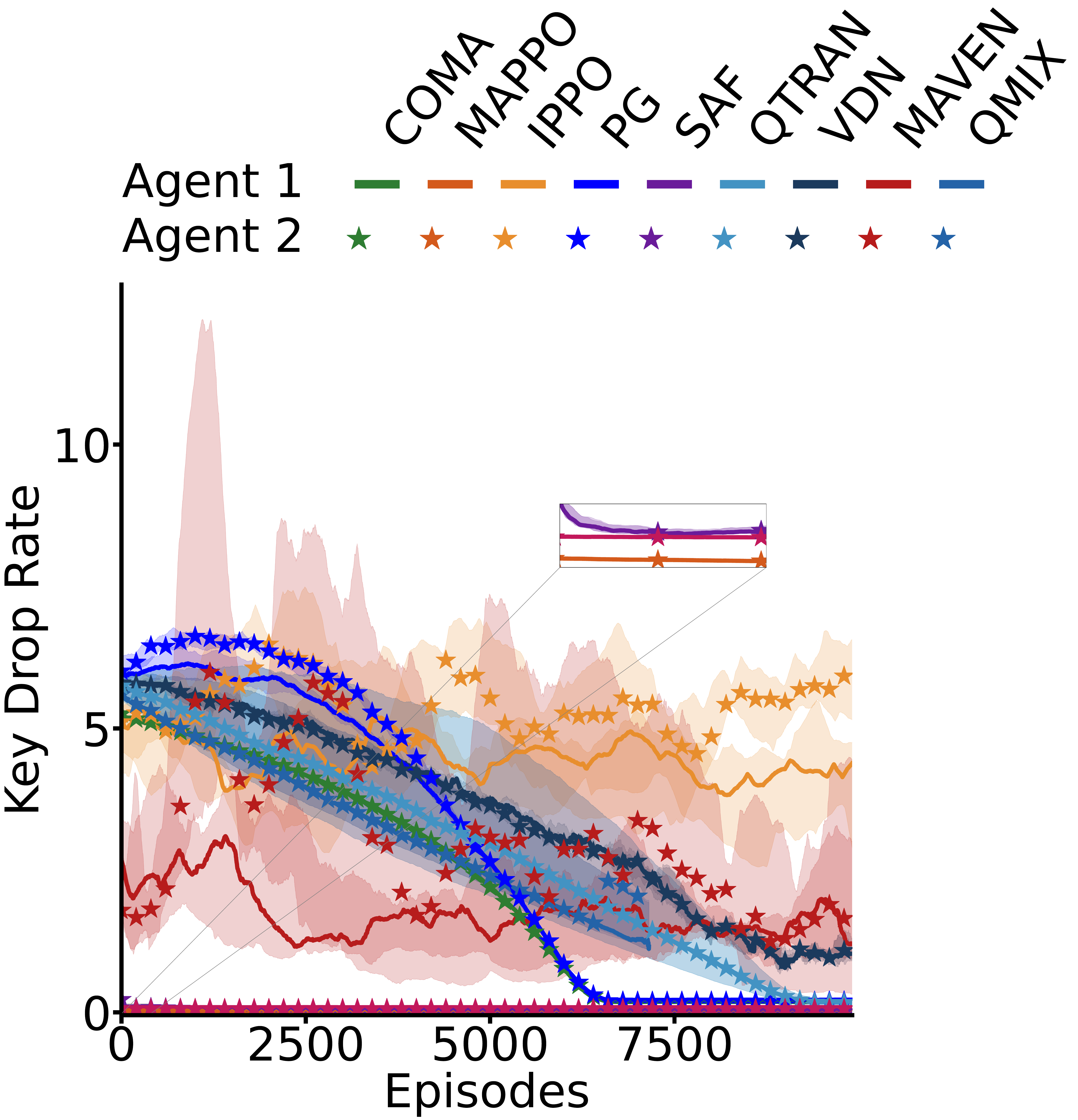}
        \label{fig:keya}
    \end{subfigure}
     \hspace{0.01\textwidth}
    \begin{subfigure}[b]{0.45\textwidth}
        \centering
       \includegraphics[width=5cm]{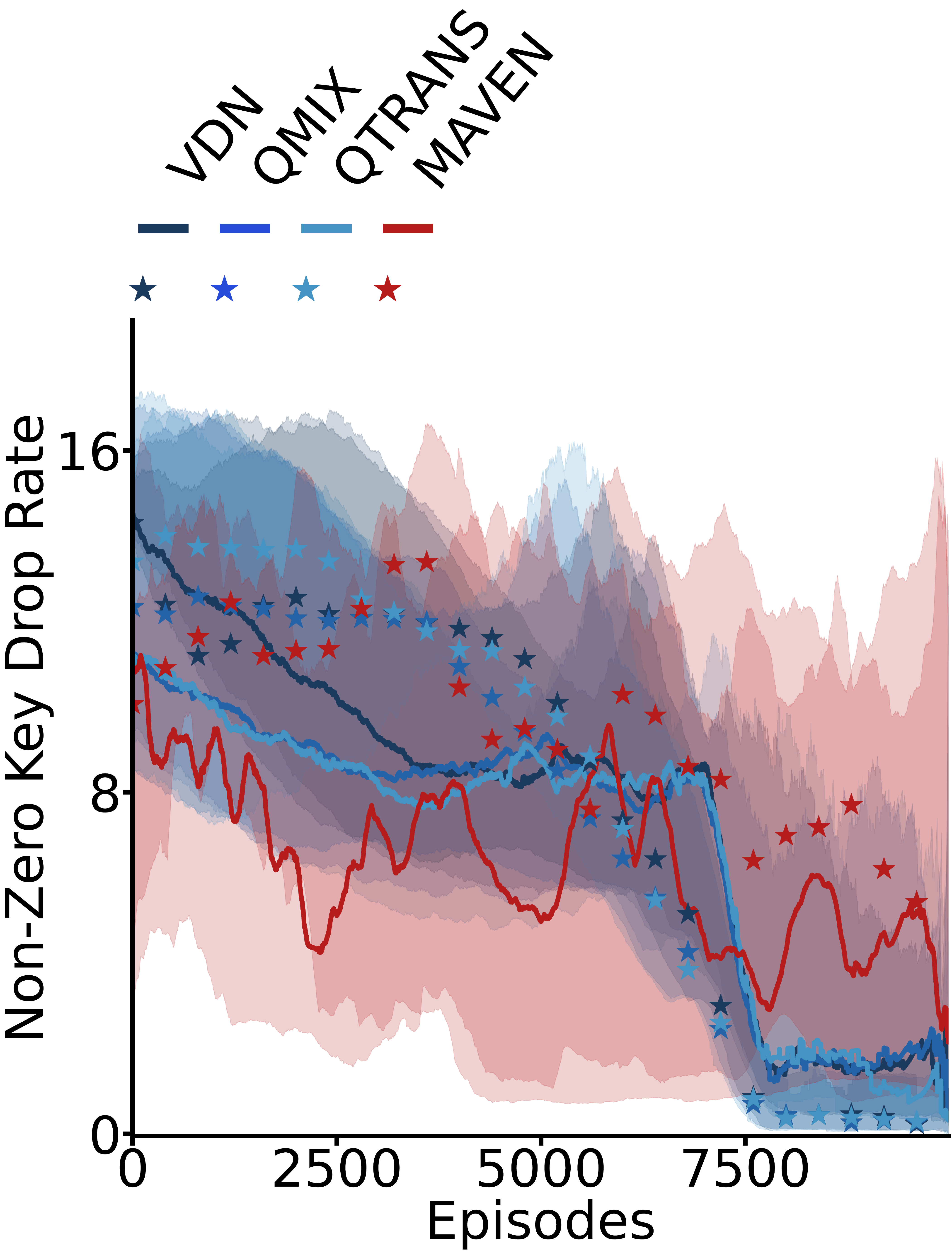}
    
        \label{fig:keyb}
    \end{subfigure}
    \caption{a) Key drop rate (i.e. cumulative key drops) averaged across parallel episodes and runs. b) Non-zero key drop rate (i.e. cumulative key drops) averaged across parallel episodes that had key drops and runs.
 }
    \label{fig:keydrop_rates}
\end{figure}

\FloatBarrier

For most of the MARL agents (VDN, QMIX, QTRAN, MAVEN) the key drop rate always converged to exactly zero (Fig. \ref{fig:keydrop_rates}a), hence the total collapse in collective success in the task. In the case of MAPPO, and SAF, we observed that the agents learned to pick up the key and open their individual doors, but minimized the number of key drops to close to zero (Fig. \ref{fig:keydrop_rates}b). As a result, the collective success rate was also close to zero. In contrast, IPPO  did not exhibit a collapse in key drops but had an oscillatory effect where one would agent increase their keydrops while the other reduces theirs. This explains IPPO's slightly better success in obtaining the collective reward (Fig. \ref{fig:2}a). Interestingly, COMA and decentralized PG showed very low, but non-zero rates of key drop (Fig. \ref{fig:keydrop_rates}a), however only PG exhibited a non-zero collective success rate (Fig. \ref{fig:2}a). This was because even though COMA agents learned to occasionally drop the key, the counter-factual baseline caused the loss to become excessively negative (see E.\ref{coma_neg}). 

One complication with measuring the key drop rate is that if the agents never even pick up the key then the key drop rate is necessarily zero. To better understand what was happening in here, we examined the ``non-zero key drop rate'', meaning the rate at which keys were dropped if they were picked up. The non-zero key drop rate showed that the value mixer MARL agents begin by dropping the key after picking it up some of the time, but eventually converge to a policy of simply holding or avoiding the key (Fig. \ref{fig:keydrop_rates}b). The variance in drop rates is increased except at the end for VDN, QMIX and QTRAN. This further emphasizes the challenge of hidden gifts.

\FloatBarrier
\newpage
\esubsection{Changing which agent steps first between episodes harms performance}\label{e:turnorder}

\begin{figure}[htbp]
    \centering
    \begin{subfigure}[b]{0.48\textwidth}
        \centering
        \includegraphics[width=5.5cm]{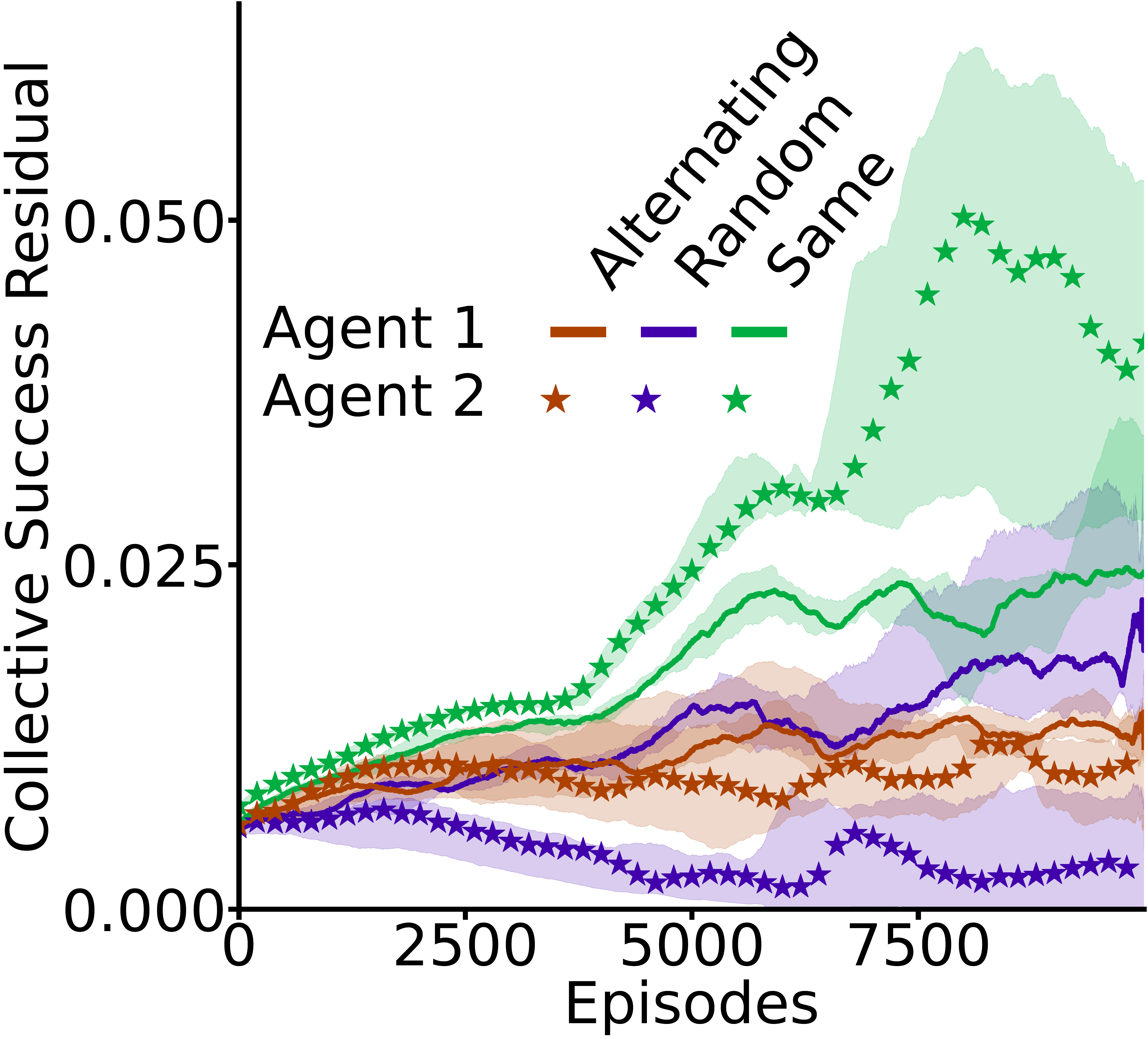}
        \caption{Collective Success Rate Residual}
        \label{fig:sub1}
    \end{subfigure}
      \hspace{0.01\textwidth}
    \begin{subfigure}[b]{0.48\textwidth}
        \centering
       \includegraphics[width=5.5cm]{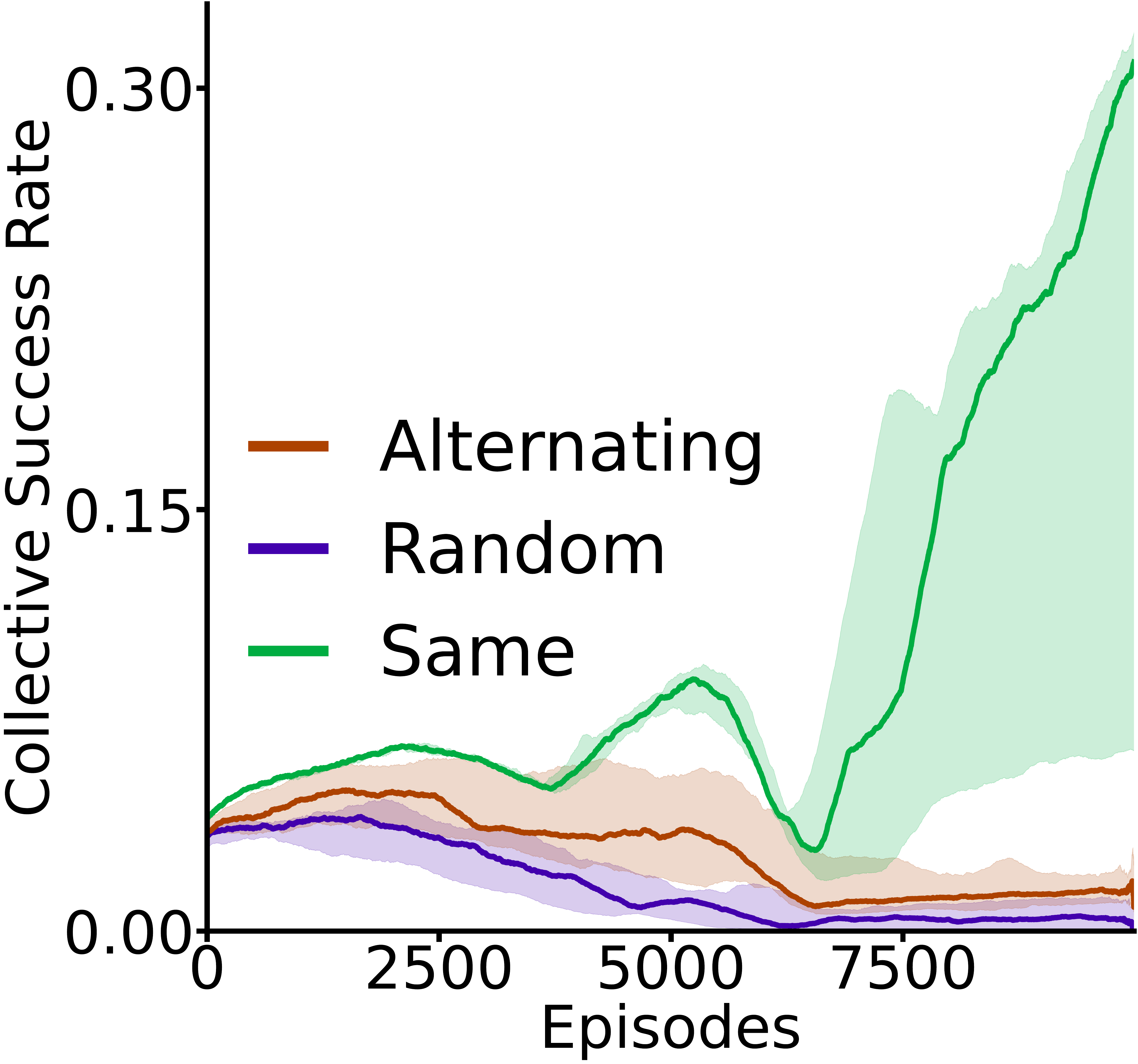}
        \caption{Collective Success Rate}
        \label{fig:sub2}
    \end{subfigure}
    \caption{a) The contribution of an agent's reward accumulation to success weighted by their total reward comparing policy gradient agents with action history of the same agent stepping first (i.e. agent 1 then agent 2), alternating agents stepping first (i.e. agent 1 steps first on odd numbered episodes and agent 2 steps first in even numbers episodes), and a random agent is selecting to step first.  b) Success rate between different step ordering each episode.}
    \label{fig:subfigureExample}
\end{figure}
\FloatBarrier
The collective success residual is calculated as $(r^c-r^i)\times r^i$ where $(r^c-r^i)$ describes how much an agent $i$ is contributing to the collective success while weighting it by $r^i$ shows if the agents are increasing that success rate. Interestingly, alternating which agent goes first between episodes creates oscillations in the collective success rate residual where one agent receiving more reward means the other agent receives less. Greatly reducing the success. Moreover, randomly selecting an agent to go first biases the first agent to increase their reward and almost removes all success. These effect may be caused by uncertainty associated with which agent can reach the key when the other agent is in sight. For example in the random case, if agent $i$'s current policy has learned for the past five updates that it will pick up the key, when both agents are equal distance from the key, there will be a action prediction error. This uncertainty increases the difficulty of the credit assignment problem.

\newpage
\esubsection{Randomizing the policy can increase collective success slightly} \label{rand}

\begin{figure}[htbp]
    \centering
    \begin{subfigure}[b]{0.48\textwidth}
        \centering
        \includegraphics[width=5.5cm]{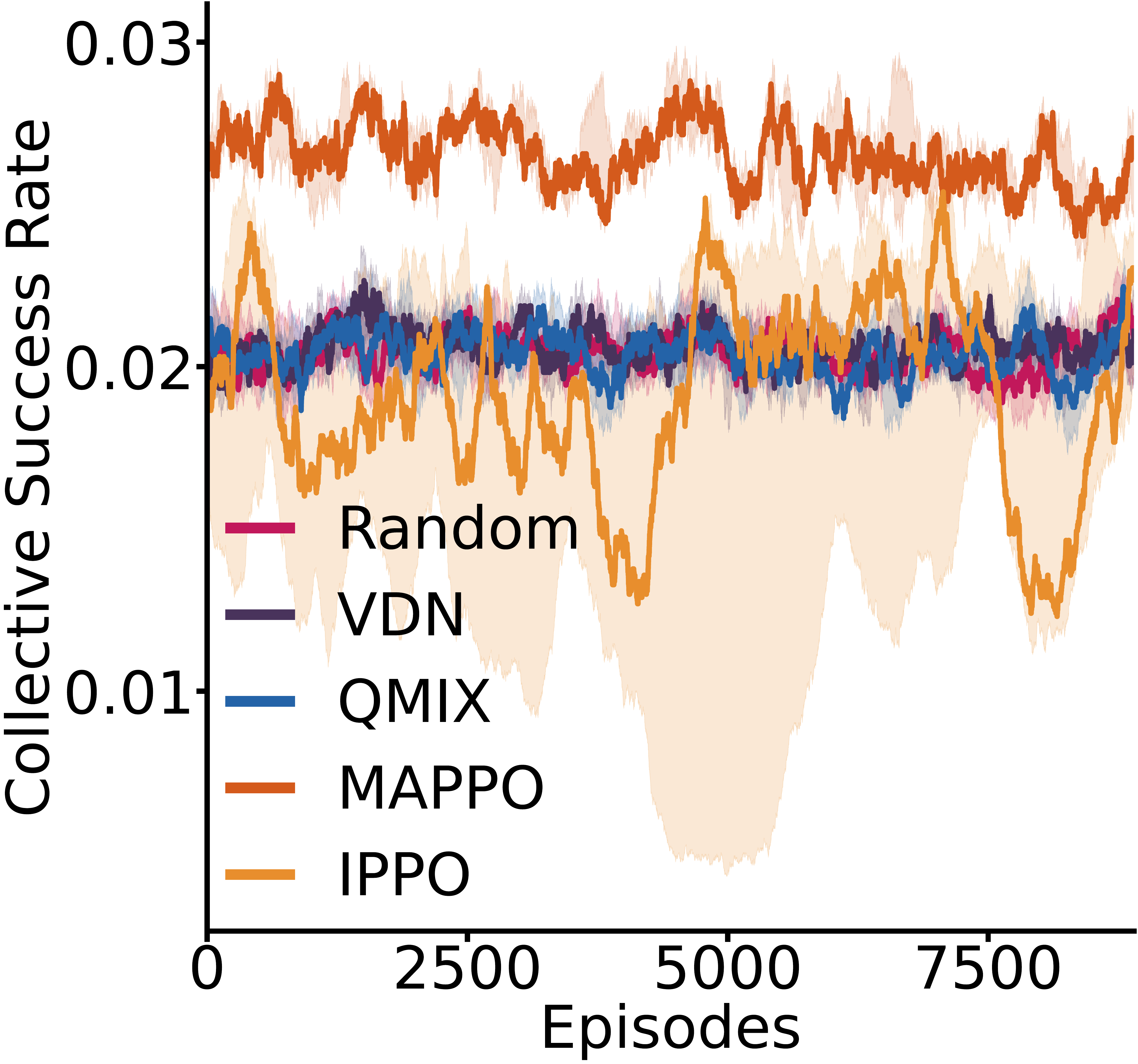}
        \caption{Collective Success Rate}
        \label{fig:sub1}
    \end{subfigure}
      \hspace{0.01\textwidth}
    \begin{subfigure}[b]{0.48\textwidth}
        \centering
       \includegraphics[width=5.5cm]{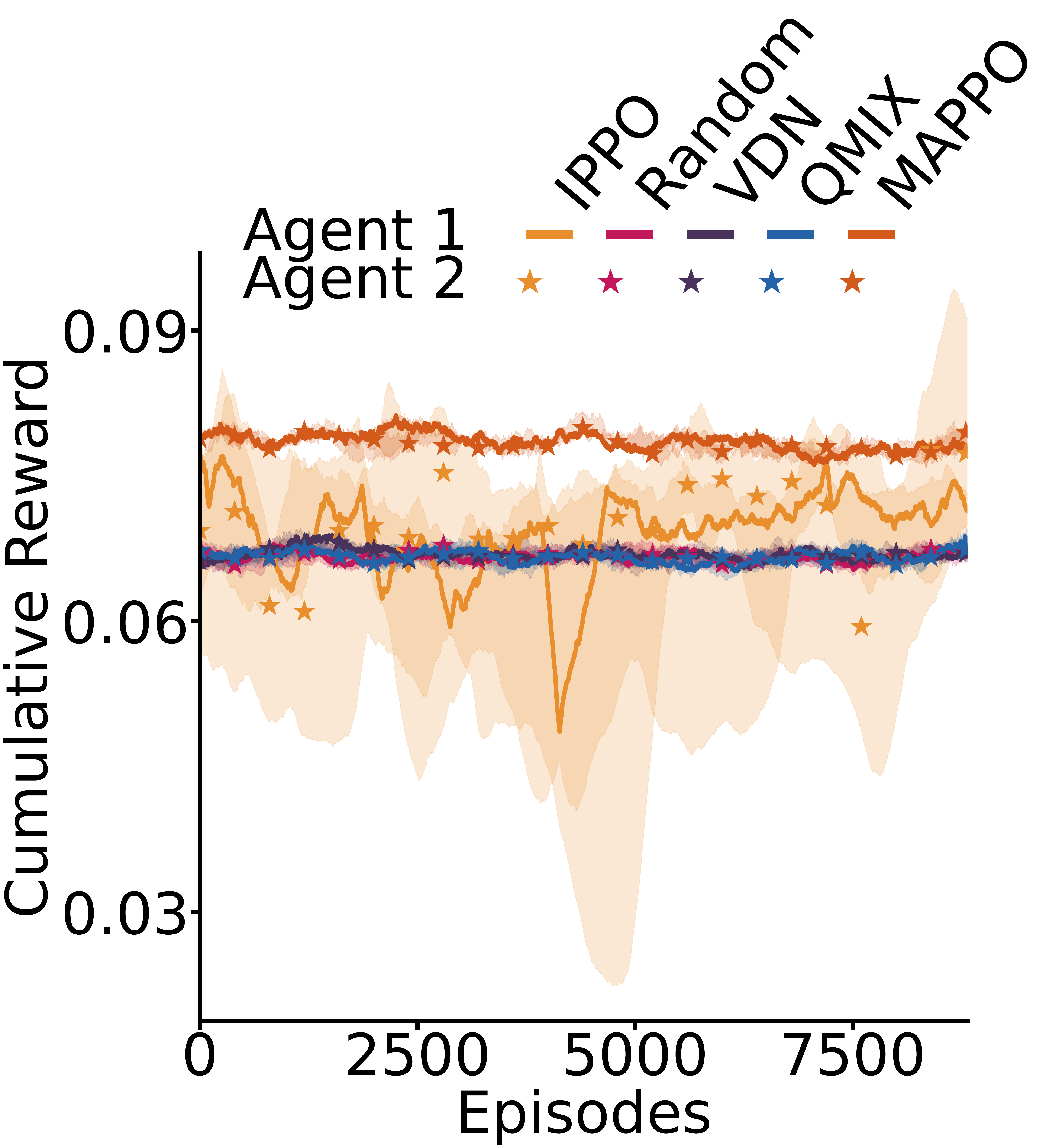}
        \caption{Cumulative Reward}
        \label{fig:sub2}
    \end{subfigure}
    \caption{a) Comparing agents of MAPPO, IPPO, VDN and QMIX models with a randomization applied to their policies b) The cumulative reward for randomized policy agents}
    \label{fig:subfigureExample}
\end{figure}
\FloatBarrier
PPO agents had their value function learning rates set to $0.001$ while the policy learning rates where kept as $0.000001$. This meant the policy would always prefer initial episodes and converge quickly to those while the value function weighting them more evenly to converge further in the training process. VDN and QMIX use epsilon greedy in their strategy and simply increasing the time of  decay for this mechanism led these agents to be more random throughout the experiment. 

This policy randomization process very slightly improved these agents the success rates' compared to those in the main results Fig \ref{fig:2}a but decreased the cumulative reward for the PPO agents than those in Fig \ref{fig:2}b. The random policy aligned VDN and QMIX to the random action baseline more or less, and avoided collapse.

\newpage
\esubsection{Behavioural variations appear between models with inter agent distance and minimizing the steps to the first reward}
\FloatBarrier

\begin{figure}[htbp]
    \centering
    \begin{subfigure}[b]{0.48\textwidth}
        \centering
        \includegraphics[width=5.5cm]{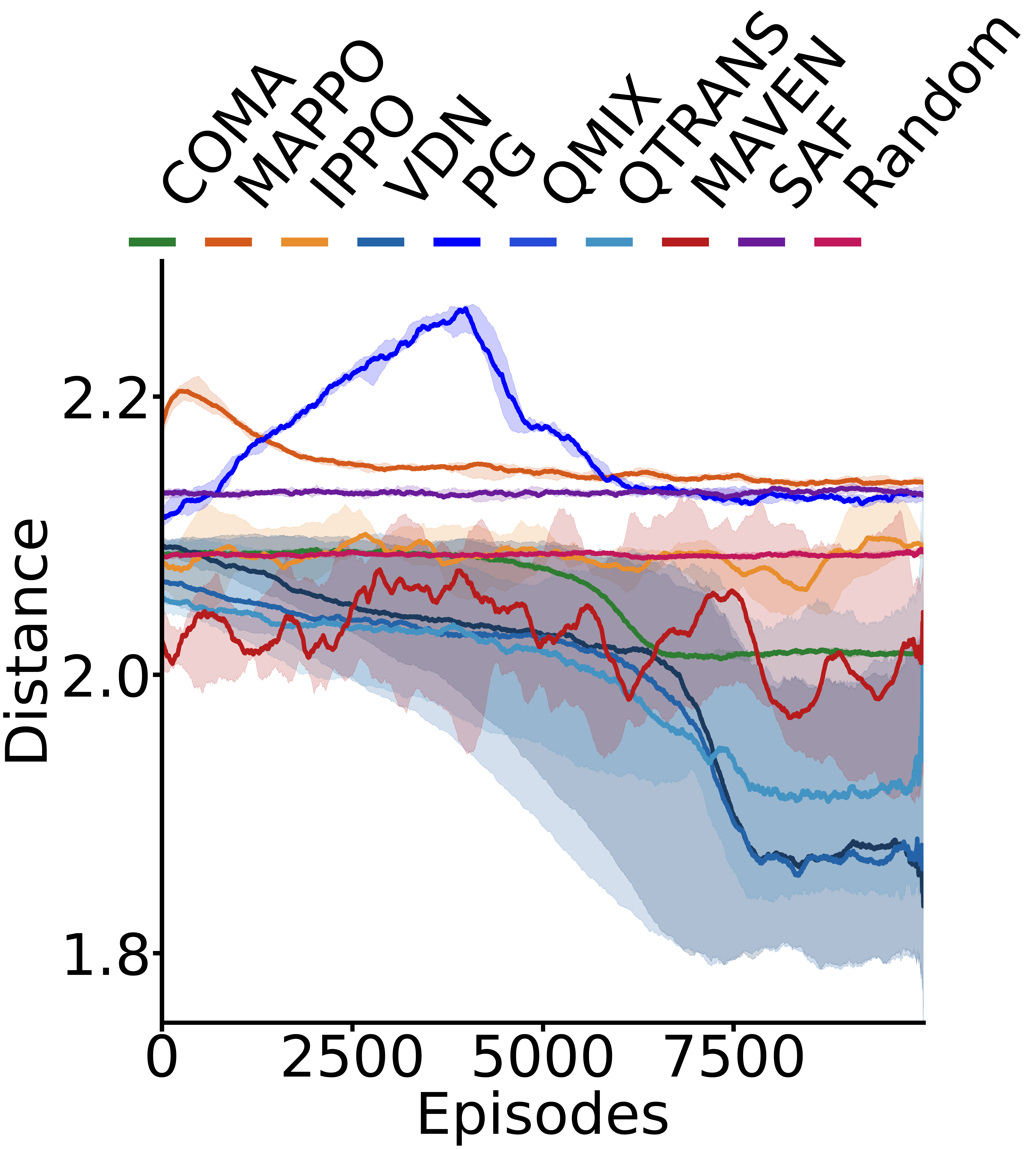}
        \caption{Distance}
        \label{fig:behav_1a}
    \end{subfigure}
     \hspace{0.01\textwidth}
    \begin{subfigure}[b]{0.48\textwidth}
        \centering
       \includegraphics[width=5.5cm]{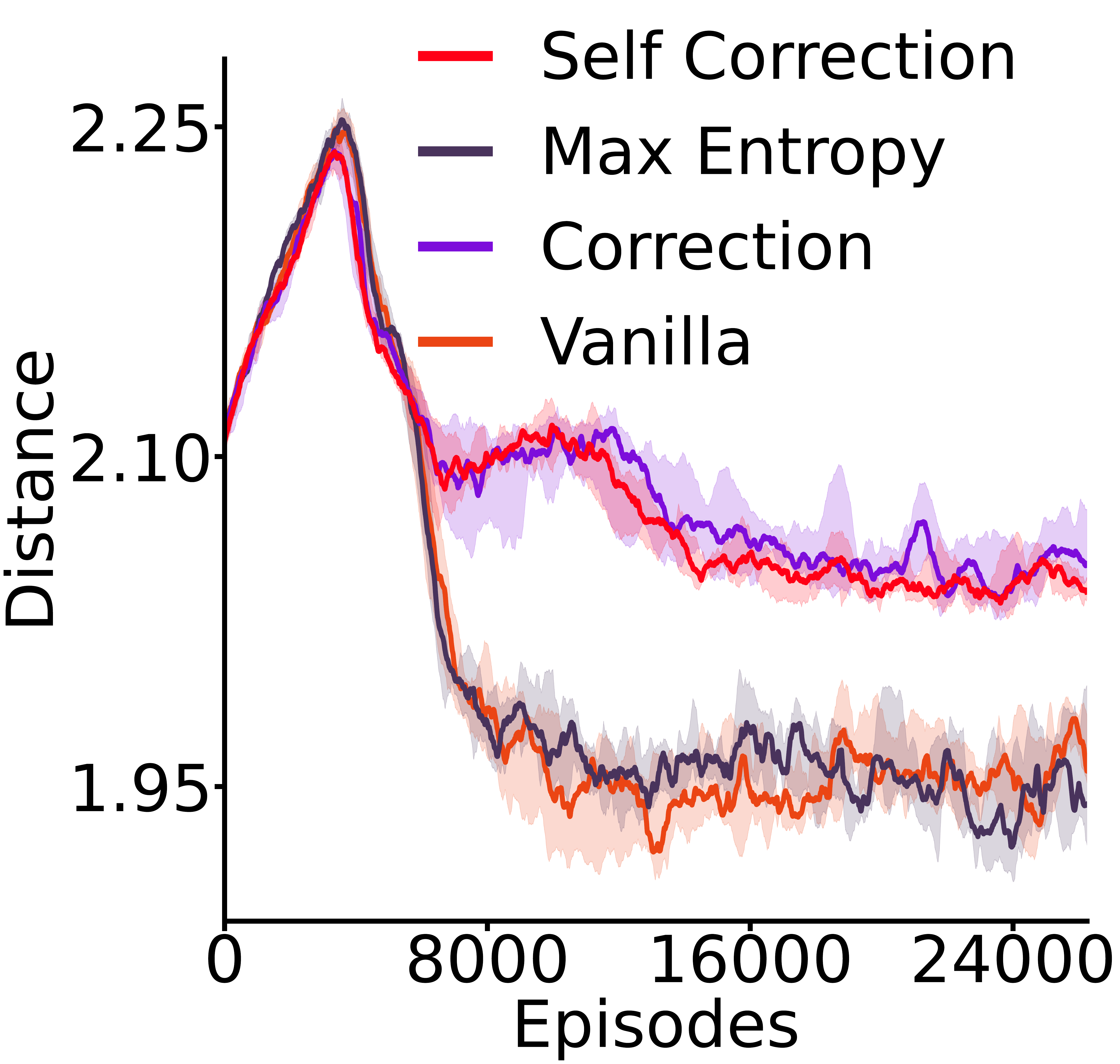}
        \caption{Correction Term Distance}
        \label{fig:behav_1b}
    \end{subfigure}
    \caption{a) Euclidean distance between agents averaged over parallel environments and simulations across our tested models  b) Euclidean distance comparing policy gradient agents with action history and variance reduction terms.}
    \label{fig:behav1}
\end{figure}
Although the 2-agent Manitokan Task is a four by four grid world, we measured the euclidean distance between agents to see if they become more coordinated or adversarial when learning hidden gifting. In Fig \ref{fig:behav1}a, PG agents exhibited the highest exploration phase but eventually converged to a lower distance. MAPPO agents also has a similar but substantially smaller exploration effect in the very beginning while SAF did not have any exploration phases. IPPO and MAVEN agents similarly hovered below the random baseline but MAVEN agents were closer to each other. COMA agents begin around random but converge to be closer to each other as well.  Value mixer agents VDN, QMIX and QTRAN all are on average closer to each other but QTRAN agent agents converge further apart. 

In Fig \ref{fig:behav1}b, vanilla and max entropy PG agents with action history become asymptotically closer to each other while the correction term agents converge further apart from them. The variance reduction in self correcting agents is also noticeable.

\begin{figure}[htbp]
    \centering
    \begin{subfigure}[b]{0.48\textwidth}
        \centering
        \includegraphics[width=5.5cm]{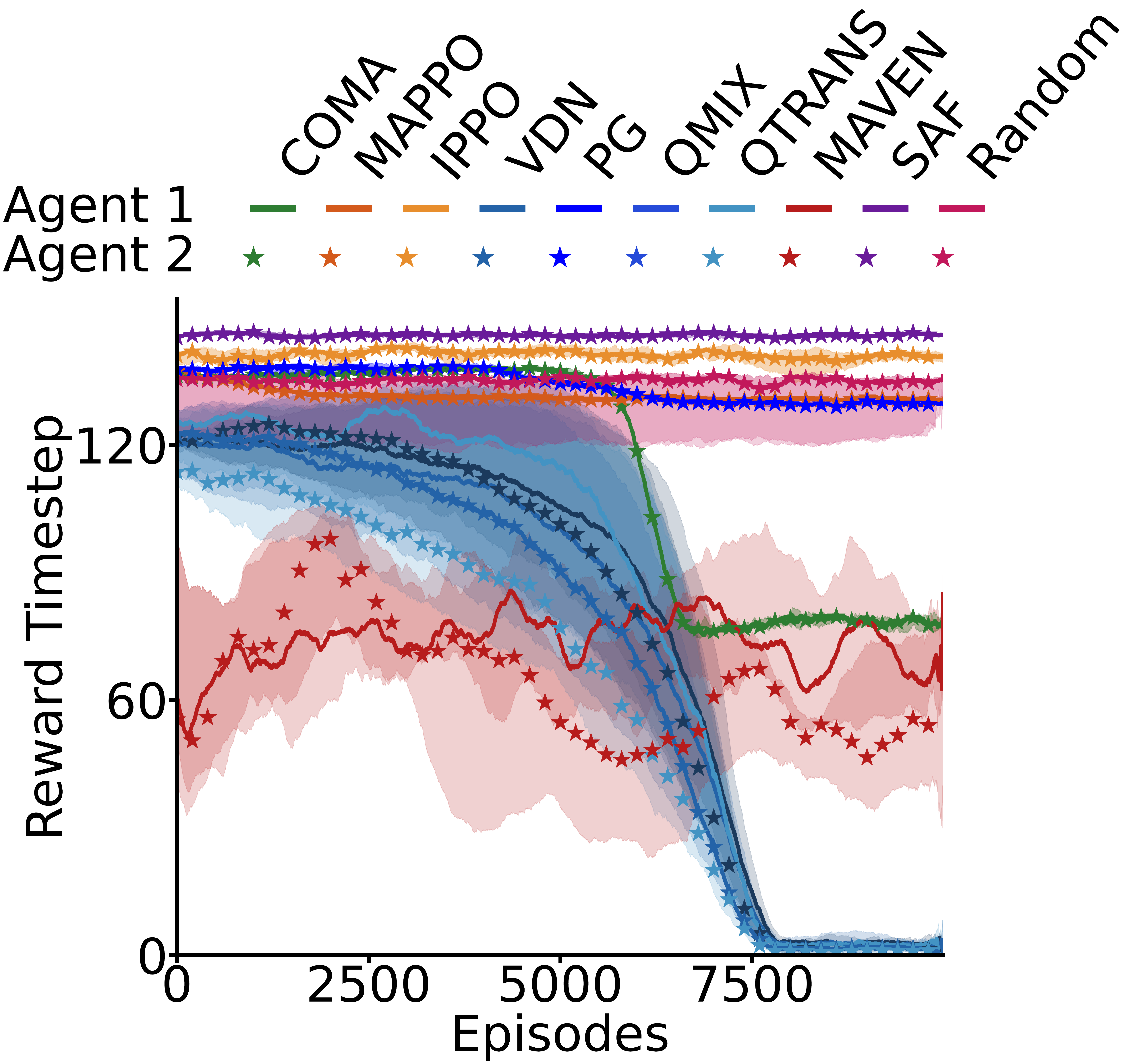}
        \caption{Timestep of First Reward}
        \label{fig:behav_2a}
    \end{subfigure}
      \hspace{0.01\textwidth}
    \begin{subfigure}[b]{0.48\textwidth}
        \centering
       \includegraphics[width=5.5cm]{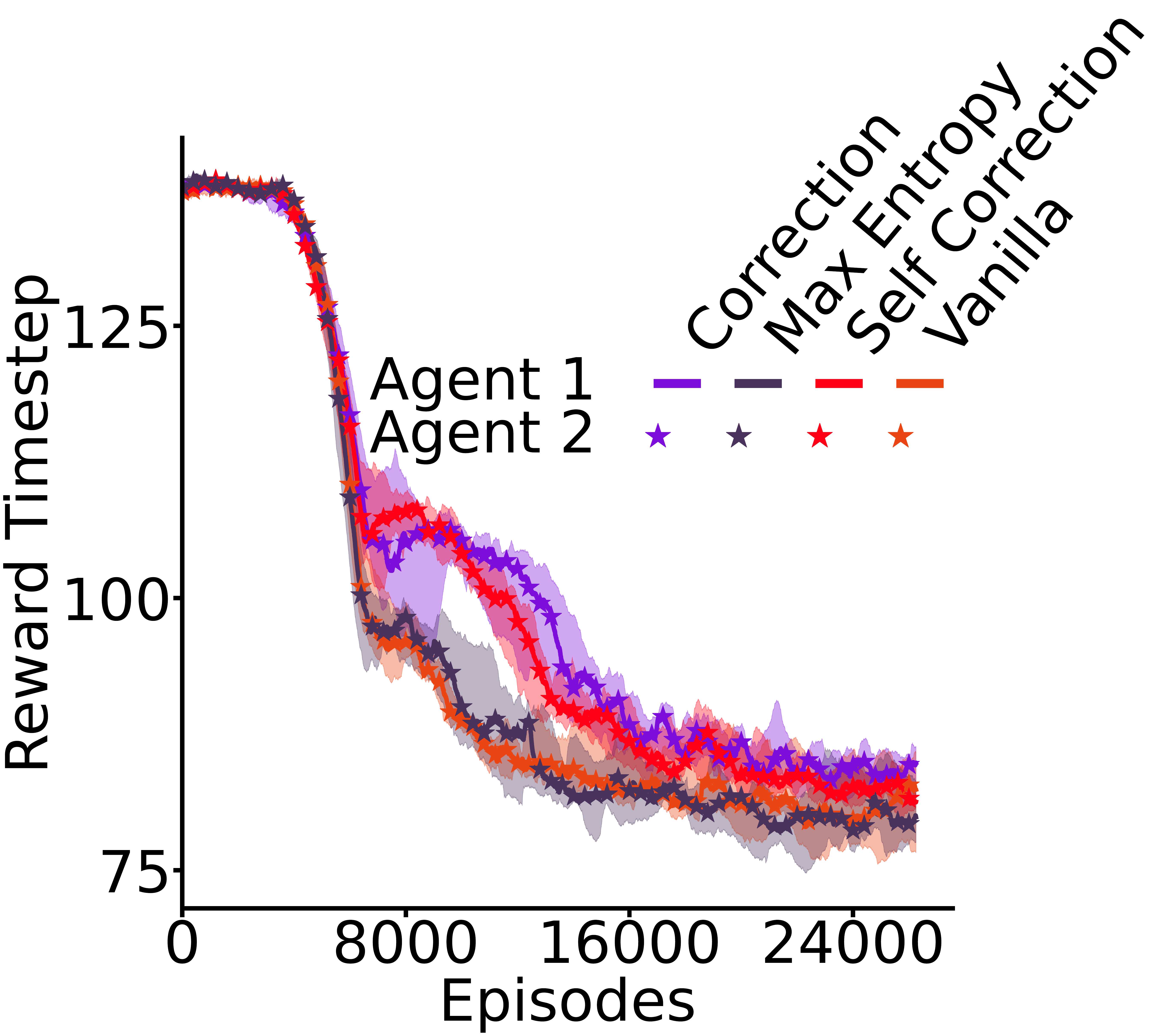}
        \caption{Correction Term Timestep of First Reward}
        \label{fig:behav_2b}
    \end{subfigure}
    \caption{ a) Timestep the first reward an agent received. b) Timestep the first reward a policy gradient agent with action history received.}
    \label{fig:behav2}
\end{figure}

\FloatBarrier

The reducing the timestep of the first reward is a way to measure if agents are improving their policies if cumulative reward also increases. In (Fig \ref{fig:behav2}a), PG, IPPO, MAPPO and SAF all converge quickly while PG and MAPPO learn policies of reducing the step slightly below random. COMA converges at a low timestep but this is most likely due to the collapse. MAVEN  oscillates at a timestep better than random but never converges and doesn't seem to learn a good policy and VDN, QMIX, and QTRAN collapse consistently with other results in Section 3. 

While in Fig \ref{fig:behav2}b, all decentralized PG algorithms with action history reduce their initial reward timesteps but models with the correction term converge slower. 
\newpage

\esubsection{Modifying the reward function enhances perspective on the challenge of the Manitokan task}\label{e:rmod}
\FloatBarrier
\begin{figure}[htbp]
    \centering
    \begin{subfigure}[b]{0.48\textwidth}
        \centering
        \includegraphics[width=5.5cm]{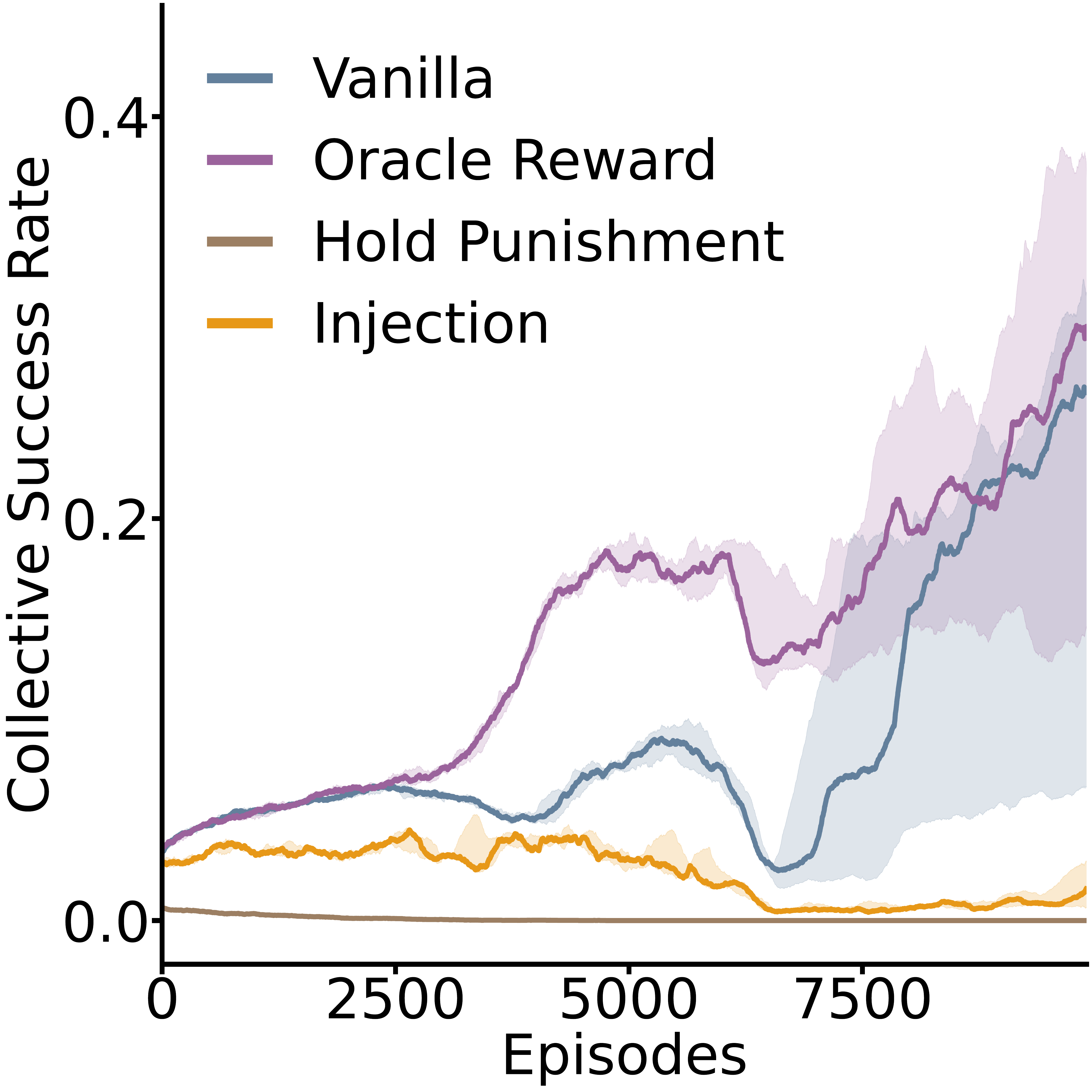}
        \caption{Collective Success Rate}
        \label{fig:sub1}
    \end{subfigure}
      \hspace{0.001\textwidth}
    \begin{subfigure}[b]{0.48\textwidth}
        \centering
       \includegraphics[width=5.5cm]{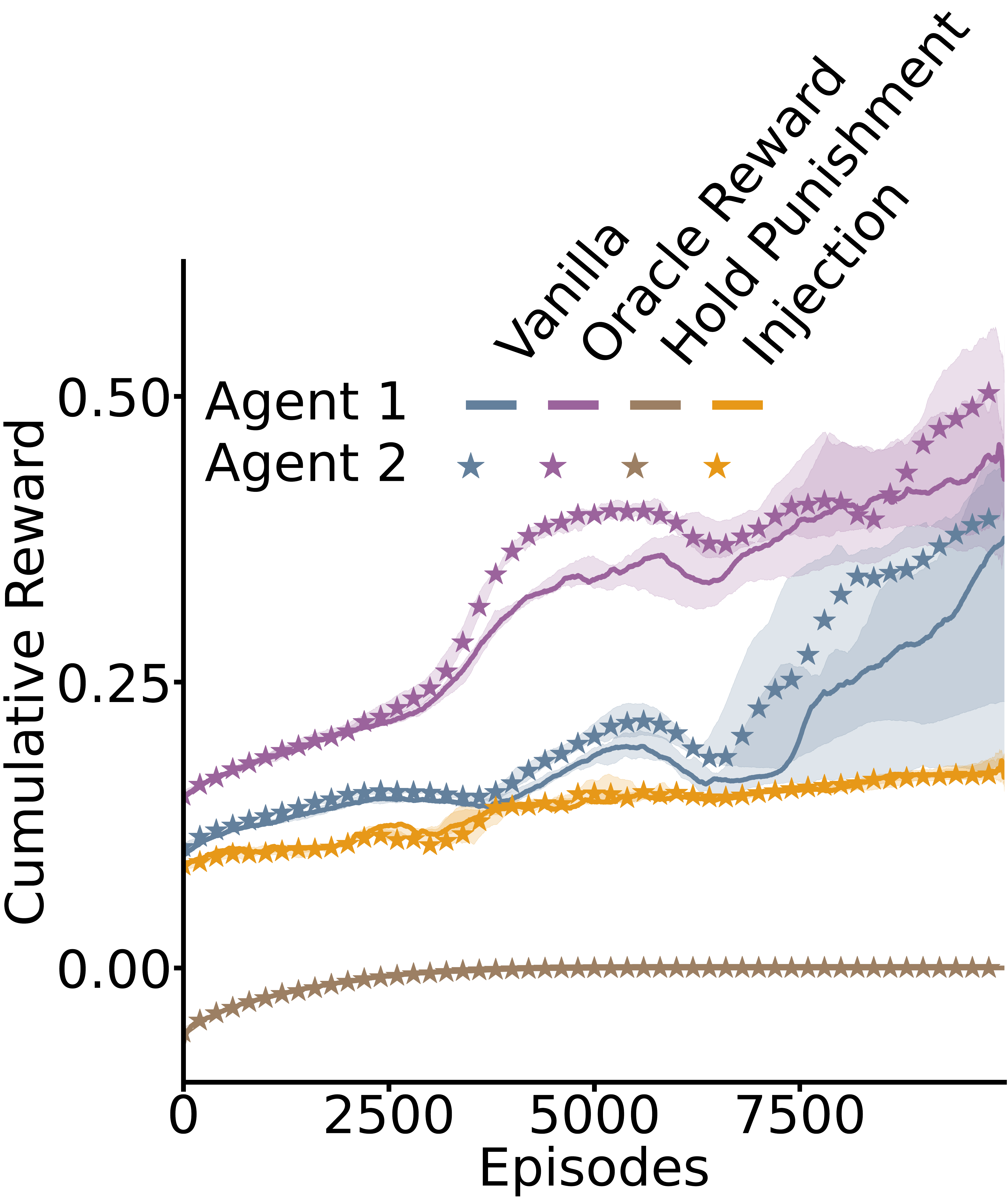}
        \caption{Cumulative Reward}
        \label{fig:sub2}
    \end{subfigure}
    \caption{a) Success rate of policy gradient agents with action history comparing the normal reward function with an oracle reward term (i.e. an agent receives a reward of 1 once for dropping the key after opening their door), a punishment term (i.e.. a negative reward of 1 is applied each step an agent holds their key after opening their door) and a reward injection term (i.e. randomly distributing normally smaller rewards around the standard rewards decaying over episodes) b) Cumulative reward to compare the modified reward functions }
    \label{fig:subfigureExample}
\end{figure}
\FloatBarrier
The reward function $\mathcal{R}$ in equation 1 to study hidden gifting behavior is both sparse with a hard to predict collective reward conditioned on the other agent's policy. We tested additional  reward conditions on PG agents with action history to see if  sample efficiency improvement can be found. Particularly, the oracle reward: $r_t^i \text{ the first key dropped after agent } i \text{'s door is opened } $, is the critical step to take for hidden gifting and when implemented the collective success rate increased quicker than the normal reward function. The punishment reward: $-0.5 \text{ for each step agent } i \text{ is holding the key after their door was opened}$, is also meant to induce gifting behavior but agents seemed to avoid the key altogether. Lastly, the injection reward where a set of rewards $r^d<r^i$ are normally distributed around rewards $r^i$ and $r^c$ which also served as the mean. $r^d$ was additionally reduced each episode for agents to prefer the standard rewards. Injection reduced the success rate severely but also reduced variance in accumulating the expected reward.

These minor modifications reemphasize the difficulty in hidden gifting, where our most performative agents still struggle even when rewarded for the optimal action.

\begin{figure}[htbp]
    \centering
    \begin{subfigure}[b]{0.47\textwidth}
        \centering
        \includegraphics[width=5.5cm]{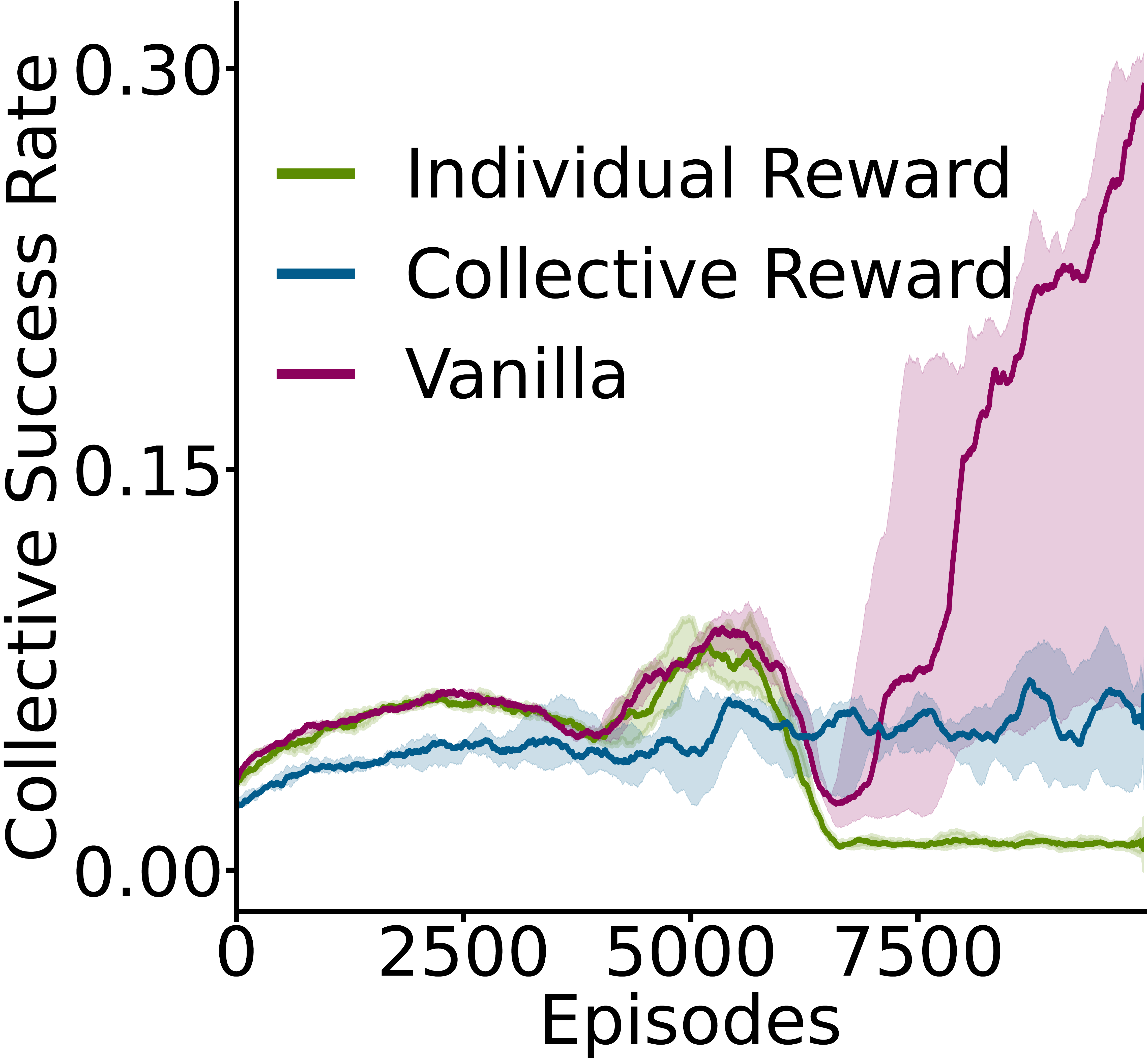}
        \caption{Collective Success Rate}
        \label{fig:sub1}
    \end{subfigure}
      \hspace{0.01\textwidth}
    \begin{subfigure}[b]{0.47\textwidth}
        \centering
       \includegraphics[width=5.5cm]{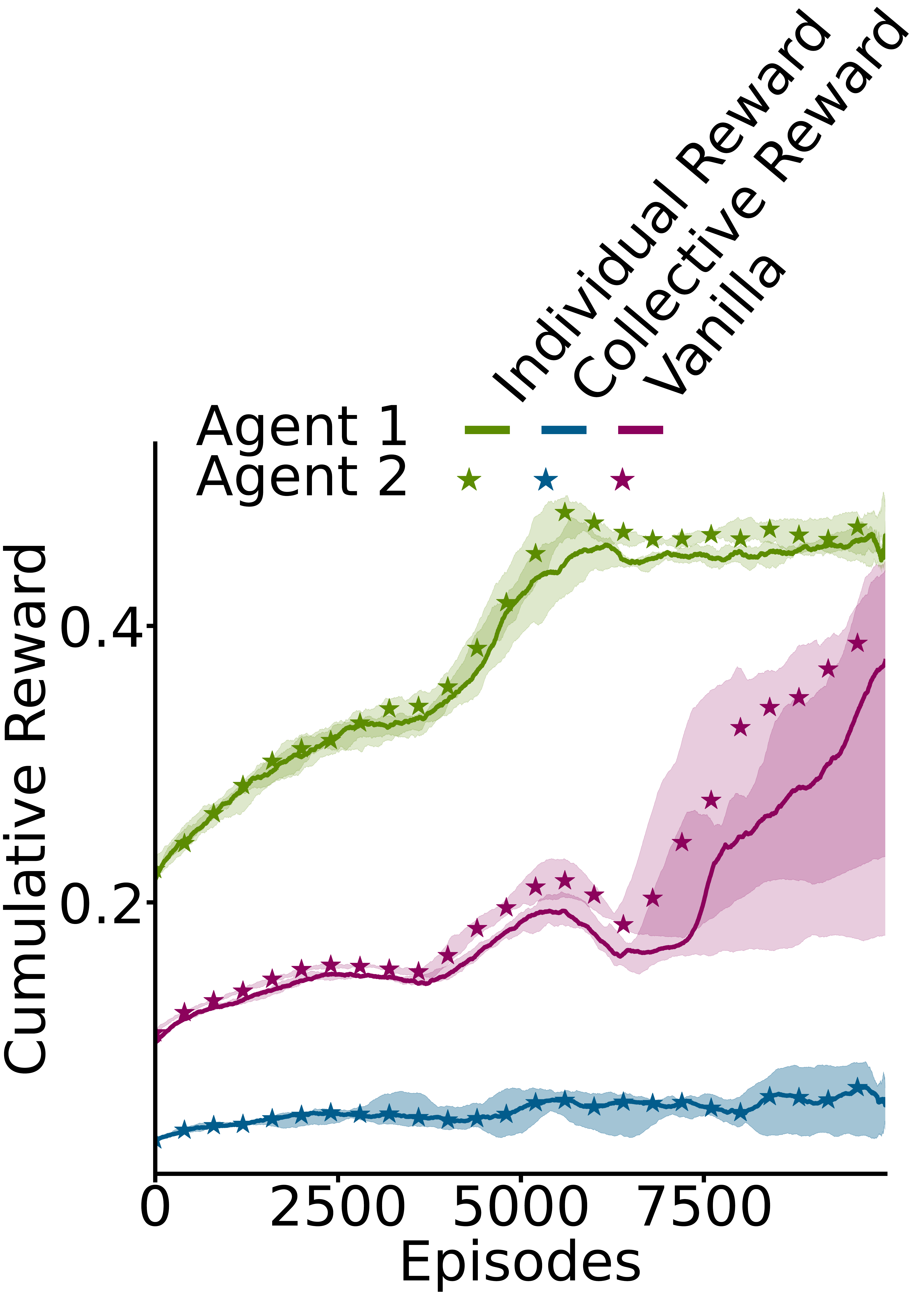}
        \caption{Cumulative Reward}
        \label{fig:sub2}
    \end{subfigure}
    \caption{a) Success rate between policy gradient agents comparing a disassociation of the reward function (i.e.. just the individual reward and the collective rewards)  b) Cumulative reward of the same dissociated reward function agents}
    \label{fig:subfigureExample}
\end{figure}

\FloatBarrier
For a further investigation of the reward function, we tested a dissociation of the individual reward $r^i$ and the collective reward $r^c$ with action history PG agents. Using only the individual reward removed collective success altogether but agents converged at a higher percentage of the cumulative reward (i.e.. whoever gets to the key first). The sole collective reward did not cause a failure in collective behavior but severely inhibited it. With both these reward dissociation, agents fail to learn hidden gifting.

\newpage

\esubsection{Minimizing self correction}\label{e:safe}
\FloatBarrier
\begin{figure}[htbp]
    \centering
    \begin{subfigure}[b]{0.48\textwidth}
        \centering
        \includegraphics[width=5.5cm]{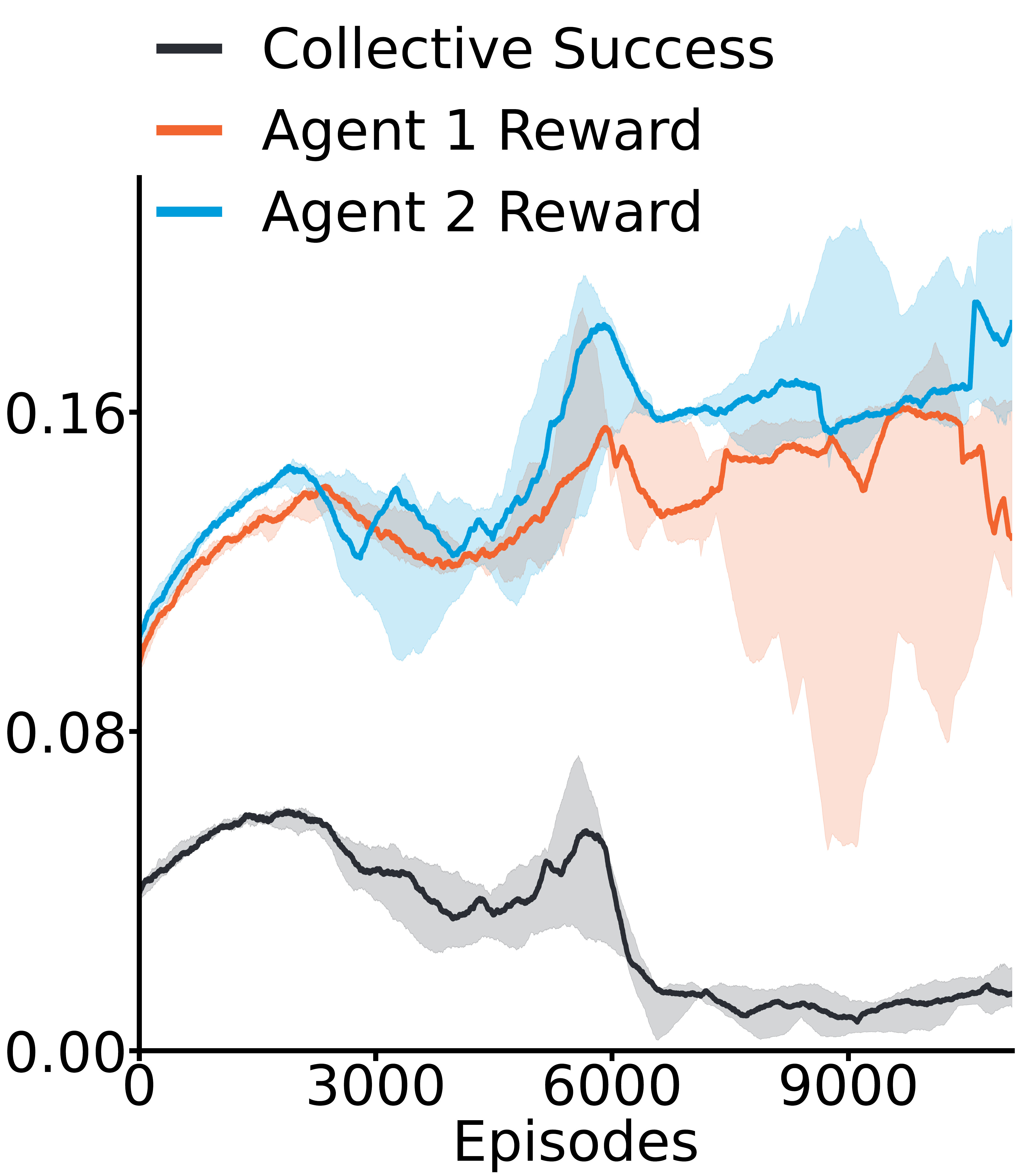}
        \caption{Collective Success Rate}
        \label{fig:anti_a}
    \end{subfigure}
      \hspace{0.01\textwidth}
    \begin{subfigure}[b]{0.48\textwidth}
        \centering
       \includegraphics[width=5.5cm]{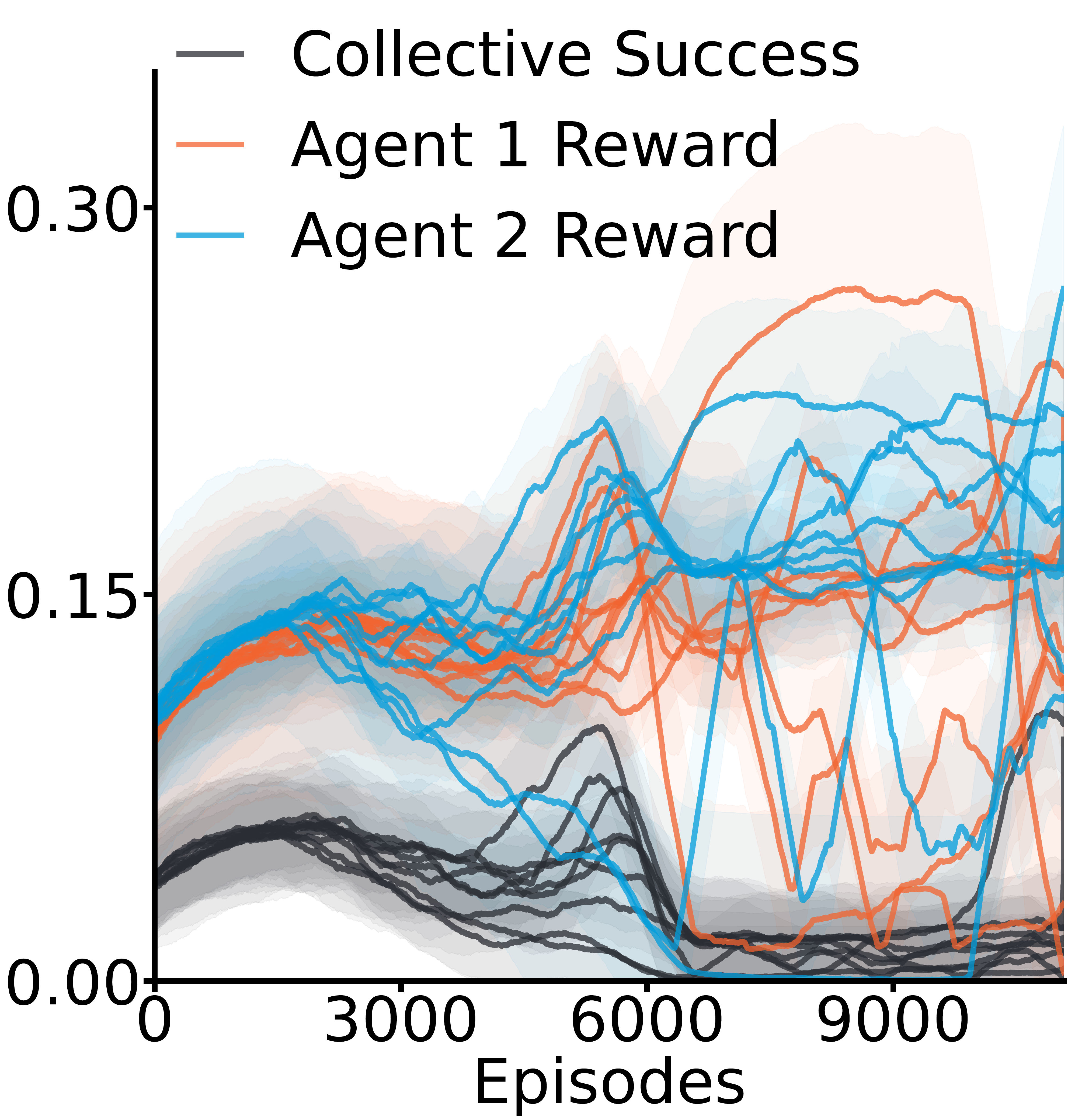}
        \caption{Collective Success Rate Across Simulations}
        \label{fig:anti_b}
    \end{subfigure}
    \caption{a) The percentage of cumulative reward and collective success for anti-collective policy gradient agents with action history (i.e. optimizing the negated self correction term) across 11000 episodes  b) 9 individual simulations for anti-collective behaviour averaged each across a different set of 32 parallel environments}
    \label{fig:anti}
\end{figure}

\FloatBarrier
For all previous experiments, the correction term was maximized to induce agents towards dropping the key for the other agent (i.e. hidden gifting). However, this term for an agent $i$ could also be minimized through negation $-\mathbb{E}[\nabla_{\Theta^i}\nabla_{\Theta^i}J_c(\Theta^i)\Psi(\pi_c^i,a^i,o^i)]$ in the policy update and doing so led agents to actively "compete" for the key and avoid dropping it all together. In Fig \ref{fig:anti}a, the rewards for both agents increases with variance spikes while the collective success rate goes down. These results demonstrate a stronger implication of the self-correction in the collective behaviour of agents than just as a variance reducer. 

Fig \ref{fig:anti}b displays the individual simulations with standard deviation of the 32 parallel environments. Specifically, the reward curves sharply drop and return after agents have learned to open their doors. This tradeoff in the individual reward accumulation is a detriment to the collective success rate but perhaps in other situations, the negative correction term can help avoid undesired rewarded behaviour.

Alternatively, if we set $\Psi$ to be $\frac{1}{\mathbb{E}[\pi_c^j(a^j|o^j)]}$ instead of $\frac{1}{\mathbb{E}[\log\pi_c^j(a^j|o^j)]}$, the self-correction term is now weighted by the actual collective policy rather than it's entropy. This is refereed to as $\hat{\Psi}$. This is a plug in adjustment and did not have a theoretical motivation or derivation. The policy independence from \ref{p:correction} should still hold but there is no proof for this adjustment. However in looking at Fig. \ref{fig:anti2}, there is a different change in behaviour of the agents. 

In Fig. \ref{fig:anti_a}, the agents follow a very similar reward accumulation path but sharply drop around 3000 episodes where a slight switch in agent 1 achieving a higher reward. This happens again at a slower rate at around 7000 episodes where agent 2 accumulates more reward than the other agent and eventually surpassing agent 1 with some increase in variance until the end of the experiment. Agent 1's reward accumulation deteriorates after 8000 episodes implying that agent 2 is better at finding the key and always holding onto it. Then this happens for a final time in this experiment at 16000 episodes where the variance blows up. The negated self-correction here is inhibiting agents more sharply, perhaps due to the smaller range of values that $\pi_c(a|o)$ has than $\log \pi_c(a|o)$. The success rate is severely reduced and does not pickup in variance or on average through the remainder of the experiment.

In Fig.\ref{fig:anti_b}, all simulations are plotted with their within simulation variance. The behaviour between both agents in performance and variance is similar until 6000 episodes where agent 2 over takes agent 1 in all but three simulations. This then reverses around 16000 episodes with the majority of agent 1 simulations overtaking agent 2. The success rate simulations are near identical to eachother, showing how more impactful  $\hat{\Psi}$ is to inhibiting cooperation.

\begin{figure}[htbp]
    \centering
    \begin{subfigure}[b]{0.48\textwidth}
        \centering
        \includegraphics[width=5.5cm]{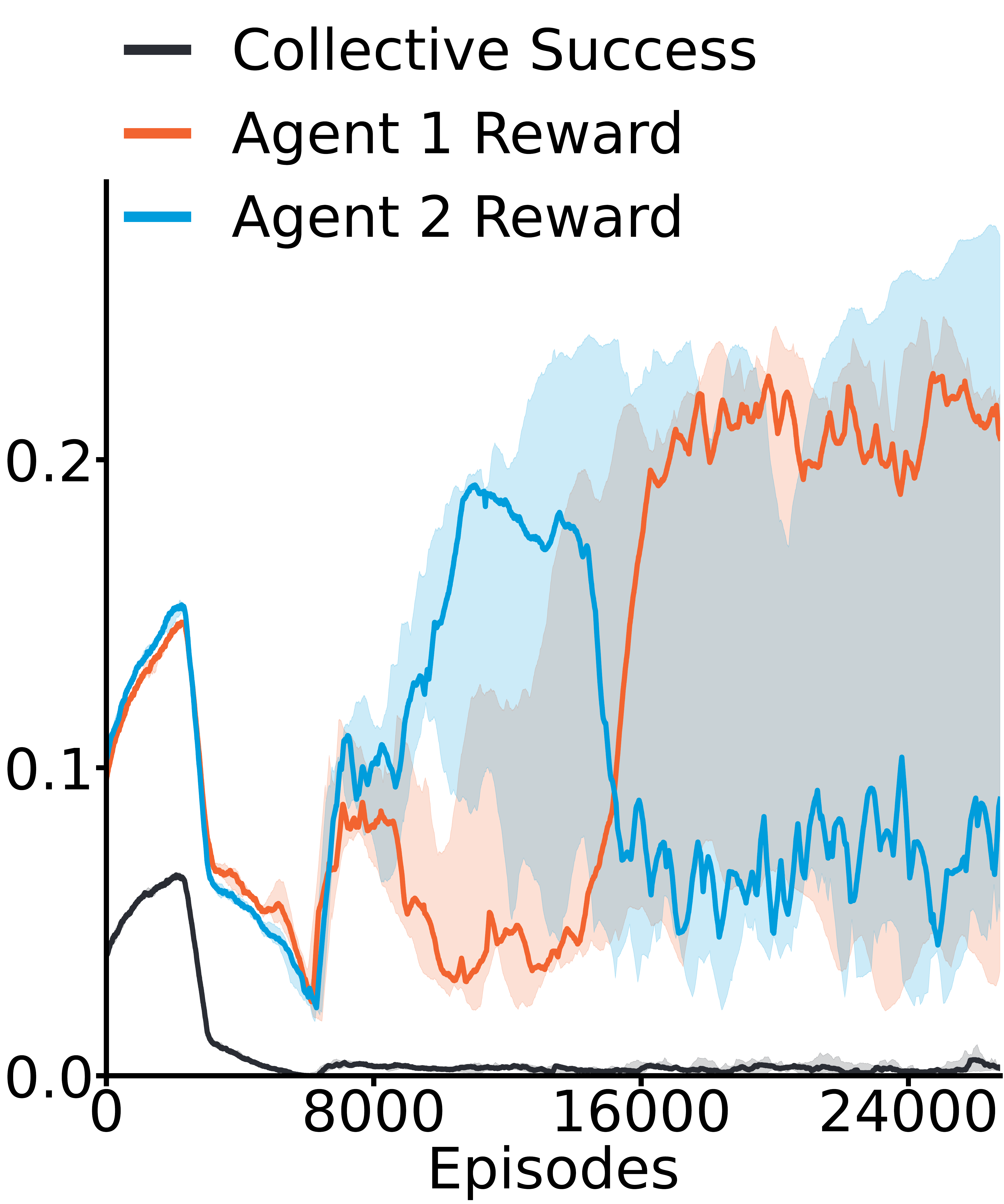}
         \caption{Collective Success Rate}
        \label{fig:anti_a}
    \end{subfigure}
      \hspace{0.01\textwidth}
    \begin{subfigure}[b]{0.48\textwidth}
        \centering
       \includegraphics[width=5.5cm]{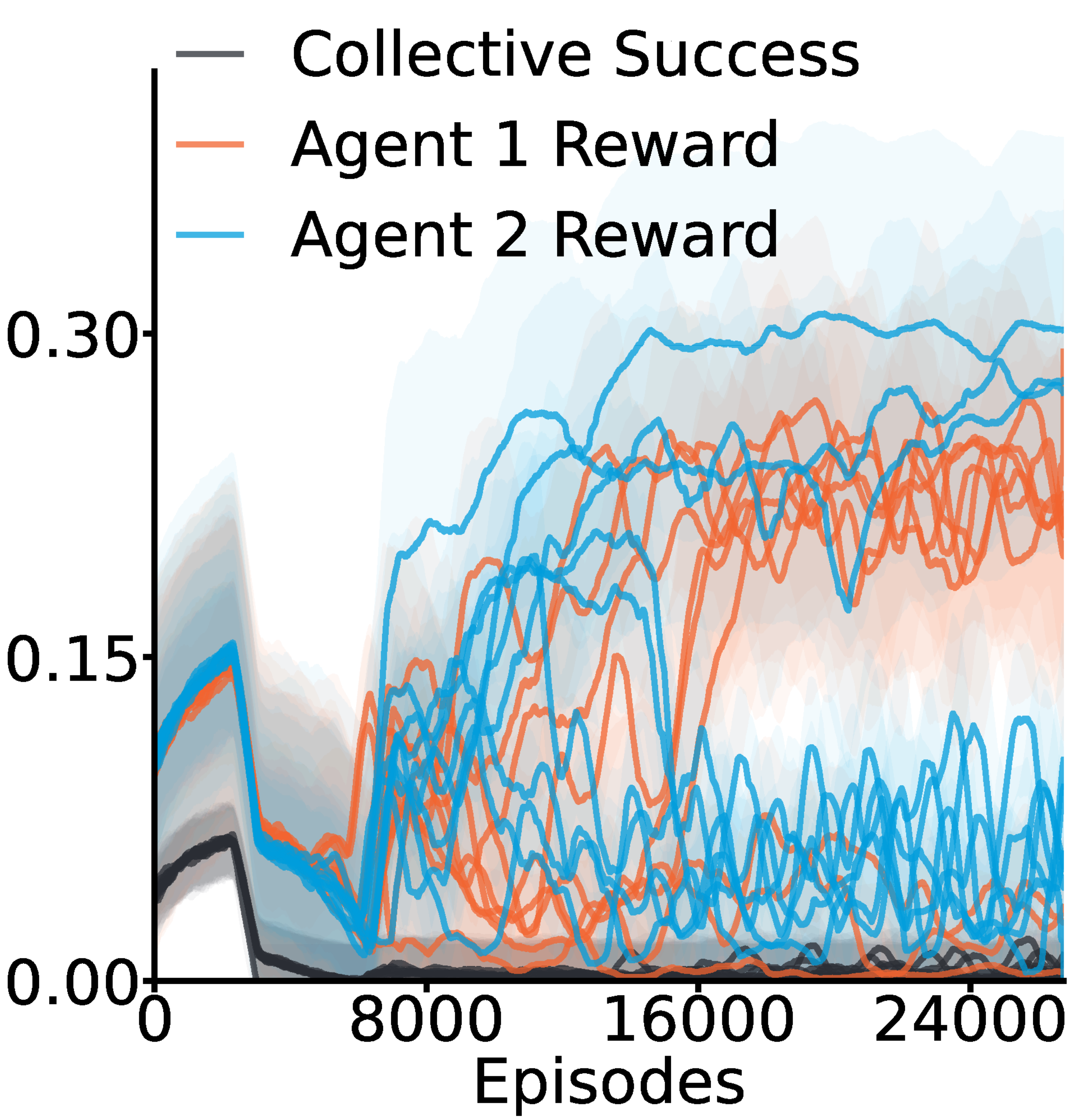}
     \caption{Collective Success Rate Across Simulations}
        \label{fig:anti_b}
    \end{subfigure}
    \caption{a) The percentage of cumulative reward and collective success for anti-collective policy gradient agents with action history (i.e. optimizing the negated self correction term) across 11000 episodes  b) 9 individual simulations for anti-collective behaviour averaged each across a different set of 32 parallel environments}
    \label{fig:anti}
\end{figure}

\FloatBarrier

The original self-correction contraposition is theoretically motivated and showed a more smoother slower inhibition on average where agents continued to compete with similar performance until almost the end of the experiment. The adjusted self-correction inhibition has a more sharper effect which makes sense since the policy is a categorical distribution. The agents do not compete comparatively though. Agent 2 overtakes agent 1 earlier than in the first inhibition experiment. Overall, these experiments further implicate self-correction in learning the collective sub policy for leaving hidden gifts for the other agent.

\newpage
\esubsection{The policy gradient objective is better than the q-learning in single agent key-to-door} \label{e:single_agents}
\FloatBarrier
\begin{figure}[htbp]
    \centering
 
        \includegraphics[width=5.5cm]{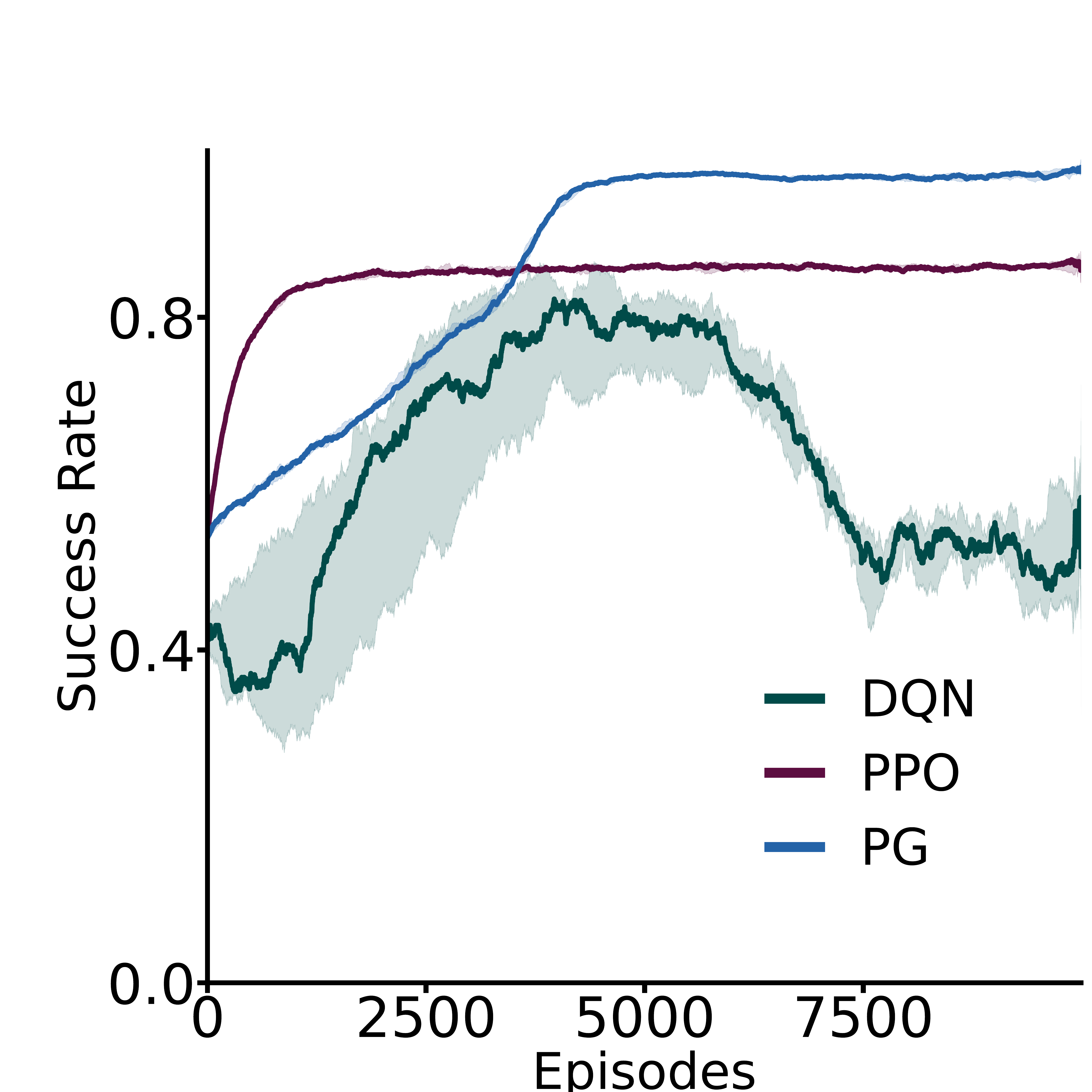}

    \caption{a) Comparison of single agent PPO, PG and DQN agents where one agent needs to open a one door after finding one key}

\end{figure}

As a baseline, PPO, PG and DQN agents are compared on the individual objective of the main task (eg. opening a door). This is a normal key-to-door task and success is defined by opening a door for a sparse reward of 1. PPO and PG agents retain the same hyperparameter except the learning rate for both actor and critic in PPO was reduced after a grid search to tune against overfitting. The DQN agent required 1 simulation at a time rather than 32 in parallel but was not able to converge above 50\%  success after an extensive hyperparameter search. This demonstrates the performance of on-policy policy gradient objective over the off-policy q-learning objective in temporal credit assignment tasks.

\newpage
\esubsection{A self correcting naive agent is more stable  than a naive LOLA agent without a second learning rate} \label{e:lola_fail}
\FloatBarrier

\begin{figure}[htbp]
    \centering
    \begin{subfigure}[b]{0.48\textwidth}
        \centering
        \includegraphics[width=4.5cm]{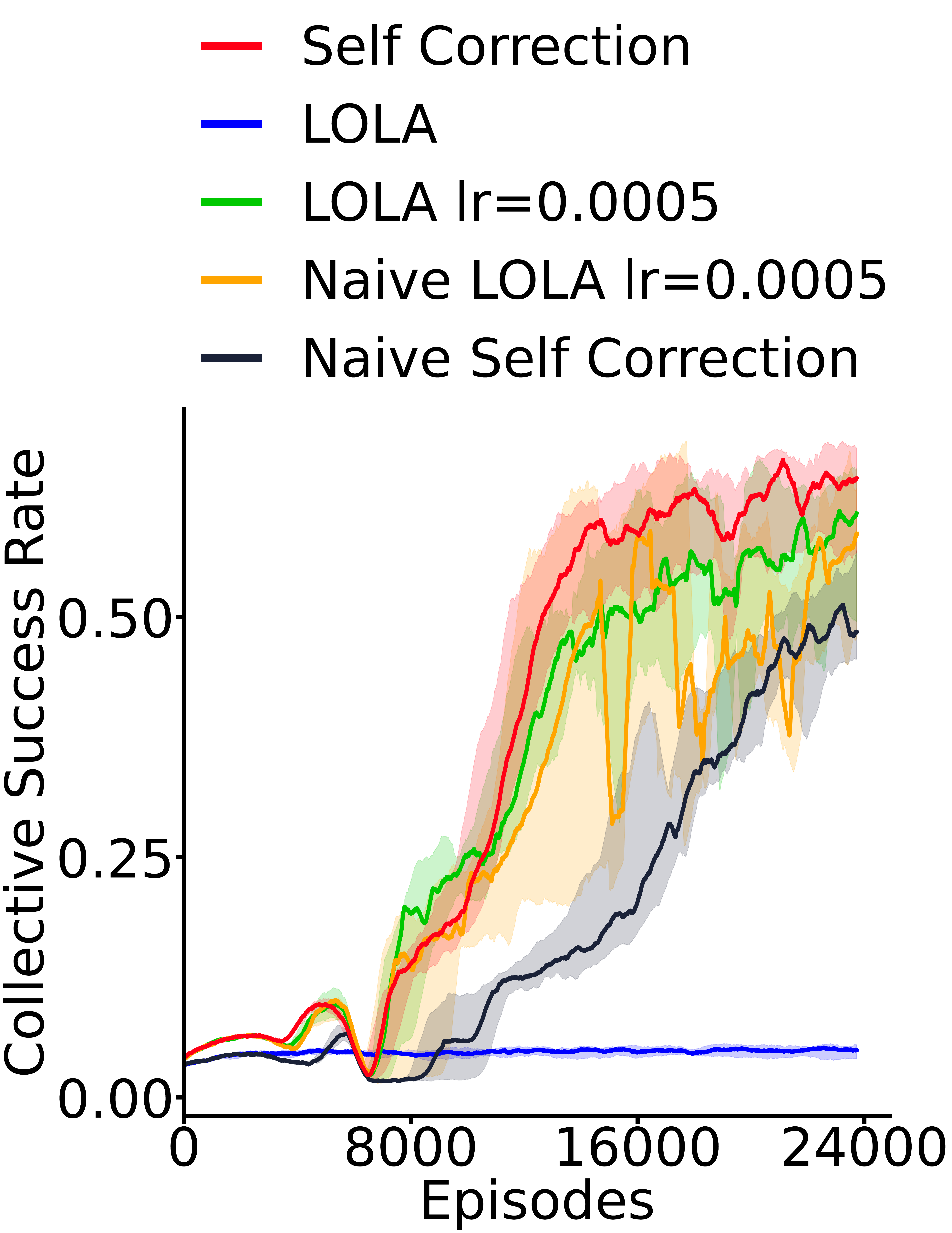}
        \caption{Collective Success Rate}
    \end{subfigure}
    \begin{subfigure}[b]{0.48\textwidth}
        \centering
       \includegraphics[width=5cm]{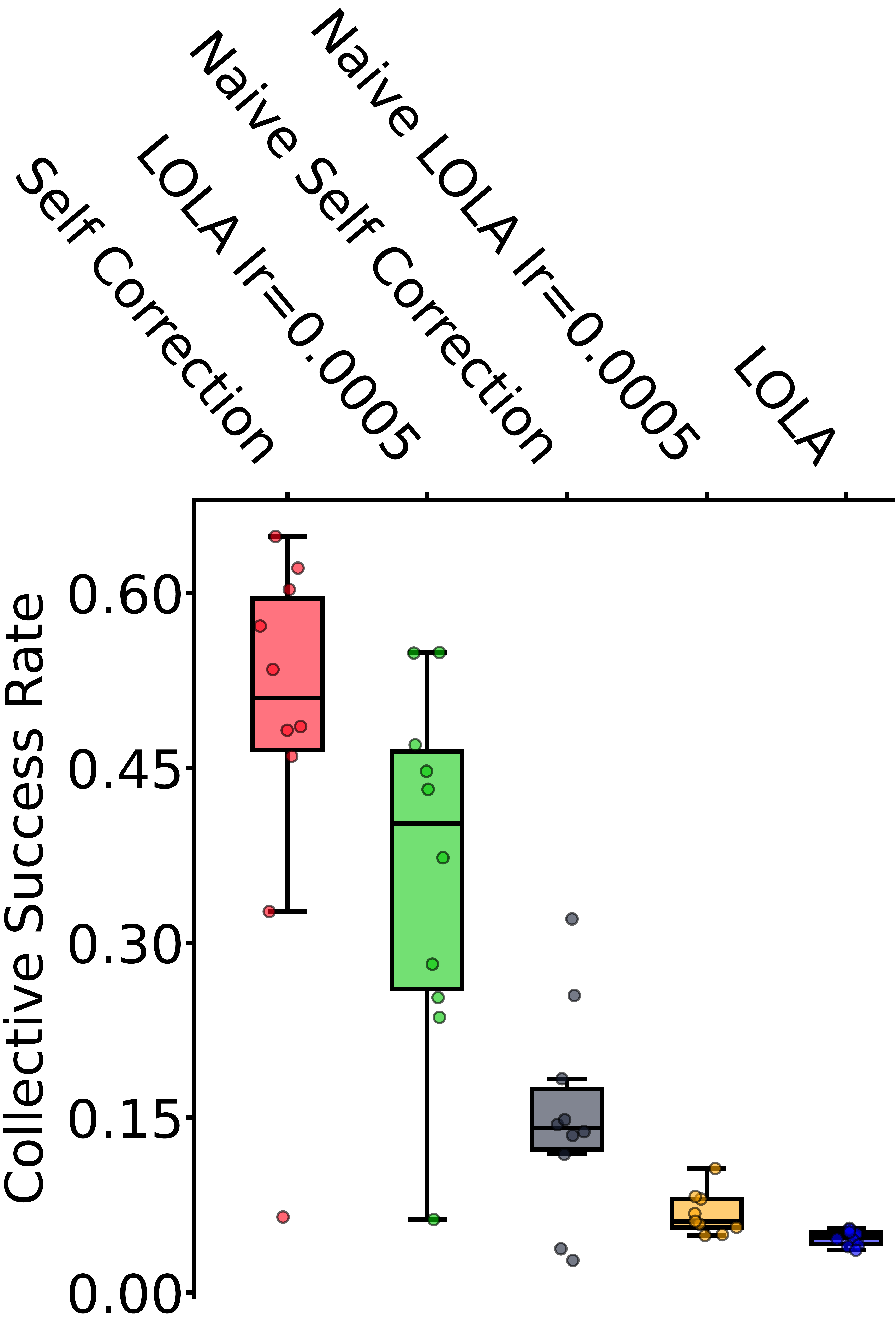}
        \caption{Global Collective Success}
    \end{subfigure}
    \caption{a) A comparison of LOLA with a learning rate of 1 and 0.0005, Naive LOLA with a learning rate of 0.0005, Self-correction, and Naive Self-correction. The naive learner framework only permits one agent to optimize the additional hessian objective rather than two. b)}
\end{figure}

Learning with Opponent Learning Awareness (LOLA) \citep{foerster2017learning} is the original learning aware gradient update. In the original work, only one policy gradient agent was a LOLA agent with the other being a naive policy gradient agent. To test how self-correction compares, 32 parallel simulations were ran for  LOLA with a learning rate of 1 and 0.0005, Naive LOLA with a learning rate of 0.0005, Self-correction, and Naive Self-correction. LOLA with a learning rate of 1 did not learn cooperative behaviour but decreasing the learning rate to 0.0005 improved learning but with high ossicilations in median performance as well as variance. When both agents were LOLA agents, similar to \citep{willi2022cola}, there was greater stability in collective success but less than self-correction. For thoroughness, a naive learner experiment for self-correction was ran. Interestingly, the variance reduction effect was maintained but performance was delayed and reduced compared to self-correction. However, this implies that one self-correction agent can help stabilize collective success.

\newpage
\esubsection{One last action is sufficent for policy gradient agents to solve the Manitokan task} \label{e:more_acts}
\FloatBarrier

\begin{figure}[htbp]
    \centering
    \begin{subfigure}{0.48\textwidth}
        \centering
        \includegraphics[width=4cm]{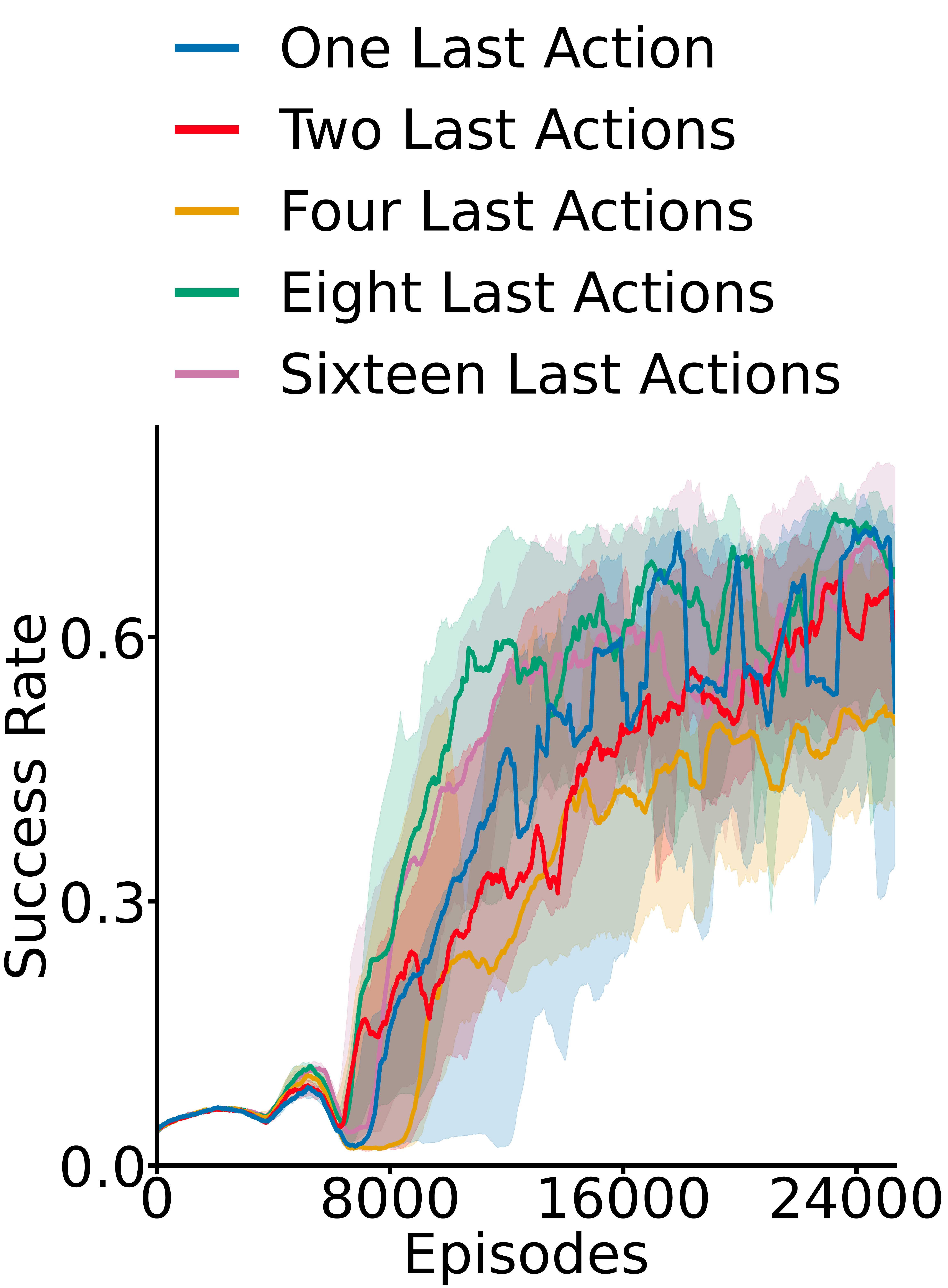}
        \caption{Collective Success Rate}
    \end{subfigure}
      \centering
    \begin{subfigure}{0.48\textwidth}
        \centering
       \includegraphics[width=4.5cm]{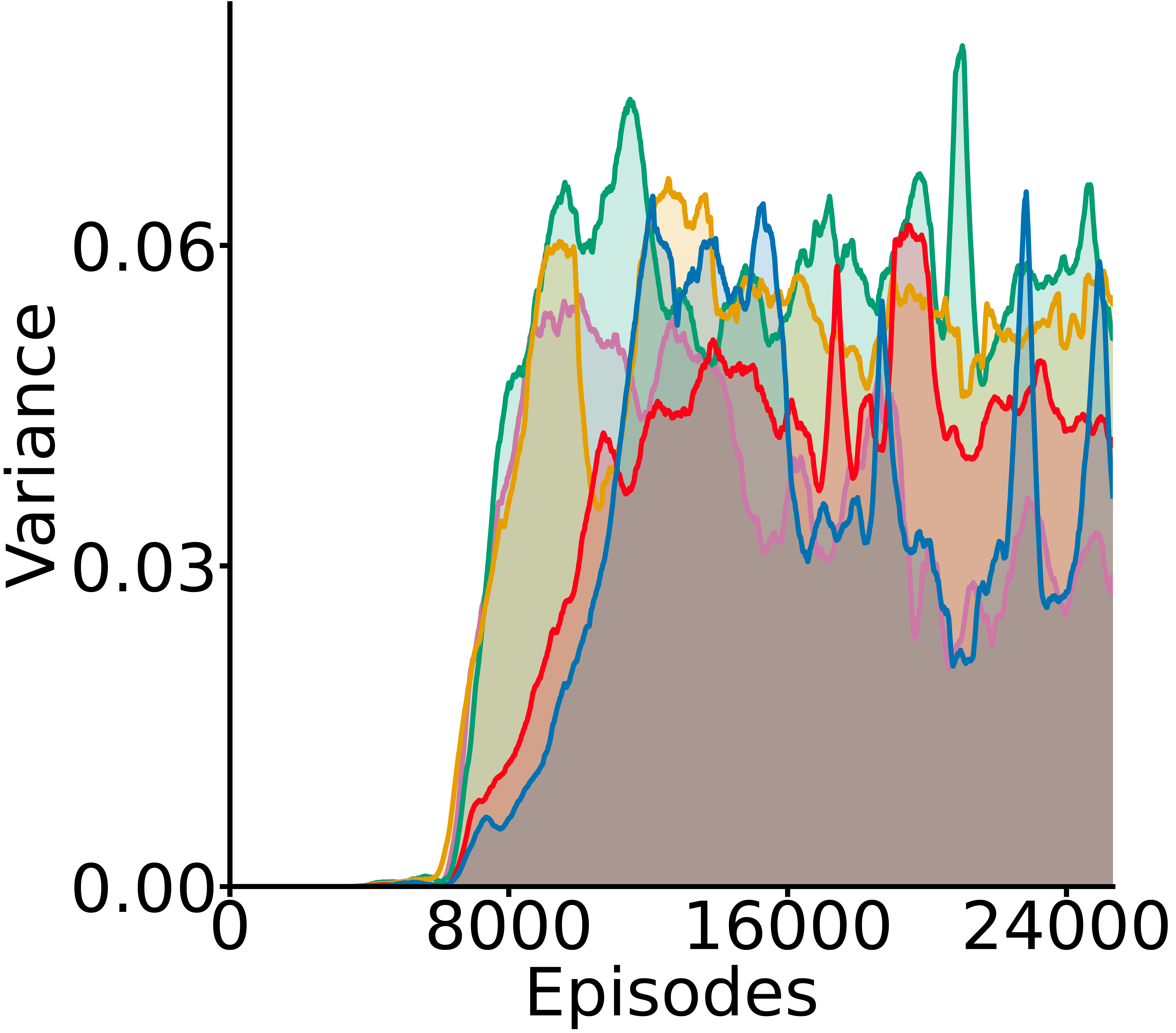}
       \caption{Variance}
    \end{subfigure}
        \begin{subfigure}{0.48\textwidth}
        \centering
       \includegraphics[width=4.5cm]{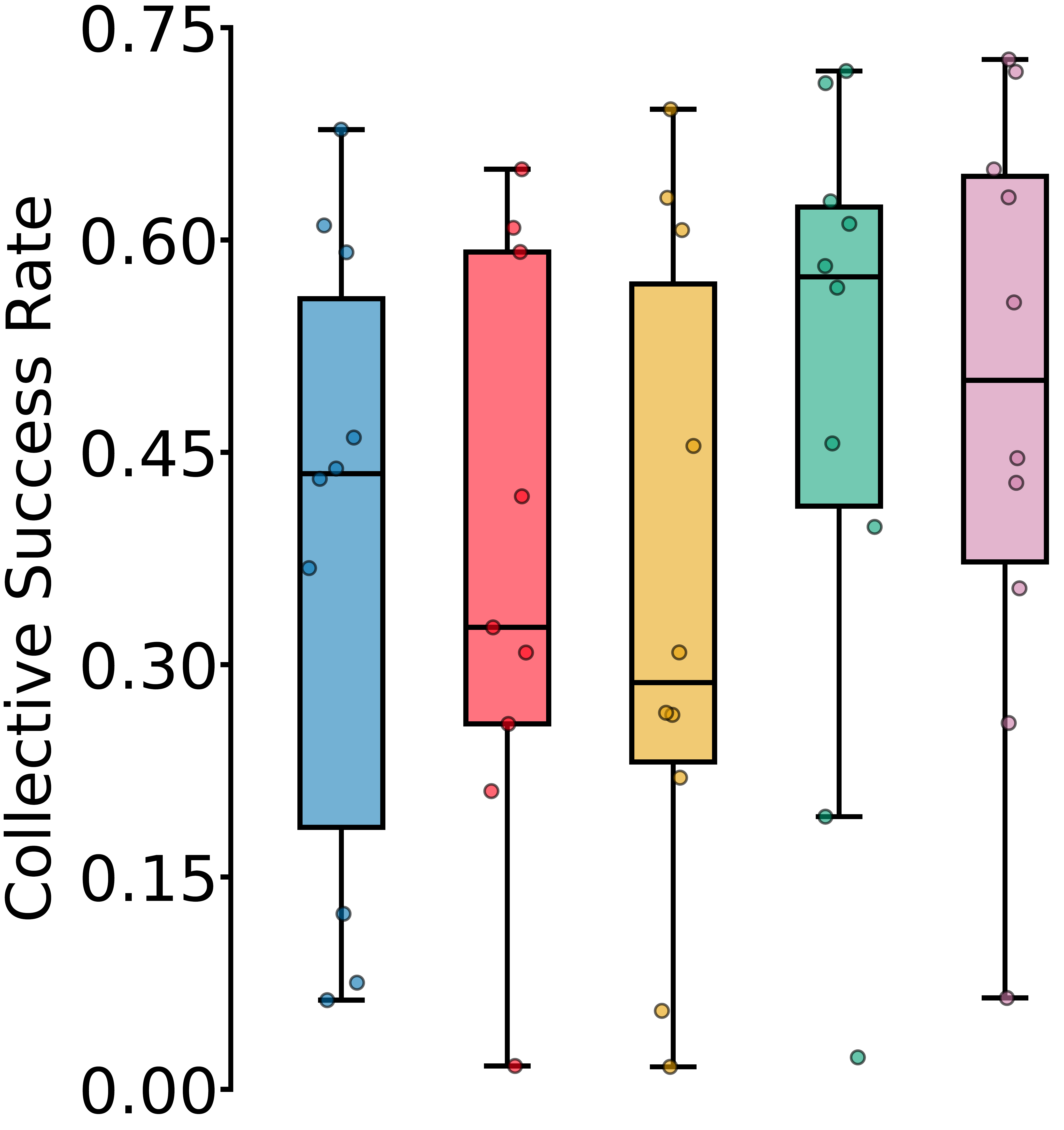}
       \caption{Global Collective Success}
    \end{subfigure}
      \centering
    \begin{subfigure}{0.48\textwidth}
        \centering
       \includegraphics[width=5.cm]{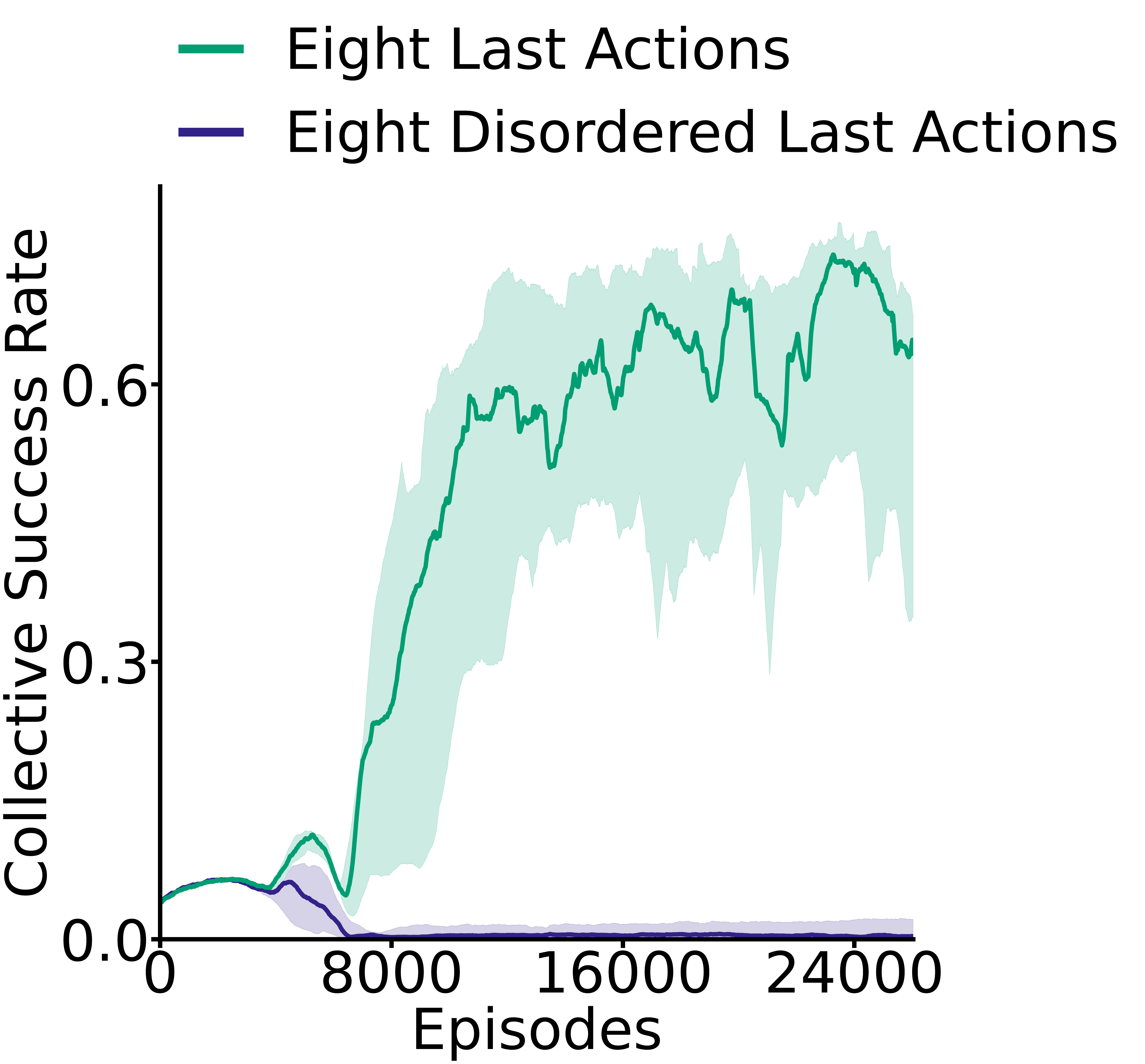}
       \caption{Disorderd Actions Collective Success Rate}
    \end{subfigure}
    \caption{a) A comparison of collective success rate between PG agents with last action inputs with one last action, two last actions, four last actions, eight last actions and sixteen last actins b) A comparison of variance over time between PG agents with last action inputs c) A comparison of global collective success rate between PG agents with last action inputs d) A comparison between PG agents with eight last actions as input and PG agents with randomly permutated eight last actions as input.}
    \label{fig:m_last_acts}

\end{figure}

Due to the performance of one last action, we were curious if more last actions could further increase the collective success rate. Two and Four last actions inhibit performance, while eight and sixteen last actions recover performance comparable to one last action. These results indicate that a single previous action provides a great deal of the necessary information to solve the task, though more history could also help. There did not seem to be a large effect on variance over time; however, sixteen last actions did exhibit a reduction. The importance of temporal structure rather than quantity of information is demonstrated in \cref{fig:m_last_acts}d where randomly permutating or disordering 8 last actions nearly brings collective success rate to zero.
\newpage
\esubsection{The Self Correction value is more correlated to collective success than policy entropy in maximum entropy policy agents} \label{e:corr_self}
\FloatBarrier

\begin{figure}[htbp]
    \centering
    \begin{subfigure}{0.48\textwidth}
        \centering
        \includegraphics[width=5.5cm]{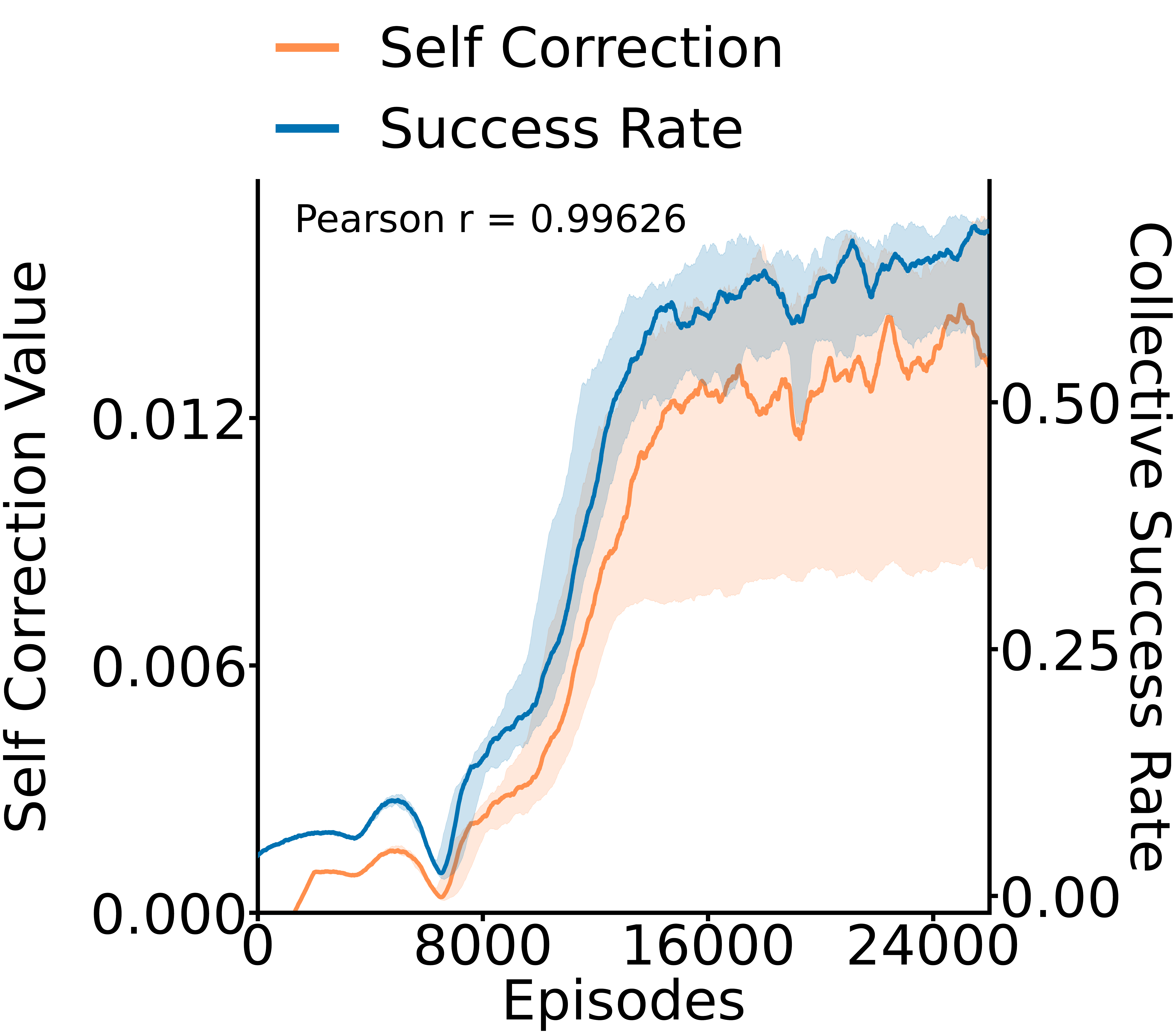}
        \caption{Self Correction Correlation}
    \end{subfigure}
    \begin{subfigure}{0.48\textwidth}
        \centering
       \includegraphics[width=5.5cm]{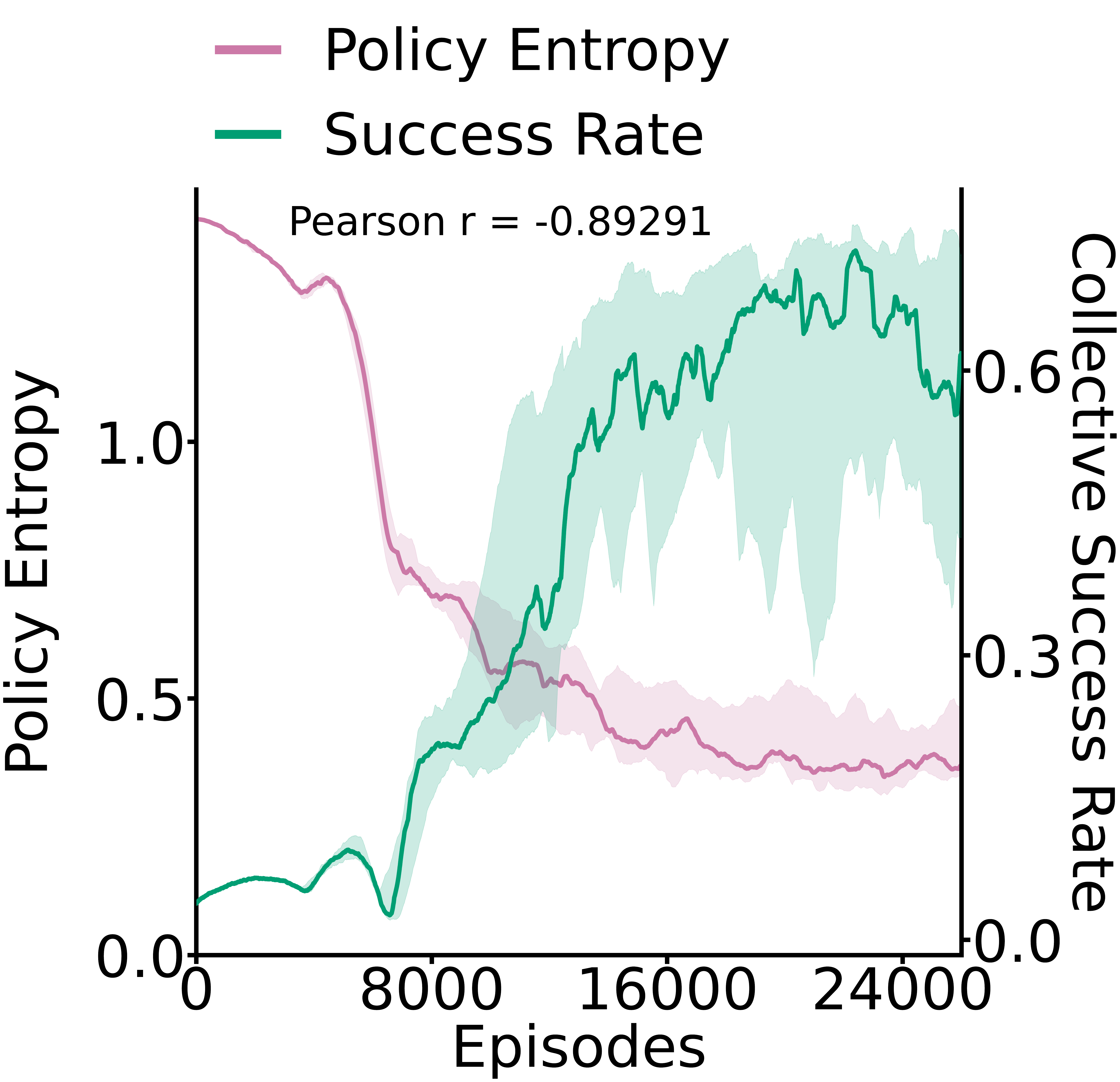}
       \caption{Policy Entropy Correlation}
    \end{subfigure}
    \caption{a) A correlation analysis between the collective success rate rate of PG agents with one last action input and a self correction term, and the value of the self-correction term. b) A correlation analysis between the collective success rate rate of PG agents with one last action input and a maximum entropy term, and the value the value of policy entropy.}
    \label{fig:term_values}

\end{figure}

In \cref{fig:term_values}a, we tested  the relationship between the value of the self correction term and our performance metric  of collective success rate. The self correction term is highly correlated with success, since Pearson's r = 0.99626. The variance in the self-correction term is noticeably larger than the success rate. While in In \cref{fig:term_values}b we did the same test between the policy entropy values and the collective success rate of the maximum entropy PG agents. The entropy is inversely correlated with the collective success rate but not as strongly than as the self correction values with a Pearson's r = -0.89291. The variance in policy entropy is markedly low.

\FloatBarrier
\newpage
\FloatBarrier
\psection{}

\psubsection{Correction term}
\label{p:correction}
\FloatBarrier

We begin by deriving the standard policy gradient theorem \citep{sutton1998reinforcement,sutton1999policy} under the assumptions in Section 4 that the collective reward was returned and that the collective reward $r^c$ is differentiable through another agent $j$s objective. The objective $J(\Theta^i)$ for agent $i$ is to maximize the expected cumulative sum of rewards within an episode $\mathbb{E}[\sum_t^T\gamma^t\mathcal{R}^i(o^i_t,a^i_t)]$ with the reward function $\mathcal{\hat{R}}$ in equation 1 where a value function $V(\Theta^i,o^i) = \mathbb{E}[\gamma^t\mathcal{\hat{R}}^i(o^i,a^i)]$. 
\\

\begin{equation}\label{eq6}\nabla_{\Theta^i}J(\Theta^i)=\nabla_{\Theta^i} (\sum_{a^i\in A}\pi^i(a^i|o^i)Q(o^i,a^i))\end{equation} is the differentiated objective with respect to agent $i$. 
\\

\begin{equation}\label{eq7}\sum_{a^i\in A}(\nabla_{\Theta^i}\pi^i(a^i|o^i)Q(o^i,a^i)+\pi^i(a^i|o^i)\nabla_{\Theta^i}Q(o^i,a^i))\end{equation} by product rule expansion.
\\

\begin{equation}\label{eq:8}\sum_{a^i\in A}\nabla_{\Theta^i}\pi^i(a^i|o^i)Q(o^i,a^i)+\pi^i(a^i|o^i)\nabla_{\Theta^i}(\sum_{o^{i'},R^i}\mathcal{T}(o^{i'},r^i+r^c|o^i,a^i)(r^i+r^c+V(\Theta^i,o^{i'})))\end{equation}Here, \cref{eq:8} is summed over all actions $\sum_{a^i\in A}$. In particular, the value function can be used to predict a look-ahead of the next collective reward $r^c$ with a next observation $o^{i'}$ and $\mathcal{T}$ is the transition probability.
\\

\begin{equation}\label{eq:8}\sum_{a^i\in A}\nabla_{\Theta^i}\pi^i(a^i|o^i)Q(o^i,a^i)+\pi^i(a^i|o^i)(\sum_{o^{i'},R^i}\mathcal{T}(o^{i'},r^i+r^c|o^i,a^i)(\nabla_{\Theta^i}r^i+\textcolor{red}{\nabla_{\Theta^i}r^c}+\nabla_{\Theta^i}V(\Theta^i,o^{i'})))\end{equation}

Notably, $\nabla_{\Theta^i}r^i=0$ since the reward does not require another agent but $\textcolor{red}{\nabla_{\Theta^i}r^c}\not=0$ since the other agents are required  this reward and therefore $\textcolor{red}{\nabla_{\Theta^i}(\nabla_{\Theta^j}r^c)}$ the reward is implicitly conditioned by other agents policies which is non-stationary and changes across episodes. To obtain a better estimate of the collective reward that accounts for the non-stationarity in the other agent's policy, we need a \emph{estimate} of the collective reward.

To isolate the sub-objective for the collective policy, start with the reward maximization objection.
\\

\begin{equation}\label{eq:10}J(\Theta^j)=\mathbb{E}_{\tau\sim \pi^j, \pi^j}[\sum_t^T\gamma^t\mathcal{\hat{R}}^j(o^j_t,a^j_t)]\end{equation}
\\

The individual reward $r^i$ and collective reward $r^c$ are single scalar rewards and can be factorized,

\begin{equation}\label{eq:11}J(\Theta^j) 
     = \mathbb{E}_{\tau\sim \pi^i, \pi^j}[\sum_{t=0}^T \gamma^tr_t^j+ \gamma^tr^c]\end{equation} by linearity in \cref{eq:2} of $\mathcal{\hat{R}}^j$.
\\

The terms can be separated,

\begin{equation}
    J(\Theta^j)= \mathbb{E}_{\tau\sim \pi^i, \pi^j}[\sum_{t=0}^{T-1} \gamma^tr_t^j] + \mathbb{E}_{\tau\sim \pi^i, \pi^j}[\sum_{t=T}^T \gamma^tr^c]
\end{equation}

but the collective reward can only appear at the terminal step of an episode, 

\begin{equation}
    J(\Theta^j)= \mathbb{E}_{\tau\sim \pi^i, \pi^j}[\sum_{t=0}^{T-1} \gamma^tr_t^j] + \mathbb{E}_{\tau\sim \pi^i, \pi^j}[ \gamma^Tr^c] 
\end{equation}

which leads to two sub-objectives for the rewards,

\begin{equation}
    J(\Theta^j)= J_d(\Theta^j) + J_c(\Theta^j)
\end{equation}

From here, we have an isolated sub-objective assuming the individual objective can be solved,

\begin{equation}
     J_c(\Theta^j) = J(\Theta^j) - J_d(\Theta^j) = \mathbb{E}_{\tau\sim \pi^i, \pi^j}\!\left[\sum_{t=0}^{T-1} \gamma^t r^{c,i}\right] = \mathbb{E}_{\tau\sim \pi^j, \pi^j}\!\left[\gamma^T r^c\right]
\end{equation}

Now, we assume an estimator of the collective objective $\hat{J_c}(\Theta^j)=\mathbb{E}_{\tau\sim\pi^j}\big[\log \pi_c^i(a^i|o^i)\,
      r^c_T\big]$ since the individual reward was acquired. Differentiating $\hat{J_c}(\Theta^j)$,

Differentiating agent $i$'s collective objective with respect to agent $j$ is

\begin{equation}\label{eq:other}
\nabla_{\Theta^j} \hat{J_c}(\Theta^j)=\mathbb{E}_{\tau\sim\pi^j}\big[\nabla_{\Theta^j}\log \pi_c^i(a^i|o^i)\,
      Q^j_c(o^j,a^j)\big] = \mathbb{E}_{\tau\sim\pi^j}\big[\log \pi_c^i(a^i|o^i)\,
      \nabla_{\Theta^j}Q^j_c(o^j,a^j)\big] 
\end{equation}

where $Q^j_c(o^j,a^j)$ is the collective $Q$ value. 

Now, since the policies between agent $i$ and agent $j$ are statistically independent, 

\begin{equation}\label{eq:other}\nabla_{\Theta^j}\hat{J_c}(\Theta^j)=\mathbb{E}_{\tau\sim\pi^j}[\log \pi_c^i(a^i|o^i)]\mathbb{E}_{\tau\sim\pi^j}[ \nabla_{\Theta^j}Q^j_c(o^j,a^j)] \end{equation} \\

By dividing the terms by $\mathbb{E}_{\tau\sim\pi^j}\big[\log \pi_c^i(a^i|o^i)\big]$ which is a scalar quantity and equivalent to the entropy of the policy at the terminal state $\mathbb{E}_{\tau\sim\pi^j}\big[\log \pi_c^j(a^j|o^j)\big] = \sum_{a^j}^{|\mathcal{A}|}\pi_c^j(a^j|o^j)\log \pi_c^j(a^j|o^j)=\mathbb{H}\big[\pi_c^j(.|o^j)\big]$, we get a surrogate for the change in Q value estimate. Let $\Psi(\pi^i_c,o^i,a^i) = \frac{1}{\mathbb{E}_{\tau\sim\pi^j}[\log \pi_c^i(a^i|o^i)]}$ where $\Psi$ is the reciprocal of the expected collective policy for agent $i$ which is a scalar in the terminal state. 

\begin{equation}\label{eq:17}
\textcolor{red}{\nabla_{\Theta^i}Q^j_c(o^j,a^j)}= \mathbb{E}_{\tau\sim\pi^j}[\frac{\nabla_{\Theta^j}\hat{J_c}(\Theta^j)}{\log \pi_c^i(a^i|o^i)}]  = \mathbb{E}_{\tau\sim\pi^j}[\nabla_{\Theta^j}\hat{J_c}(\Theta^j)\Psi(\pi_c^i,a^i,o^i)]= \nabla_{\Theta^j}J_c(\Theta^j)
\end{equation}

Then plugging  \cref{eq:17} as a  into \cref{eq:8},

\begin{equation}\label{eq:16}\begin{split}\sum_{a^i\in A}\pi^i(a^i|o^i)(\sum_{o^{i'},\mathcal{\hat{R}}^i}\mathcal{T}(o^{i'}_{+1},\mathcal{\hat{R}}^i(o^i,a^i)|o^i,a^i)( \nabla_{\Theta^i}\nabla_{\Theta^j}J_c(\Theta^j)\Psi(\pi_c^i,a^i,o^i)  +\nabla_{\Theta^i}V(\Theta^i,o^{i'}))\end{split}\end{equation} Now in \cref{eq:16} the correction term as a surrogate for the collective reward in the look ahead step from \cref{eq:8}. The remainder of this "proof" is complete the derivation of the policy gradient theorem by transitioning over all next observations and actions.
\\

Let $\Phi(o^i) = \sum_{a^i\in A}(\nabla_{\Theta^i}\pi^i(a^i|o^i)Q(o^i,a^i)$ for readability and Let $\rho^{i}(o^i\rightarrow o^{i'})=\pi^i(a^i|o^i)(\sum_{o^{i'},\mathcal{\hat{R}}^i}\mathcal{T}(o^{i'},\mathcal{\hat{R}}^i(o^i,a^i)|o^i,a^i)$ for further readability.
\\

\begin{equation}\label{eq:17}\Phi(o^i)+\sum_{o^i}\rho^{i}(o^i\rightarrow o^{i}_{+1})(\nabla_{\Theta^i}V(\Theta^i, o^i_{+1})+\nabla_{\Theta^i}\nabla_{\Theta^j}J_c(\Theta^j)\Psi(\pi_c^i,a^i,o^i))\end{equation}
\\

The previous, \cref{eq:17}, can then be recursively expanded out further $\Phi(o^i)+\sum_{o^i}\rho^{i}(o^i\rightarrow o^{i}_{+1})(\Phi(o^{i}_{+1}) +\nabla_{\Theta^i}\nabla_{\Theta^j}J_c(\Theta^j)\Psi(\pi_c^i,a^i,o^i)+\sum_{o^{i}_{+1}}\rho^{i}(o^{j}_{+1}\rightarrow o^{j}_{+2})(\nabla_{\Theta^i}V(\Theta^i,o^{+2})+\nabla_{\Theta^i}\nabla_{\Theta^j}J_c(\Theta^j)\Psi(\pi_c^i,a^i,o^i))$
\\

\begin{equation}\label{eq:18}\sum_{x^i,x^j\in O}\sum_{k=0}^\infty \rho^{i}(o\rightarrow x^i, k)(\Phi(x^i) +\nabla_{\Theta^i}\nabla_{\Theta^j}J_c(\Theta^j)\Psi(\pi_c^i,a^i,x^i))\end{equation}
\\

Let $\eta(o) =\sum_{\nabla_{\Theta^i}\nabla_{\Theta^j}J_c(\Theta^j)\Psi(\pi_c^i,a^i,o^i),k=0}^\infty \rho^{i}(o^i\rightarrow o^{i'}, k) $ to clarify the transitions.
\\

\begin{equation}\label{eq:19}\sum_o\eta(o)(\Phi(o) +\nabla_{\Theta^i}\nabla_{\Theta^j}J_c(\Theta^j)\Psi(\pi_c^i,a^i,o^i)) \propto \sum_o\frac{\eta(o)}{\sum_o \eta(o)}(\Phi(o) +\nabla_{\Theta^i}\nabla_{\Theta^j}J_c(\Theta^j)\Psi(\pi_c^i,a^i,o^i)\end{equation} since the normalized distribution is a factor of the sum.
\\

Then let $\sum_s\frac{\eta(o)}{\sum_o \eta(o)} = \sum_{o\in O} d(o)$
\\

\begin{equation}\label{eq:20}\sum_{o\in O} d(o)(\sum_{a^i\in A}(\nabla_{\Theta^i}\pi^i(a^i|o^i)Q(o^i,a^i) + \nabla_{\Theta^i}\nabla_{\Theta^j}J_c(\Theta^j)\Psi(\pi_c^i,a^i,o^i)) \end{equation}
\\

\begin{equation}\label{eq:21}\sum_{o\in O} d(o)(\sum_{a^i\in A}(\pi^i(a^i|o^i)Q(o^i,a^i)\frac{\nabla_{\Theta^i}\pi^i(a^i|o^i)}{\pi^i(a^i|o^i)} + \nabla_{\Theta^i}\nabla_{\Theta^j}J_c(\Theta^j)\Psi(\pi_c^i,a^i,o^i)) \end{equation}, the log-derivative trick can pull out the gradient.
\\

\begin{equation}\label{eq:22}\sum_{o\in O} d(o)(\sum_{a^i\in A}((a^i|o^i)Q(o^i,a^i)\nabla_{\Theta^i} \log {\pi^i(a^i|o^i)} + \nabla_{\Theta^i}\nabla_{\Theta^j}J_c(\Theta^j)\Psi(\pi_c^i,a^i,o^i)) \end{equation}
\\

Finally, the full gradient objective from \cref{eq:5} is constructed $$\nabla_{\Theta^i}J(\Theta^i)=\mathbb{E}_{\tau\sim\pi^i,\tau\sim\pi^j}[Q(o^i,a^i)\nabla_{\Theta^i}\log{\pi^i(a^i|o^i)} + \nabla_{\Theta^i}\nabla_{\Theta^j}J_c(\Theta^j)\Psi(\pi_c^i,a^i,o^i))]\text{  }\square$$

\psubsection{Correction terms do not conflict with individual objectives}\label{p:conf}

A corollary to the construction of the correction term is that if there is no collective reward signal (ex. the agent is performing a single agent task), then the correction degenerates to zero. 

For the sake of contradiction, assume that the correction term does not become zero when there is a lack of a collective reward signal such that there exists a value $b\not =0$. Then,

\begin{equation}\label{eq:27}b=\frac{\nabla_{\Theta^j}J_c(\Theta^i)}{\mathbb{E}_{\tau\sim\pi^j}[\log \pi_c^j(a^j|o^j)]}\end{equation}

\begin{equation}\label{eq:28}\mathbb{E}_{\tau\sim\pi^i}[\log \pi_c^j(a^j|o^j)]b=\nabla_{\Theta^j}J_c(\Theta^i)\end{equation}

\begin{equation}\label{eq:29}\mathbb{E}_{\tau\sim\pi^i}[\log \pi_c^j(a^j|o^j)]b=\nabla_{\Theta^j}(\sum_{t=0}^T\gamma^tr^j_c)=\nabla_{\Theta^j}(\sum_{t=0}^T\gamma^t0)=0\end{equation}

\begin{equation}\label{eq:29}\mathbb{E}_{\tau\sim\pi^i}[\log \pi_c^j(a^j|o^j)]b=0\end{equation}

Since $\mathbb{E}_{\tau\sim\pi^i}[\log \pi_c^j(a^j|o^j)]$ is the entropy of the terminal policy and  it cannot be zero$^*$. There is only one possibility: $b=0 $.

Therefore, $b=0$ contradicts the claim. $\text{  }\square$

Intuitively, $b$ is actually equal to $r^c$ which is obviously zero when there is no collective reward. This result, although quick, shows that an agent can theoretically learn to solve an individual task without conflicting with learned policies for nonstationary coordination behaviours.
\\

\end{document}